\documentclass[accepted]{article}

\usepackage[numbers]{natbib}
\usepackage[preprint]{neurips_2024}

\usepackage{microtype}
\usepackage{graphicx}
\usepackage{subfigure}
\usepackage{booktabs} 

\usepackage[dvipsnames]{xcolor}
\usepackage[colorlinks=true, citecolor=Blue]{hyperref}

\usepackage[textsize=tiny]{todonotes}
\usepackage{url}
\usepackage{color, float}
\usepackage{xspace}
\usepackage{multirow}
\usepackage{fancyhdr}
\usepackage{microtype}
\usepackage{booktabs} 
\usepackage[ruled, vlined, linesnumbered]{algorithm2e}

\usepackage[utf8]{inputenc}
\usepackage{microtype}
\usepackage{graphicx}
\usepackage{booktabs} 
\usepackage{url}            
\usepackage{nicefrac}       
\usepackage{xspace}
\usepackage{enumerate}
\usepackage{amsthm, amsmath, mathtools, amsfonts, amssymb, dsfont, enumitem, bm, bbm, mathabx}
\usepackage{graphicx}
\usepackage[mathscr]{euscript}
\usepackage[capitalize,noabbrev]{cleveref}
\usepackage{adjustbox}
\usepackage{footmisc}
\usepackage[skip=5pt]{caption}
\usepackage[skip=3pt]{subcaption}
\usepackage{wrapfig}
\usepackage{titlesec}
\usepackage{soul}
\usepackage{array}
\usepackage[capitalize,noabbrev]{cleveref}
\usepackage{longtable}
\sethlcolor{lightgray}
\renewcommand{\cite}{\citep}

\graphicspath{{./icml_fig/}}

\numberwithin{figure}{section}
\numberwithin{table}{section}
\newcolumntype{P}[1]{>{\centering\arraybackslash}p{#1}}
\newcolumntype{M}[1]{>{\centering\arraybackslash}m{#1}}

\newcommand{\ahat}{\widehat{a}}
\newcommand{\eps}{\varepsilon}
\renewcommand{\hat}{\widehat}
\renewcommand{\bar}{\widebar}

\newcommand{\Doffline}{\mathcal{D}_\text{offline}}

\newcommand{\set}[1]{\left\{#1\right\}}
\newcommand{\shat}{\widehat{s}}
\newcommand{\rhat}{\widehat{r}}

\newcommand{\Qhat}{\hat{Q}}
\newcommand{\geval}{g_\text{eval}}

\newcommand{\R}{\mathbb{R}}

\newcommand{\N}{\mathcal{N}}
\newcommand{\A}{\mathcal{A}}

\newcommand{\E}{\operatorname{\mathbb{E}}}

\renewcommand{\S}{\mathcal{S}}

\newcommand{\norm}[1]{\left\|#1 \right\|}
\newcommand{\x}[1]{{x^{(#1)}}}

\counterwithin{figure}{section}
\counterwithin{table}{section}
\counterwithin{algocf}{section}

\theoremstyle{plain}
\newtheorem{theorem}{Theorem}[section]

\theoremstyle{definition}
\newtheorem{definition}[theorem]{Definition}

\theoremstyle{remark}

\usepackage{titlesec}
\titlespacing{\paragraph}{0pt}{0pt}{5pt}
\title{Diffusion World Model:\\
Future Modeling Beyond Step-by-Step Rollout for Offline Reinforcement Learning} 
\author{%
Zihan Ding \\
 Princeton University
   \And
   Amy Zhang\\
   University of Texas at Austin\\
   Meta\\
   \And
   Yuandong Tian\\
   Meta
   \And
   Qinqing Zheng\\
   Meta
}

\begin{document}
\maketitle

\begin{abstract}
We introduce Diffusion World Model (DWM), a conditional diffusion model capable of predicting multistep future states and rewards concurrently. As opposed to traditional one-step dynamics models, DWM offers long-horizon predictions in a single forward pass, eliminating the need for recursive queries.  We integrate DWM into model-based value estimation~\cite{feinberg2018model}, where the short-term return is simulated by future trajectories sampled from DWM. In the context of offline reinforcement learning, DWM can be viewed as a conservative value regularization through generative modeling. Alternatively, it can be seen as a data source that enables offline Q-learning with synthetic data. Our experiments on the D4RL~\cite{fu2020d4rl} dataset confirm the robustness of DWM to long-horizon simulation. In terms of absolute performance, DWM significantly surpasses one-step dynamics models with a $44\%$ performance gain, and is comparable to or slightly surpassing their model-free counterparts. \looseness=-1
\end{abstract}
\section{Introduction}
\label{sec:intro}

World models are foundational blocks of AI systems to perform planning and reasoning~\cite{ha2018world}. They serve as simulators of real environments that predict the future outcome of certain actions will produce, and policies can be derived from them. Representative example usages of them in model-based reinforcement learning (MBRL) include action searching~\cite{schrittwieser2020mastering, ye2021mastering}, policy optimization within such simulators~\cite{dean2020sample, feinberg2018model, hafner2019dream, sutton1991dyna}, or a combination of both~\cite{chitnis2023iql, hansen2023td, hansen2022temporal}. 

\begin{wrapfigure}{r}{0.5\textwidth}
    \centering
    \includegraphics[width=0.5\columnwidth]{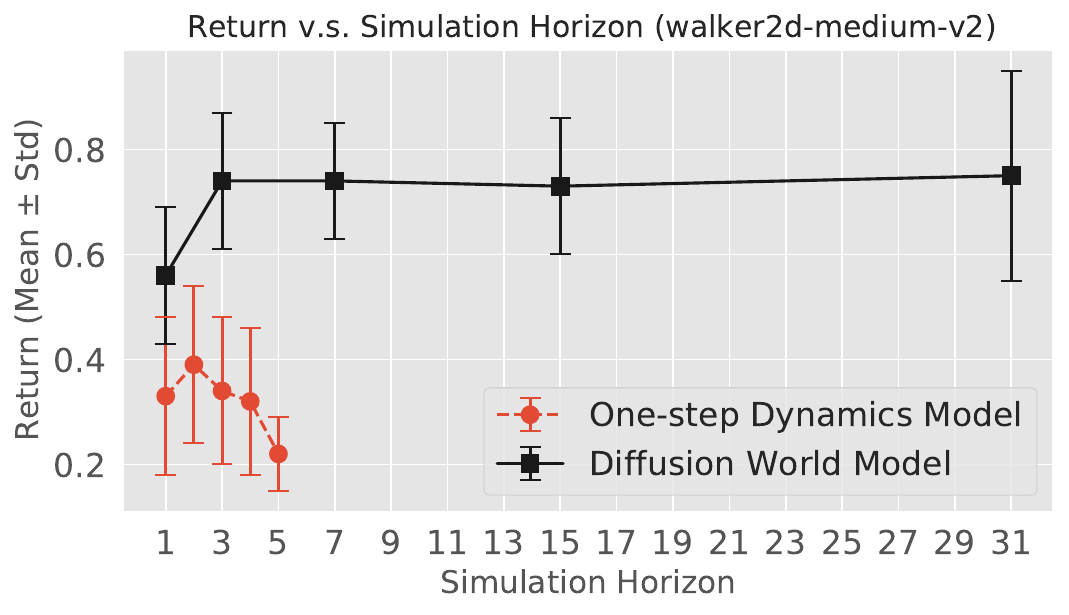}
    \caption{\small The return of TD3+BC trained using diffusion world model and one-step dynamics model.}
    \vspace{-10pt}
    \label{fig:compounding_error_comparison}
\end{wrapfigure}
The prediction accuracy of a world model is critical to the final performance of model-based RL approaches.
Traditional MB methods builds a one-step dynamics model $p_{\text{one}}(s_{t+1}, r_t | s_t, a_t)$ that predicts reward $r_t$ and next state $s_{t+1}$ based on the current state $s_t$ and the current action $a_t$~\cite{hafner2019dream, hafner2020mastering, hafner2023mastering, hansen2022modem, janner2019trust, kaiser2019model,  kidambi2020morel, yu2020mopo}. When planning for multiple steps into the future, $p_\text{one}$ is recursively invoked, leading to a rapid accumulation of errors and unreliable predictions for long-horizon rollouts.
Figure~\ref{fig:compounding_error_comparison} plots the performance of an MB approach with a one-step dynamics model. The return quickly collapses as the rollout length increases, highlighting the issue of \emph{compounding errors} for such models~\cite{asadi2019combating, lambert2022investigating, xiao2019learning}.  

Recently, there has been growing interest in utilizing sequence modeling techniques to solve decision making problems, as seen in various studies~\cite{ajay2022conditional, chen2021decision, janner2022planning, janner2021offline,  micheli2022transformers, robine2023transformer, zheng2023guided,  zheng2022online}. Under this theme, a number of works have proposed Transformer based~\cite{chen2022transdreamer, micheli2022transformers, robine2023transformer} or diffusion model based~\cite{alonso2024diffusion, janner2022planning, lu2023synthetic,rigter2023world, yang2023learning, zhang2023learning} dynamics models, or closely related approaches. As we will review at the end of this section, as well as in Section~\ref{sec:related},
while most existing approaches leverage these sequence models as dynamics models for planning, they model one-step future outcome $s_{t+1}$ and $r_t$ using information of current and previous steps. At planning time, they still plan step by step. 
This raises an intriguing question that our paper seeks to answer:
\begin{quote}
    \emph{Can sequence modeling tools effectively reduce the compounding error in long-horizon prediction via jointly predicting multiple steps into the future?}
\end{quote}
In this paper, we introduce \textbf{\emph{Diffusion World Model}} (DWM). Conditioning on current state $s_t$, action $a_t$, and expected return $g_t$, DWM simultaneously predicts \textit{multistep future states and rewards}. Namely, it models $p_\theta(r_{t}, s_{t+1}, r_{t+1}, \ldots, s_{t+T-1}, r_{t+T-1} | s_t, a_t, g_t)$ where $T$ is the sequence length of the diffusion model.  For planning $H~(H < T)$  steps into the future, DWM only needs to be called once, whereas the traditional one-step model $p_\text{one}$ needs to be invoked $H$ times. This greatly reduces the source of compounding error. As illustrated in Figure~\ref{fig:compounding_error_comparison}, diffusion world model is robust to long-horizon simulation, where the performance does not deteriorate even with simulation horizon $31$. \looseness=-1

To verify the proposed DWM, we consider the \textbf{offline RL} setup, where the objective is to learn a policy from a static dataset without online interactions. The detachment from online training circumvents the side effects of exploration and allows us to investigate the quality of world models thoroughly. We propose a generic Dyna-type~\cite{sutton1991dyna} model-based framework. In brief, 
we first train a diffusion world model using the offline dataset, then train a policy using imagined data generated by the diffusion world model, in an actor-critic manner. Particularly, to generate the target value for training the critic, we introduce \emph{\textbf{Diffusion Model Based Value Expansion} (Diffusion-MVE)} that uses diffusion world model generated future trajectories to simulate the return up to a chosen horizon. As we will elaborate later, \emph{Diffusion-MVE can be interpreted as a value regularization for offline RL through generative modeling, or alternatively, a way to conduct offline Q-learning with synthetic data}.  Our framework is flexible to carry any MF actor-critic RL method of choice, and the output policy is efficient at inference time, as the world model does not intervene with action generation. \looseness=-1

\begin{table}[t]
    \centering
    \resizebox{\textwidth}{!}{
\begin{tabular}{>{\centering\arraybackslash}p{1.9cm}|>{\centering\arraybackslash}p{1.6cm} c >{\centering\arraybackslash}p{5cm} >{\centering\arraybackslash}p{2.5cm}}
    \toprule
      Method   &  RL Setup & Diffusion Model & Model Usages & Action Prediction\\
      \midrule
         SynthER~\cite{lu2023synthetic} & Offline/Online  & $p(s_t, a_t, s_{t+1}, r_t)$ & transition-level data augmentation & MF methods\\ \hline
         DWMs~\cite{alonso2024diffusion} & Online & $p(s_{t+1}|s_t, a_t, \ldots, s_{t-T+1}, a_{t-T+1})$ & step-by-step planning & REINFORCE~\cite{williams1992simple}\\ \hline
         PolyGrad~\cite{rigter2023world} & Online& $p(r_t, s_{t+1}, \ldots, s_{t+T-1}, r_{t+T-1} | s_t, a_t, \ldots, a_{t+T-1})$ & generate on-policy trajectories for policy optimization & stochastic Langevine dynamics\\ \hline
         PGD~\cite{jackson2024policy} & Offline & $p(s_t, a_t, r_t, \ldots, s_{t+T-1}, a_{t+T-1}, r_{t+T-1}, s_{t+T})$ & generate on-policy trajectories with policy gradient guidance for data augmentation & MF offline methods\\ \hline
         UniSim$^*$~\cite{yang2023learning} & Offline & $p(s_{t+1} | s_t, a_t)$  & step-by-step planning & REINFORCE~\cite{williams1992simple} \\ \hline
         Diffuser~\cite{janner2022planning} & Offline & $p(a_t, \ldots, s_{t+T-1}, a_{t+T-1}|s_t)$ & extract $a_t$ from the sample & extract from the sample \\ \hline
         DD~\cite{ajay2022conditional} &Offline &$p(s_{t+1}, \ldots, s_{t+T-1} | s_t, g_t)$  & extract  $s_{t+1}$ from the sample & inverse dynamics model\\ \hline
         \textbf{DWM (ours)} &  Offline &$p(r_t, s_{t+1}, r_{t+1}, \ldots, s_{t+T-1}, r_{t+T-1} | s_t, a_t, g_t)$ & multistep planning & MF offline methods\\
    \bottomrule
    \multicolumn{4}{l}{*The observation of UniSim might contain multiple frames, yet the nature of their diffusion model is still a one-step model.} \\
    \end{tabular}
    }
    \caption{A comparison of representative diffusion-model based MBRL methods. }
    \label{tab:diffusion_mbrl_comparison}
\end{table}

Empirically, we benchmark diffusion-based and traditional one-step world models on 9 locomotion tasks from the D4RL datasets~\cite{fu2020d4rl}, where all the tasks are in continuous action and observation spaces.
The predominant results are:
\begin{enumerate}[leftmargin=1em]
    \item Our results confirm that DWM outperform one-step models, where DWM-based algorithms achieves a $44\%$ performance gain. 
    \item We further consider a variant of our approach where the diffusion model is substituted with a Transformer architecture~\cite{vaswani2017attention}. Although Transformer is a sequence model, its inherent autoregressive structure is more prone to compounding error. We confirm that DWM-based algorithms surpass Transformer-based algorithms with a $37.5\%$ performance gain. 
    \item We also compare our algorithm with Decision Diffuser~\cite{ajay2022conditional}, a closely related model-based offline RL method that simulates the state-only trajectory, while predicting actions using an inverse dynamics model. The performance of the two methods are comparable. 

    \item 
    Meanwhile, due to inevitable modeling error, MB methods typically exhibit worse final performance compared with their model-free (MF) counterparts that directly learn policies from interacting with the true environment. 
    Our results show that DWM-based MB algorithms is comparable to or even slightly outperforming its MF counterparts. We believe this stimulates us to conduct research in the space of model-based RL approaches, which come with an advantage of sample efficiency~\cite{dean2020sample, deisenroth2013survey} and thus are potentially more suitable for practical real-world problems.
\end{enumerate}

\paragraph{Key Differences with Other Diffusion-Based Offline RL Methods}
More recently, various forms of diffusion models like~\cite{ajay2022conditional, alonso2024diffusion, jackson2024policy, janner2022planning, lu2023synthetic, rigter2023world, yang2023learning, zhang2023learning} 
have been introduced for world modeling and related works. These works have targeted different data setups (offline or online RL), and utilize diffusion models to model different types of data distributions. When applying to downstream RL tasks, they also have distinct ways to derive a policy.  While Section~\ref{sec:related} will review them in details, we summarize our key distinctions from these works in Table~\ref{tab:diffusion_mbrl_comparison}.

\section{Related Work}
\label{sec:related}
\paragraph{Model-Based RL}
One popular MB technique is action searching.
Using the world model, one simulates the outcomes of candidate actions, which are sampled from proposal distributions or policy priors~\cite{nagabandi2018neural, williams2015model}, and search for the optimal one. This type of approaches
has been successfully applied to games like Atari and Go~\cite{schrittwieser2020mastering, ye2021mastering} and continuous control problems with pixel observations~\cite{hafner2019learning}.
Alternatively, we can optimize the policy through interactions with the world model. This idea originally comes from the Dyna algorithm~\cite{sutton1991dyna}. The primary differences between works in this regime lie in their usages of the model-generated data. For example, Dyna-Q~\cite{sutton1990integrated} and MBPO~\cite{janner2019trust} augment the true environment data by world model generated transitions, and then conduct MF algorithms on either augmented or generated dataset. \citet{feinberg2018model} proposes to improve the value estimation by unrolling the policy within the world model up to a certain horizon. The Dreamer series of work~\cite{hafner2019dream, hafner2020mastering, hafner2023mastering} use the rollout data for both value estimation and policy learning. More recently, \citet{hansen2023td, hansen2022temporal, chitnis2023iql} combine both techniques to solve continuous control problems. As we cannot go over all the MB approaches, we refer readers to \citet{wang2019benchmarking, amos2021model} for more comprehensive review and benchmarks of them. 

Most of the aforementioned approaches rely on simple one-step world models $p_\text{one}(s_{t+1}, r_t|s_t, a_t)$. The Dreamer series of work~\cite{hafner2019dream, hafner2020mastering, hafner2023mastering} use recurrent neural networks (RNN) to engage in past information for predicting the next state.
Lately, \citet{robine2023transformer, micheli2022transformers, chen2022transdreamer} have independently proposed Transformer-based world models as a replacement of RNN. \citet{janner2020gamma} uses a generative model to learn the occupancy measure over future states, which can perform long-horizon rollout with a single forward pass.
\looseness=-1

\paragraph{Offline RL} 
Directly applying online RL methods to offline RL usually leads to poor performance. The failures are typically attributed to the extrapolation error~\cite{fujimoto2019off}.
To address this issue, a number of conservatism notions have been introduced to encourage the policy to stay close to the offline data. For model-free methods, these notions are applied to the value functions~\cite{kumar2020conservative, kostrikov2021offline, garg2023extreme} or to the policies~\cite{wu2019behavior, jaques2019way, kumar2019stabilizing, fujimoto2021minimalist}. Conservatism has also been incorporated into MB techniques through modified MDPs. 
For instance, MOPO~\cite{yu2020mopo} builds upon MBPO and relabels the predicted reward when generating transitions. It subtracts the uncertainty of the world model's prediction from the predicted reward, thereby softly promoting state-action pairs with low-uncertainty outcome. 
In a similar vein, MOReL~\cite{kidambi2020morel} trains policies using a constructed pessimistic MDP with terminal state. 
The agent will be moved to the terminal state if the prediction uncertainty of the world model is high,
and will receive a negative reward as a penalty. 

\paragraph{Sequence Modeling for RL}
There is a surge of recent research interest in applying sequence modeling tools to RL problems.
\cite{chen2021decision, janner2021offline} first consider the offline trajectories as autoregressive sequences and model them using Transformer architectures~\cite{vaswani2017attention}. This has inspired a line of follow-up research, including~\cite{meng2021offline, lee2022multi}. Normalizing flows like diffusion model~\cite{ho2020denoising, sohl2015deep, song2020score}, flow matching~\cite{lipman2022flow} and consistency model~\cite{song2023consistency} have also been incorporated into various RL algorithms, see e.g., ~\cite{chi2023diffusion, ding2023consistency, du2023learning, hansen2023idql, jia2023chain, mishra2023reorientdiff, wang2022diffusion, xu2023controllable}. 
Several recent works have utilized the diffusion model (DM) for world modeling in a variety of ways. 
Here we discuss them and highlight
the key differences between our approach and theirs, see also Table~\ref{tab:diffusion_mbrl_comparison}. 
\citet{alonso2024diffusion} trains a DM-based one-step dynamics model, which predicts the next single state $s_{t+1}$, conditioning on past states $s_{t-T}, \ldots, s_t$ and actions $a_{t-T}, \ldots, a_t$. 
This concept is similarly applied in UniSim~\cite{yang2023learning}. In essence, these models still plan step by step while incorporating information from preivous steps, whereas our model plans multiple future steps at once.  Similarly, \citet{zhang2023learning} trains a discretized DM with masked and noisy input. Despite still predicting step by step at inference time, this work mainly focuses on prediction tasks and does not conduct RL experiments. 
SynthER~\cite{lu2023synthetic} is in the same spirit as MBPO~\cite{janner2019trust}, which models the collected transition-level data distribution via an unconditioned diffusion model, and augments the training dataset by its samples. We focus on simulating the future trajectory for enhancing the model-based value estimation, and our diffusion model is conditioning on $s_t$ and $a_t$.
PolyGRAD~\cite{rigter2023world} learns a DM to predict a sequence of future states $s_{t+1}, \ldots, s_{t+T-1}$  and rewards $r_t, \ldots, r_{t+T-1}$, conditioning on the initial state $s_t$ and corresponding actions $a_{t}, \ldots, a_{t+T-1}$. Given that the actions are also unknown, PolyGRAD alternates between predicting the actions (via stochastic Langevin dynamics using policy score) and denoising the states and rewards during the DM's sampling process. This approach results in generating on-policy trajectories. In contrast, our approach is off-policy, since it does not interact with the policy during the sampling process. 
Policy-guided diffusion (PGD)~\cite{jackson2024policy} shares the same intention as PolyGrad, which is to generate on-policy trajectories. To achieve this, it trains an unconditioned DM using the offline dataset but samples from it under the (classifier) guidance of policy-gradient.
Diffuser~\cite{janner2022planning} and Decision Diffuser (DD)~\cite{ajay2022conditional} are most close to our work, as they also predict future trajectories. However, the modeling details and the usage of generated trajectories significantly differs. Diffuser trains an unconditioned model that predicts both states and actions, resulting in a policy that uses the generated next action directly. DD models state-only future trajectories conditioning on $s_t$, while we model future states and rewards conditioning on $s_t$ and $a_t$. DD predicts the action by an inverse dynamics model given current state and predicted next state, hence the diffusion model needs to be invoked at inference time. Our approach, instead, can connect with any MF offline RL methods that is fast to execute for inference.

\section{Preliminaries}
\label{sec:preliminary}
\paragraph{Offline RL.}
We consider an infinite-horizon Markov decision process (MDP) defined by $(\S, \A, R, P, p_0, \gamma)$,
where $\S$ is the state space, $\A$ is the action space. Let $\Delta(\S)$ be the probability simplex of
the state space. $R: \S \times \A \mapsto \R$ is a deterministic reward function,
$P: \S \times \A \mapsto \Delta(\S)$ defines the probability distribution of transition, $p_0: \S \mapsto \Delta(\S)$ defines the distribution of initial state $s_0$,
and $\gamma \in (0,1)$ is the discount function. The task of RL is to learn a policy $\pi: \S \mapsto \A$ 
that maximizes its return $J(\pi) = \E_{s_0 \sim p_0(s), a_t \sim \pi(\cdot | s_t), s_{t+1} \sim P(\cdot|s_t, a_t) } \left[ \sum_{t=0}^\infty \gamma^t R(s_t, a_t) \right]$.
Given a trajectory $\tau=\set{s_0, a_0, r_0, \ldots, s_{|\tau|}, a_{|\tau|}, r_{|\tau|}}$, where $|\tau|$ is the total number of timesteps, the return-to-go (RTG) at timestep
$t$ is $g_t = \sum_{t'=t}^{|\tau|} \gamma^{t'-t} r_{t'}$.
In offline RL, we are constrained to learn a policy solely from a static dataset generated by certain unknown policies. Throughout this paper,
we use 
$\Doffline$ to denote the offline data distribution and use
$D_\text{offline}$ to denote the offline dataset. \looseness=-1

\paragraph{Diffusion Model.}
Diffusion probabilistic models~\cite{ho2020denoising, sohl2015deep, song2020score} are generative models that create samples from noises by an iterative denoising process. It defines a fixed Markov chain, called the \emph{forward} or \emph{diffusion process}, that iteratively adds Gaussian noise to $\x{k}$ starting from a data point $\x{0}$:
   $\x{k+1}|\x{k} \sim \mathcal{N}\left(\sqrt{1 - \beta_{k}}\x{k}, \beta_{k}\mathbf{I}\right), \; 0 \leq k \leq K-1.$
As the number of diffusion steps $K \rightarrow \infty$, $\x{K}$ essentially becomes a random noise. We learn the corresponding \emph{reverse process} that transforms random noise to data point:
    $\x{k-1}|\x{k} \sim \mathcal{N}\left(\mu_\theta(\x{k}), \Sigma_\theta(\x{k}) \right), \; 1 \leq k \leq K.$
Sampling from a diffusion model amounts to first sampling a random noise $\x{K} \sim \N(0, \mathbf{I})$ then running the reverse process. 
Let $\varphi(z;\mu, \Sigma)$ denote the density function of a random variable $z \sim \N(\mu, \Sigma)$. To learn the reverse process, we parameterize
    $p_\theta(\x{k-1}|\x{k}) = \varphi\left(\x{k-1};\mu_\theta(\x{k}), \Sigma_\theta(\x{k}) \right), \; 1 \leq k \leq K,$
and optimize the variational lower bound of the marginal likelihood $p_\theta(x^{(0):(K)})$.
There are multiple equivalent ways to optimize the lower bound~\cite{kingma2021variational}, 
and we take the noise prediction route as follows. One can rewrite 
$ \x{k} = \sqrt{\bar{\alpha}_k} \x{0} + \sqrt{1 - \bar{\alpha}_k} \eps$,
where $\bar{\alpha}_k = \prod_{k'=1}^K (1 -\beta_{k'})$, and $\eps \sim \N(0, \mathbf{I})$ is the noise injected for $\x{k}$ (before scaling). We then parameterize a neural network $\eps_\theta(\x{k}, k)$ to predict $\eps$ injected for $\x{k}$. Moreover, a conditional variable $y$ can be easily added into both processes via formulating the corresponding density functions $q(\x{k+1}|\x{k}, y)$ and $p_\theta(\x{k-1}|\x{k}, y)$, respectively. We further deploy classifier-free guidance~\cite{ho2022classifier} to promote the conditional information, which essentially learns both conditioned and unconditioned noise predictors. More precisely, we optimize the following loss function:\looseness=-1
\begin{equation}
    \E_{(\x{0}, y), k, \eps, b } \norm{ \eps_\theta\big(\x{k}(\x{0}, \eps), k, (1-b)\cdot y + b \cdot \varnothing \big) - \eps }^2_2,
\label{eq:diffusion_loss}
\end{equation}
where $\x{0}$ and $y$ are the true data point and conditional information sampled from data distribution,
$\eps \sim \N(0, \mathbf{I})$ is the injected noise,
$k$ is the diffusion step sampled uniformly between $1$ and $K$,
$b \sim \text{Bernoulli}(p_\text{uncond})$ is used to indicate whether we will use null condition,
and finally, $\x{k} = \sqrt{\bar{\alpha}_k} x_0 + \sqrt{1 - \bar{\alpha}_k} \eps$.
Algorithm~\ref{algo:diffusion_sampling_general} details how to sample from a guided diffusion model. In section~\ref{sec:method}, we shall introduce the form of $\x{0}$ and $y$ in the context of offline RL, and discuss how we utilize diffusion models to ease planning. \looseness=-1

\section{Diffusion World Model}
\label{sec:method}

In this section, we introduce a general recipe for model-based offline RL with diffusion world model. 
Our framework consists of two training stages, which we will detail in Section~\ref{sec:method_diffusion} and~\ref{sec:method_mbrl}, respectively.
In the first stage, we train a diffusion model to predict a sequence of future states and rewards, conditioning on the current 
state, action and target return. Next, we train an offline policy using an actor-critic method, where we utilize the pretrained diffusion model for model-based value estimation.
Algorithm~\ref{algo:dwm_training}-\ref{algo:diffusion_mbrl} presents this framework with a simple actor-critic algorithm with delayed updates, where we assume a deterministic offline policy. 
Our framework can be easily extended in a variety of ways.
First, we can generalize it to account for stochastic policies. 
Moreover, the actor-critic algorithm we present is of the simplest form. It can be extended to combine with various existing offline learning algorithms. In Section~\ref{sec:expr}, we discuss three instantiations of
Algorithm~\ref{algo:diffusion_mbrl}, which embeds TD3+BC~\cite{fujimoto2021minimalist}, IQL~\cite{kostrikov2021offline} and Q-learning with pessimistic reward~\cite{yu2020mopo} respectively. \looseness=-1

\begin{figure}[t]
\begin{minipage}{0.43\textwidth}
\begin{algorithm}[H]
\DontPrintSemicolon
\small
\caption{Diffusion World Model Training}
\label{algo:dwm_training}
\SetCommentSty{pinktcp}
\textbf{Hyperparameters}: number of diffusion steps $K$, null conditioning probability $p_\text{uncond}$, noise parameters $\bar{\alpha}_k, k\in[K]$\; 
\While{not converged}{
    Sample a length-$T$ subtrajectory from $D_\text{offline}$: 
        $\x{0} \leftarrow (s_t, a_t, r_t, s_{t+1}, r_{t+1}, $\\
        \hskip40pt $ \ldots, s_{t+T-1}, r_{t+T-1})$\;
    Compute RTG $g_t \leftarrow \sum_{h=0}^{T-1}  \gamma^{h} r_{t+h}$\;
    \SetCommentSty{bluetcp}
    \tcp{optimize DWM via Eq.~\eqref{eq:diffusion_loss}}
    Sample $\eps \sim \N(0, \mathbf{I})$ and $k \in [K]$ uniformly\;
    Compute $\x{k} \leftarrow \sqrt{\bar{\alpha}_k} \x{0} + \sqrt{1 - \bar{\alpha}_k}$\;
    $y\leftarrow \varnothing$ with probability $p_\text{uncond}$, otherwise $y\leftarrow g_t$ \;
    Take gradient step on $\nabla_\theta \big\|\eps_\theta(\x{k}, k, y) - \eps\big\|^2_2 $\;
}
\textbf{Return}: diffusion world model $p_\theta$\;
\end{algorithm}
\end{minipage}
\hfill
\begin{minipage}{0.54\textwidth}
\begin{algorithm}[H]
\DontPrintSemicolon
\small
\caption{General Actor-Critic Framework for Offline Model-Based RL with DWM}
\label{algo:diffusion_mbrl}
\SetCommentSty{bluetcp}
\textbf{Input}: pretrained diffusion world model $p_\theta$\;
\textbf{Hyperparameters}: rollout length $H$, conditioning RTG $\geval$, guidance parameter $\omega$, target network update frequency $n$\;
Initialize the actor and critic networks $\pi_\psi$, $Q_\phi$\;
Initialize the weights of target networks $\bar{\psi} \leftarrow \psi$, $\bar{\phi} \leftarrow \phi$\;
\For{$i=1, 2, \ldots$ until convergence}{
    Sample state-action pair $(s_t, a_t)$ from $D_\text{offline}$\;
    \tcp{diffusion model value expansion}
    Sample $\rhat_t, \shat_{t+1}, \rhat_{t+1}, \ldots, \shat_{t+T-1},\rhat_{t+T-1} \sim p_\theta(\cdot | s_t, a_t, \geval)$ with guidance parameter $\omega$\;
    Compute the target $Q$ value $y \leftarrow \sum_{h=0}^{H-1} \gamma^{h} \rhat_{t+h} + \gamma^H Q_{\bar{\phi}}(\shat_{t+H}, \pi_{\bar{\psi}}(\shat_{t+H})) $ \label{algo:diffusion_mve}\;
    Update the critic: $\phi \leftarrow \phi - \eta \nabla_\phi \norm{Q_\phi(s_t, a_t) - y}^2_2$\;
    Update the actor: $\psi \leftarrow \psi  + \eta \nabla_\psi Q_{\phi}(s_t, \pi_\psi(s_t))$\;
    \If{$i$ mod $n$}{
        $\bar{\phi} \leftarrow \bar{\phi} + w (\phi - \bar{\phi}) $ \;
        $\bar{\psi} \leftarrow \bar{\psi} + w (\psi - \bar{\psi}) $ \;
    }
}
\end{algorithm}
\end{minipage}
\end{figure}

\subsection{Conditional Diffusion Model}
\label{sec:method_diffusion}
We train a return-conditioned diffusion model $p_\theta(\x{0}|s_t, a_t, y)$
on length-$T$ subtrajectories,  where the conditioning variable is the RTG of a subtrajectory. That is, $y=g_t$ and $\x{0} = (r_t, s_{t+1}, r_{t+1}, \ldots, s_{t+T-1}, r_{t+T-1})$. 
As introduced in Section~\ref{sec:preliminary}, we employ classifier-free guidance to promote the role of RTG. 
Stage 1 of Algorithm~\ref{algo:diffusion_mbrl} describes
the training procedure in detail.
For the actual usage of the trained diffusion model in the second stage of our pipeline, 
we predict future $T-1$ states and rewards based on a target RTG $\geval$ and also current state $s_t$ and action $a_t$. These predicted states and rewards are used to facilitate the value estimation in policy training, see Section~\ref{sec:method_mbrl}.
As the future actions are not needed, we do not model them in our world model.
To enable the conditioning of $s_t$ and $a_t$, we slightly adjust the standard sampling procedure (Algorithm~\ref{algo:diffusion_sampling_general}),
where we fix $s_t$ and $a_t$ as conditioning for every denoising step in the reverse process, see Algorithm~\ref{algo:diffusion_sampling_rl}.
\looseness=-1

\subsection{Model-Based RL with Diffusion World Model}
\label{sec:method_mbrl}
As summarized in Algorithm~\ref{algo:diffusion_mbrl}, we propose an actor-critic algorithm where the critic is trained on synthetic data generated by the diffusion model, and the actor is then trained with policy evaluation based on the critic. In a nutshell, we estimate the $Q$-value by the sum of a short-term return, simulated by the DWM, and a long-return value, estimated by a proxy state-action value function $\Qhat$ learned through temporal difference (TD) learning. It is worth noting that in our framework, DWM only intervenes the critic training, and Algorithm~\ref{algo:diffusion_mbrl} is general to connect with any MF value-based algorithms. We shall present 3 different instantiations of it in Section~\ref{sec:expr}.
\begin{definition}[\textbf{$H$-step Diffusion Model Value Expansion}]
\label{def:dmve}
Let $(s_t, a_t)$ be a state-action pair. Sample $\rhat_t, \shat_{t+1}$$, \rhat_{t+1}, \ldots,$$ \shat_{t+T-1},\rhat_{t+T-1}$ from the diffusion model $p_\theta(\cdot|s_t, a_t, \geval)$. Let $H$ be the simulation horizon, where $H < T$. The $H$-step \emph{diffusion model value expansion} estimate of the value of $(s_t, a_t)$ is given by
\begin{equation}
    \Qhat_\text{diff}^H(s_t, a_t) = \mathbb{E}_{p_\theta(\cdot|s_t, a_t, \geval)}\left[\textstyle \sum_{h=0}^{H-1} \gamma^{h} \rhat_{t+h} + \gamma^H \Qhat(\shat_{t+H}, \ahat_{t+H})\right],
\end{equation}
where $\ahat_{t+H} = \pi(\shat_{t+H})$ and $\Qhat(\shat_{t+H}, \ahat_{t+H})$ is the proxy value for the final state-action pair. \looseness=-1
\end{definition}
We employ this expansion to compute the target value in TD learning, see Algorithm~\ref{algo:diffusion_mbrl}. 
This mechanism is key to the success of our algorithm and has several appealing properties. \looseness=-1
\begin{enumerate}[leftmargin=1em]
    \item In deploying the standard model-based value expansion~(MVE, \citet{feinberg2018model}),
 the imagined trajectory is derived by recursively querying the one-step dynamics model $p_\text{one}(s_{t+1}, r_t | s_t, a_t)$, which is the root cause of error accumulation. As an advantage over MVE, our DWM generates the imagined trajectory (without actions) as a whole. 
    \item More interestingly, MVE uses the policy predicted action $\ahat_t = \pi(\hat{s}_t)$ when querying $p_\text{one}$. This can be viewed as an on-policy value estimation of $\pi$ in a simulated environment. In contrast, Diffusion-MVE operates in an off-policy manner, as $\pi$ does not influence the sampling process. As we will explore in Section~\ref{sec:expr}, the off-policy diffusion-MVE excels in offline RL, significantly surpassing the performance of one-step-MVE. We will now delve into two interpretations of this, each from a unique perspective. 
\end{enumerate}
\paragraph{(a)} Our approach can be viewed as a policy iteration algorithm, alternating between policy evaluation (Algorithm~\ref{algo:diffusion_mbrl} line 7-9) and policy improvement (line 10) steps. Here, $\Qhat^H_\text{diff}$ is the estimator of the policy value function $Q^\pi$, with adjustable lookahead horizon $H$ and pessimistic or optimistic estimation through changing $g_\text{eval}$. Parameterized $Q_\phi$ is optimized towards the target $\Qhat^H_\text{diff}$ through a mean squared error.
In the context of offline RL, TD learning often lead to overestimation of $Q^\pi$~\cite{thrun2014issues, kumar2020conservative}. This is because $\pi$ might produce out-of-distribution actions, leading to erroneous values for $\Qhat$, and the policy is defined to maximize $\Qhat$. Such overestimation negatively impacts the generalization capability of the resulting policy when it is deployed online. To mitigate this, a broad spectrum of offline RL methods 
apply various forms of regularization to the value function~\cite{garg2023extreme, kostrikov2021offline, kumar2020conservative}, to ensure the resulting policy remains close to the data. As the DWM is trained exclusively on offline data, it can be seen as a synthesis of the behavior policy that generates the offline dataset. In other words, diffusion-MVE introduces a type of \emph{\textbf{value regularization for offline RL through generative modeling}}. 

Moreover, our approach significantly differs from existing value pessimism notions. One challenge of offline RL is that the behavior policy that generates the offline dataset is often of low-to-moderate quality, 
so that the resulting dataset might only contain trajectories with low-to-moderate returns. As a result, many regularization techniques introduced for offline RL are often \emph{overly pessimistic}~\cite{ghasemipour2022so, nakamoto2023cal}. To address this issue, we typically condition on large out-of-distribution (OOD) values of $\geval$ when sampling from the DWM. Putting differently, we ask the DWM to output an imagined trajectory under an \emph{optimistic goal}.  

\paragraph{(b)} Alternatively, we can also view the approach as an offline Q-learning algorithm~\cite{watkins1992q}, where $\Qhat$ is estimating the optimal value function $Q^*$ using off-policy data. Again, the off-policy data is generated by the diffusion model, conditioning on OOD RTG values. In essence, our approach can be characterized as \emph{\textbf{offline $Q$-learning on synthetic data}}.

\paragraph{Comparison with Transformer-based World Models.}  Curious readers may wonder about the key
distinctions between DMW and existing Transformer-based world models~\cite{chen2022transdreamer, micheli2022transformers, robine2023transformer}.
These models, given the current state $s_t$ and action $a_t$,
leverage the autoregressive structure of Transformer to incorporate past information to predict $s_{t+1}$. 
To forecast multiple steps into the future, they must make iterated predictions. 
In contrast, DWM makes long-horizon predictions in a single query. It is worth noting that it is entirely possible
to substitute the diffusion model in our work with a Transformer, and we justify our design choice in Section~\ref{sec:expr_ablation}.\looseness=-1

\section{Experiments}
\label{sec:expr}

Our experiments are designed to answer the following questions. (1) Compared with the one-step dynamics model, does DWM effectively reduces the compounding error and lead to better performance in MBRL? (2) How does the proposed Algorithm~\ref{algo:diffusion_mbrl} compare with other diffusion model model-based methods, and (3) their model-free counterparts?

To answer these questions, we consider three instantiations of Algorithm~\ref{algo:diffusion_mbrl}, where we integrate TD3+BC~\cite{fujimoto2021minimalist}, IQL~\cite{kostrikov2021offline} and Q-learning with pessimistic reward (which we refer to as PQL) as the offline RL algorithm in the second stage. These algorithms come with different conservatism notions defined on the action (TD3+BC), the value function (IQL), and the reward (PQL), respectively.  Specifically, the PQL algorithm is inspired by the MOPO algorithm~\cite{yu2020mopo}, where we penalize the world model predicted reward by the uncertainty of its prediction. Nonetheless, it is distinct from MOPO in the critic learning. MOPO uses standard TD learning on model-generated transitions, whereas we use MVE or Diff-MVE for value estimation. In the sequel, we refer to our algorithms as DWM-TD3BC, DWM-IQL and DWM-PQL respectively. For DWM-IQL, we have observed performance enhancement using a variant of Diff-MVE based on the $\lambda$-return technique~\cite{schulman2015high}, therefore we incorporate it as a default feature. Detailed descriptions of these algorithms are deferred to Appendix~\ref{sec:app_instantiations}. We present the comparisons in Section~\ref{sec:expr_dwm_vs_onestep}-\ref{sec:expr_dwm_vs_mf}, and ablate the design choices we made for DWM in Section~\ref{sec:expr_ablation}. In Appendix~\ref{subsec:add_baselines}, we conduct experimental comparison with additional baselines including data augmentation~\cite{lu2023synthetic} and autoregressive diffusion~\cite{rigter2023world} methods. 

\paragraph{Benchmark and Hyperparameters.} 
We conduct experiments on 9 datasets of locomotion tasks from the D4RL~\cite{fu2020d4rl} benchmark, and report the obtained normalized return (0-1 with 1 as expert performance). Throughout the paper, we train each algorithm for 5 instances with different random seeds, and evaluate them for 10 episodes. All reported values are means and standard deviations aggregated over 5 random seeds. 
We set the sequence length of DWM to be $T=8$ (discussed in Section~\ref{sec:expr_dwm_vs_onestep}). The number of diffusion steps is $K=5$ for training. For DWM inference, an accelerated inference technique is applied with a reduced number of diffusion steps $N=3$, as detailed in Section~\ref{sec:expr_ablation}.
The training and sampling details of DWM refer to Appendix~\ref{subsec:app_diffusion_model}, and the training details of each offline algorithm refer to Appendix~\ref{subsec:app_mf_mb_details}. We further conduct extensive experiments on sparse-reward tasks, and results are detailed in Appendix Sec.~\ref{sec:add_sparse_reward}.

\subsection{DWM v.s. One-Step Dynamics Model}
\label{sec:expr_dwm_vs_onestep}
We first investigate the effectiveness of DWM in reducing the compounding error for MBRL, and compare it with the counterparts using one-step dynamics model. Next, we evaluate the performance of our proposed Algorithm~\ref{algo:diffusion_mbrl} and the one-step dynamics model counterparts, where we substitute DWM by one-step dynamics models and use standard MVE. We call these baselines OneStep-TD3BC, OneStep-IQL and OneStep-PQL, correspondingly. 

\begin{figure}[t]
     \centering
    \includegraphics[width=1\columnwidth]{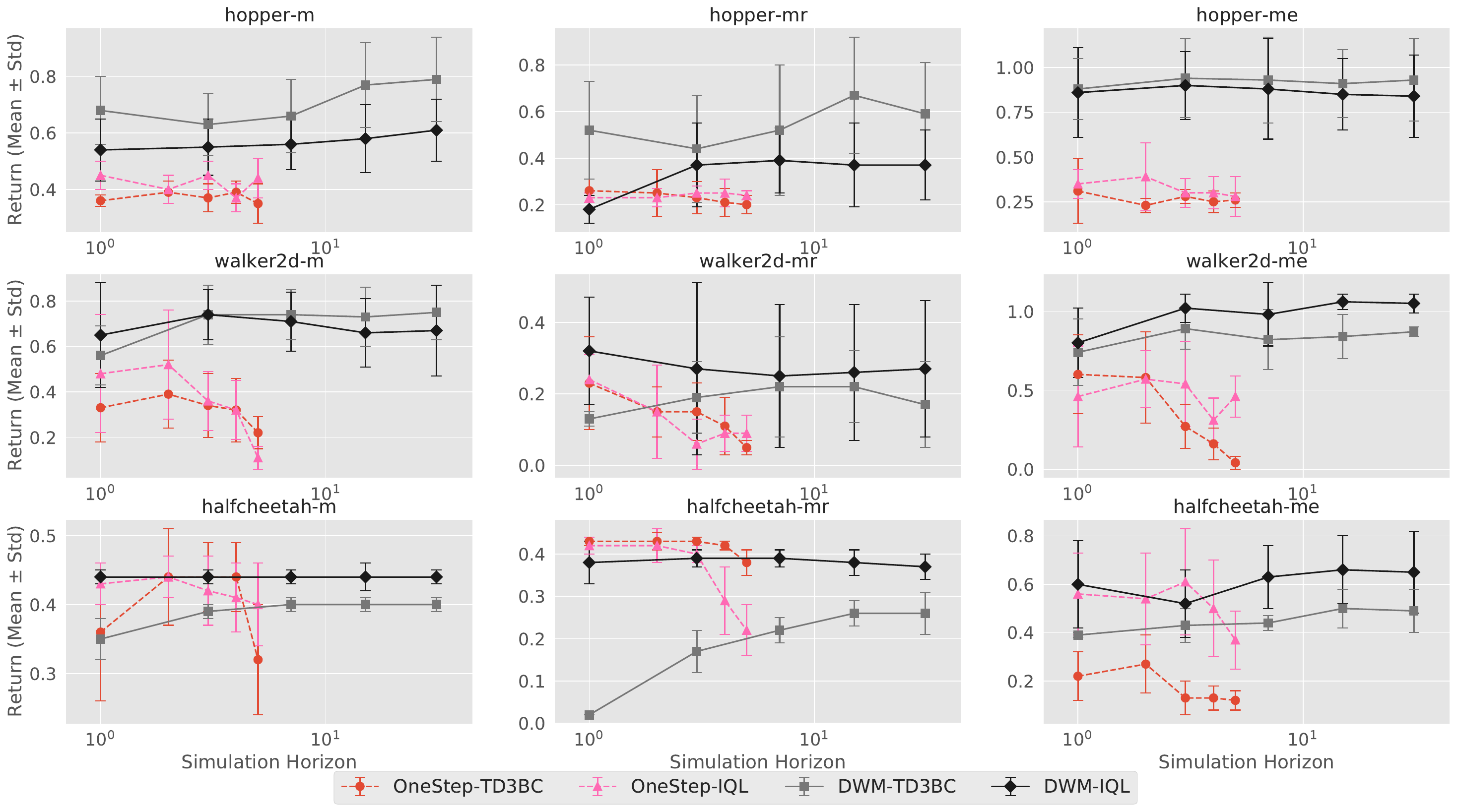}
    \caption{Performances of Algorithm~\ref{algo:diffusion_mbrl} with DWM and one-step models, using different simulation horizons. 
    The x-axis has range $[1, 31]$ in a logarithm scale.}
    \label{fig:one_step_compare}
\end{figure}
\paragraph{Long Horizon Planning and Compounding Error Comparison.} To explore the response of different world models to long simulation horizons, we compare the performance
DWM methods (DWM-TD3BC and DWM-IQL) with their one-step counterparts (OneStep-TD3BC and OneStep-IQL) when the simulation horizon $H$ used in policy training changes.
To explore the limit of DWM models, we train another set of DWMs with longer sequence length $T=32$ and investigate the performance of downstream RL algorithms for $H\in\{1,3,7,15,31\}$.
The algorithms with one-step dynamics models have simulation horizon from 1 to 5. Figure~\ref{fig:one_step_compare} plots the results across 9 tasks.
OneStep-IQL and OneStep-TD3BC exhibit a clearly performance drop as the simulation horizon increases. For most tasks,
their performances peak with relatively short simulation horizons, like one or two. This suggests that longer model-based rollout with one-step dynamics models suffer from severe compounding errors. On the contrary, DWM-TD3BC and DWM-IQL maintain relatively high returns without significant performance degradation, even using horizon length 31.
Note that in the final result Table~\ref{tab:compare_mbrl}, we report results using DWM with sequence length $T=8$, because the performance gain of using $T=32$ is marginal. See Appendix~\ref{app:seq_length_dwm} for details. We additionally conduct experiments on analyzing the compounding error for DWM and one-step model predictions. The results in Appendix~\ref{subsec:compounding} indicate the superior performance of DWM in reducing the compounding errors, which verifies our hypothesis. \looseness=-1

\paragraph{Offline RL Performance.} Table~\ref{tab:compare_mbrl} reports the performance of
Algorithm~\ref{algo:diffusion_mbrl} using DWM and one-step dynamics models on the D4RL datasets. We sweep over the simulation horizon $H\in\{1,3,5,7\}$ and a set of evaluation RTG values. The RTG values we search vary across environments, see Table~\ref{tab:hyperparam}. The predominant trends we found are: \emph{the proposed DWM significantly outperforms the one-step counterparts, with a notable $44\%$ performance gain.} 
This is attributed to the strong expressivity of diffusion models and the prediction of entire sequences all at once, which circumvents the compounding error issue in multistep rollout. This point will be further discussed in the studies of simulation horizon as next paragraph.

\begin{table*}[t]
\centering
\resizebox{0.95\textwidth}{!}{
\begin{tabular}{c|ccc|ccc}
\toprule
 Env.  &  OneStep-TD3BC & OneStep-IQL & OneStep-PQL & DWM-TD3BC & DWM-IQL & DWM-PQL  \\  \midrule
hopper-m  &  0.39 $\pm$ 0.04 & 0.45 $\pm$ 0.05 & 0.63 $\pm$ 0.12 & \textbf{0.65 $\pm$ 0.10} & 0.54 $\pm$ 0.11 & 0.50 $\pm$ 0.09  \\
walker2d-m  &  0.39 $\pm$ 0.15 & 0.52 $\pm$ 0.24 & 0.74 $\pm$ 0.14 & 0.70 $\pm$ 0.15 & 0.76 $\pm$ 0.05 & \textbf{0.79 $\pm$ 0.08}\\
halfcheetah-m &  0.44 $\pm$ 0.05 & 0.44 $\pm$ 0.03 & 0.45 $\pm$ 0.01 & \textbf{0.46 $\pm$ 0.01} & 0.44 $\pm$ 0.01 & 0.44 $\pm$ 0.01   \\
hopper-mr  &  0.26 $\pm$ 0.05 & 0.25 $\pm$ 0.03 & 0.32 $\pm$ 0.03 & 0.53 $\pm$ 0.09 & \textbf{0.61 $\pm$ 0.13} & 0.39 $\pm$ 0.03\\
walker2d-mr &  0.23 $\pm$ 0.13 & 0.24 $\pm$ 0.07 & 0.62 $\pm$ 0.22 & 0.46 $\pm$ 0.19 & 0.35 $\pm$ 0.14 & 0.35 $\pm$ 0.13 \\
halfcheetah-mr & \textbf{0.43 $\pm$ 0.01} & 0.42 $\pm$ 0.02 & 0.42 $\pm$ 0.01 & \textbf{0.43 $\pm$ 0.01} & 0.41 $\pm$ 0.01 & \textbf{0.43 $\pm$ 0.01} \\
hopper-me &  0.31 $\pm$ 0.18 & 0.39 $\pm$ 0.19 & 0.43 $\pm$ 0.18 & \textbf{1.03 $\pm$ 0.14} & 0.90 $\pm$ 0.25 & 0.80 $\pm$ 0.18 \\
walker2d-me &  0.60 $\pm$ 0.25 & 0.57 $\pm$ 0.18 & 0.61 $\pm$ 0.22 & \textbf{1.10 $\pm$ 0.00} & 1.04 $\pm$ 0.10 & \textbf{1.10 $\pm$ 0.01} \\
halfcheetah-me &  0.27 $\pm$ 0.12 & 0.61 $\pm$ 0.22 & 0.61 $\pm$ 0.22 & \textbf{0.75 $\pm$ 0.16} & 0.71 $\pm$ 0.14 & 0.69 $\pm$ 0.13 \\ \hline
\multirow{2}{*}{Average} &  0.368 $\pm$ 0.105 & 0.432 $\pm$ 0.115 & 0.537 $\pm$ 0.128 & \textbf{0.679 $\pm$ 0.098} & 0.641 $\pm$ 0.117 & 0.610 $\pm$ 0.080   \\
\cline{2-7}
& \multicolumn{3}{c|}{0.446$\pm$0.116} & \multicolumn{3}{c}{\textbf{0.643$\pm$0.07}} \\\bottomrule
\end{tabular}
}
\caption{\small Comparison of MB methods with one-step model versus DWM on the D4RL dataset.}
\label{tab:compare_mbrl}
\end{table*}

\subsection{DWM v.s. Decision Diffuser}
\label{sec:expr_dwm_vs_other_mb}

We further compare DWM-TD3BC (the best-performing DWM-based algorithms) with Decision Diffuser (DD)~\cite{ajay2022conditional}, another closely related approach that also use diffusion models to model the trajectory in the offline dataset.
As noted in Section~\ref{sec:intro}, our approach is significantly different from theirs, though. DWM conditions on both state $s_t$ and action $a_t$, where DD only conditions on $s_t$. More importantly, we train a downstream model-free policy using imagined rollout, whereas DD predicts the action via an inverse dynamics model, using current state $s_t$ and predicted next state $\hat{s}_{t+1}$. This means, at inference time, DD needs to generate the whole trajectory, which is computationally inefficient. On the contrary, DWM based approaches are efficient as we do not need to sample from the trained DWM anymore. Table~\ref{tab:compare_diffusion_dd} reports the performance and the inference time of DWM-TD3BC and DD. The inference time is averaged over $600$ evaluation episodes. The performance of DWM-TD3BC is comparable to Decision Diffuser, and it enjoys $4.6$x faster inference speed. We anticipate the difference in speed amplifies for higher dimensional problems.
\begin{table*}
    \centering
\resizebox{0.95\textwidth}{!}{
\small
\begin{tabular}{c|cc | cccc}
\toprule
 Env.  &  \multicolumn{2}{c|}{DD} & \multicolumn{2}{c}{DWM-TD3BC}  \\  
  & Normalized Return & Inference Time (sec) & Normalized Return & Inference Time (sec)\\
  \midrule
hopper-m  & 0.49 $\pm$ 0.07 & 4.11 $\pm$ 4.64 & \textbf{0.65 $\pm$ 0.10} & \textbf{1.18 $\pm$ 0.51}\\
walker2d-m & 0.67 $\pm$ 0.16  & 8.09 $\pm$ 1.24 & 0.70 $\pm$ 0.15 &  \textbf{2.00 $\pm$ 0.63} \\ 
halfcheetah-m & \textbf{0.49 $\pm$ 0.01}  &8.18 $\pm$ 3.77 & 0.46 $\pm$ 0.01 &  \textbf{1.81 $\pm$ 0.54} \\
hopper-mr  & \textbf{0.66 $\pm$ 0.15}  & 6.21 $\pm$ 4.21  & 0.53 $\pm$ 0.09 & \textbf{0.64 $\pm$ 0.52}\\
walker2d-mr  & 0.44 $\pm$ 0.26 &5.94 $\pm$ 4.32 & \textbf{0.46 $\pm$ 0.19} & \textbf{0.83 $\pm$ 0.45} \\
halfcheetah-mr  & 0.38 $\pm$ 0.06 &7.46 $\pm$ 9.72  & \textbf{0.43 $\pm$ 0.01} & \textbf{0.60 $\pm$ 0.17} \\
hopper-me  & \textbf{1.06 $\pm$ 0.11} & 8.82 $\pm$ 2.96 & 1.03$\pm$ 0.14 & \textbf{2.62 $\pm$ 1.22} \\
walker2d-me  & 0.99 $\pm$ 0.15  & 9.26 $\pm$ 1.30 &\textbf{1.10 $\pm$ 0.00}  & \textbf{3.60  $\pm$ 3.90} \\
halfcheetah-me  & \textbf{0.91 $\pm$ 0.01} &9.53 $\pm$ 2.50  & 0.75 $\pm$ 0.16 &  \textbf{3.77 $\pm$ 2.43} \\ \hline
{Average} & \textbf{0.677$\pm$0.109} & 7.531 $\pm$ 3.651 & \textbf{0.679$\pm$0.098} & \textbf{1.620 $\pm$ 1.379} \\
\bottomrule
\end{tabular}
}
\caption{\small The performance of DWM-TD3BC and Decision Diffuser (DD) are comparable, while DWM-TD3BC is $4.6$x faster than DD. }
\label{tab:compare_diffusion_dd}
\end{table*}

\subsection{DWM v.s. Model-Free Counterparts}
\label{sec:expr_dwm_vs_mf} 

Finally, we compare DWM-based algorithms with their MF counterparts, namely, TD3+BC vs DWM-TD3BC, and IQL vs DWM-IQL. Table~\ref{tab:dwm_vs_mf} reports the results. For each group of comparison, we highlight the winning performance. 

\begin{wraptable}{r}{0.56\textwidth}
\centering
\resizebox{0.56\textwidth}{!}{
\begin{tabular}{c|cc|cc}
\toprule
 Env.          &  TD3+BC & DWM-TD3BC                 & IQL  & DWM-IQL \\  \midrule
hopper-m       & 0.58 $\pm$ 0.11 &  \textbf{0.65 $\pm$ 0.10}  & 0.48 $\pm$ 0.08 & \textbf{0.54 $\pm$ 0.11} \\
walker2d-m     & \textbf{0.77 $\pm$ 0.09} &  0.70 $\pm$ 0.15  & 0.75 $\pm$ 0.15 & \textbf{0.76 $\pm$ 0.05} \\
halfcheetah-m  & \textbf{0.47 $\pm$ 0.01} &  0.46 $\pm$ 0.01 & \textbf{0.46 $\pm$ 0.07}  & 0.44 $\pm$ 0.01\\
hopper-mr      & 0.53 $\pm$ 0.19 & \textbf{0.53 $\pm$ 0.09}  & 0.25 $\pm$ 0.02& \textbf{0.61 $\pm$ 0.13}\\
walker2d-mr    & \textbf{0.75 $\pm$ 0.19} & 0.46 $\pm$ 0.19&  \textbf{0.48 $\pm$ 0.23}  & 0.35 $\pm$ 0.14 \\
halfcheetah-mr & \textbf{0.43 $\pm$ 0.01}  & \textbf{0.43 $\pm$ 0.01}& \textbf{0.44 $\pm$ 0.01}  & 0.41 $\pm$ 0.01\\
hopper-me      & 0.90 $\pm$ 0.28  & \textbf{1.03 $\pm$ 0.14} & 0.86 $\pm$ 0.22& \textbf{0.90 $\pm$ 0.25}\\
walker2d-me    & 1.08 $\pm$ 0.01  & \textbf{1.10 $\pm$ 0.00} & \textbf{1.09 $\pm$ 0.00} & 1.04 $\pm$ 0.10 \\
halfcheetah-me & 0.73 $\pm$ 0.16  &\textbf{0.75 $\pm$ 0.16} & 0.60 $\pm$ 0.23  & \textbf{0.71 $\pm$ 0.14}\\ \hline
Average        & \textbf{0.693 $\pm$ 0.116} & 0.679 $\pm$ 0.098 & 0.601 $\pm$ 0.112  & \textbf{0.641 $\pm$ 0.117}\\
\bottomrule
\end{tabular}
}
\caption{\small Performance of DWM methods and its MF counterparts.}
\label{tab:dwm_vs_mf}
\end{wraptable}
We can see DWM-IQL outperforms IQL, DWM-TD3BC is comparable to TD3+BC. 
Different from MF algorithms with ground-truth samples, MB algorithms inevitably suffers additional modeling errors from approximating the dynamics. It is worth noting that MB algorithms using traditional one-step dynamics model significantly underperforms the MF counterparts (results reported in Table~\ref{tab:compare_mbrl}), while DWM alleviates the downside of dynamics modeling through reducing compounding errors. \looseness=-1

\subsection{Ablation Studies}
\label{sec:expr_ablation}

\paragraph{Diffusion Model v.s. Transformer.}
Algorithm~\ref{algo:diffusion_mbrl} is capable of accommodating various types of sequence models, including
Transformer~\cite{vaswani2017attention}, one of the most successful sequence models.
However, analogous to the compounding error issue for one-step dynamics model, 
Transformer is subject to inherent error accumulation due to its autoregressive structure.
Therefore, we hypothesize Transformer will underperform and choose diffusion model.
\begin{wraptable}{r}{0.5\textwidth}
\resizebox{0.5\textwidth}{!}{
\begin{tabular}{c|cc|cc}
\toprule
\small
 Env. & \multicolumn{1}{c}{\textbf{T-TD3BC}} & \multicolumn{1}{c|}{\textbf{T-IQL}} & \multicolumn{1}{c}{\textbf{DWM-TD3BC}} & \multicolumn{1}{c}{\textbf{DWM-IQL}} \\ \midrule  
hopper-m  & 0.58 $\pm$ 0.08 & 0.55 $\pm$ 0.08 & \textbf{0.65 $\pm$ 0.10} & 0.54 $\pm$ 0.11 \\
walker2d-m & 0.60 $\pm$ 0.16 & 0.72 $\pm$ 0.12  & 0.70 $\pm$ 0.15 &  \textbf{0.76 $\pm$ 0.05}\\
halfcheetah-m & 0.42 $\pm$ 0.03 & 0.43 $\pm$ 0.01  & \textbf{0.46 $\pm$ 0.01} & 0.44 $\pm$ 0.01    \\
hopper-mr & 0.25 $\pm$ 0.06 & 0.26 $\pm$ 0.09  & 0.53 $\pm$ 0.09 & \textbf{0.61 $\pm$ 0.13} \\
walker2d-mr & 0.13 $\pm$ 0.06 & 0.23 $\pm$ 0.12 & \textbf{0.46 $\pm$ 0.19} & 0.35 $\pm$ 0.14  \\
halfcheetah-mr & 0.40 $\pm$ 0.01 & 0.39 $\pm$ 0.01 & \textbf{0.43 $\pm$ 0.01} & 0.41 $\pm$ 0.01 \\
hopper-me & 0.66 $\pm$ 0.25 & 0.62 $\pm$ 0.16  & \textbf{1.03$\pm$ 0.14} & 0.90 $\pm$ 0.25  \\
walker2d-me & 0.58 $\pm$ 0.15 & 1.03 $\pm$ 0.09 & \textbf{1.10 $\pm$ 0.00} & 1.04 $\pm$ 0.10\\
halfcheetah-me & 0.36 $\pm$ 0.17 & 0.44 $\pm$ 0.08 & \textbf{0.75 $\pm$ 0.16} & 0.71 $\pm$ 0.14  \\ \hline
Avg. & 0.442$\pm$0.101 &  0.519$\pm$0.084 & \textbf{0.679 $\pm$ 0.098} & 0.641 $\pm$ 0.117   \\
\bottomrule
\end{tabular}
}
\caption{\small Performance of Algorithm~\ref{algo:diffusion_mbrl} using DWM and Transformer-based world models. }
\label{tab:rollout_compare_dt}
\end{wraptable}
To verify this hypothesis, we replace the diffusion model with Transformer in our proposed algorithms,
and compare the resulting performance with DWM methods. We particularly consider the combination with
TD3+BC and IQL, where we call the obtained algorithms T-TD3BC and T-IQL.
We test T-TD3BC and T-IQL with parameter sweeping over simulation horizon $H\in\{1,3,5,7\}$, as the same as DWM methods. For the evaluation RTG, we take the value used in Decision Transformer~\cite{chen2021decision} and apply the same normalization as used for DWM.
As shown in Table~\ref{tab:rollout_compare_dt}, DWM consistently outperforms Transformer-based world models across offline RL algorithm instantiations and environments. The experiment details refer to Appendix~\ref{subsec:app_transformer}. \looseness=-1

\paragraph{Additional Ablation Experiments.} Due to the space limit, we refer the readers to Appendix for other ablation experiments regarding other  design choices of our approach. Appendix~\ref{app:diffusion_steps} discusses the number of diffusion steps we use in training DWM and trajectory sampling. Appendix~\ref{app:ood_rtg} discusses the evaluation RTG values we use when sampling from the DWM. Appendix~\ref{app:lambda_return} ablates the $\lambda$-return technique we incorporate for DWM-IQL. Last, we also investigate the effects of fine-tuning DWM with relabelled RTGs~\cite{yamagata2023q}. We have found this technique is of limited utility, so we did not include it in the final design for the simplicity of our algorithm. See the results in Appendix~\ref{app:rtg_relabel}.

\section{Conclusion and Future Work}
We present a general framework of leveraging diffusion models as world models, in the context of offline RL. This framework can be easily extended to accommodate online training. Specifically, we utilize DWM generated trajectories for model-based value estimation.  
Our experiments show that this approach effectively reduces the compounding error in MBRL. We benchmarked DWM against the traditional one-step dynamics model, by training 3 different types of offline RL algorithms using imagined trajectories generated by each of them. 
DWM demonstrates a notable performance gain. DWM-based approaches are also caparable with or marginally outperforming their MF counterparts.
However, there are also limitations of our work. 
Currently, DWM is trained for each individual environment and is task-specific. An intriguing avenue for future research would be extending DWM to multi-environment and multi-task settings. Additionally, to circumvent the side effects of exploration, we only investigate DWM in the offline RL setting. This raises an interesting question regarding the performance of DWM in online settings. Lastly but most importantly, although we adopt 
the stride sampling technique to accelerate the inference, the computational demand of DWM remains high. Further enhancements to speed up the sampling process could be crucial for future usages of DWM to tackle larger scale problems.

\bibliography{example_paper}

\begin{thebibliography}{}

\bibitem[Ajay et~al., 2022]{ajay2022conditional}
Ajay, A., Du, Y., Gupta, A., Tenenbaum, J., Jaakkola, T., and Agrawal, P. (2022).
\newblock Is conditional generative modeling all you need for decision-making?
\newblock {\em arXiv preprint arXiv:2211.15657}.

\bibitem[Alonso et~al., 2024]{alonso2024diffusion}
Alonso, E., Jelley, A., Kanervisto, A., and Pearce, T. (2024).
\newblock Diffusion world models.

\bibitem[Amos et~al., 2021]{amos2021model}
Amos, B., Stanton, S., Yarats, D., and Wilson, A.~G. (2021).
\newblock On the model-based stochastic value gradient for continuous reinforcement learning.
\newblock In {\em Learning for Dynamics and Control}, pages 6--20. PMLR.

\bibitem[Asadi et~al., 2019]{asadi2019combating}
Asadi, K., Misra, D., Kim, S., and Littman, M.~L. (2019).
\newblock Combating the compounding-error problem with a multi-step model.
\newblock {\em arXiv preprint arXiv:1905.13320}.

\bibitem[Chen et~al., 2022]{chen2022transdreamer}
Chen, C., Wu, Y.-F., Yoon, J., and Ahn, S. (2022).
\newblock Transdreamer: Reinforcement learning with transformer world models.
\newblock {\em arXiv preprint arXiv:2202.09481}.

\bibitem[Chen et~al., 2021]{chen2021decision}
Chen, L., Lu, K., Rajeswaran, A., Lee, K., Grover, A., Laskin, M., Abbeel, P., Srinivas, A., and Mordatch, I. (2021).
\newblock Decision transformer: Reinforcement learning via sequence modeling.
\newblock In {\em Thirty-Fifth Conference on Neural Information Processing Systems}.

\bibitem[Chi et~al., 2023]{chi2023diffusion}
Chi, C., Feng, S., Du, Y., Xu, Z., Cousineau, E., Burchfiel, B., and Song, S. (2023).
\newblock Diffusion policy: Visuomotor policy learning via action diffusion.
\newblock {\em arXiv preprint arXiv:2303.04137}.

\bibitem[Chitnis et~al., 2023]{chitnis2023iql}
Chitnis, R., Xu, Y., Hashemi, B., Lehnert, L., Dogan, U., Zhu, Z., and Delalleau, O. (2023).
\newblock Iql-td-mpc: Implicit q-learning for hierarchical model predictive control.
\newblock {\em arXiv preprint arXiv:2306.00867}.

\bibitem[Dean et~al., 2020]{dean2020sample}
Dean, S., Mania, H., Matni, N., Recht, B., and Tu, S. (2020).
\newblock On the sample complexity of the linear quadratic regulator.
\newblock {\em Foundations of Computational Mathematics}, 20(4):633--679.

\bibitem[Deisenroth et~al., 2013]{deisenroth2013survey}
Deisenroth, M.~P., Neumann, G., Peters, J., et~al. (2013).
\newblock A survey on policy search for robotics.
\newblock {\em Foundations and Trends{\textregistered} in Robotics}, 2(1--2):1--142.

\bibitem[Ding and Jin, 2023]{ding2023consistency}
Ding, Z. and Jin, C. (2023).
\newblock Consistency models as a rich and efficient policy class for reinforcement learning.
\newblock {\em arXiv preprint arXiv:2309.16984}.

\bibitem[Du et~al., 2023]{du2023learning}
Du, Y., Yang, M., Dai, B., Dai, H., Nachum, O., Tenenbaum, J., Schuurmans, D., and Abbeel, P. (2023).
\newblock Learning universal policies via text-guided video generation.
\newblock {\em arXiv preprint arXiv:2302.00111}.

\bibitem[Emmons et~al., 2021]{emmons2021rvs}
Emmons, S., Eysenbach, B., Kostrikov, I., and Levine, S. (2021).
\newblock Rvs: What is essential for offline rl via supervised learning?
\newblock {\em arXiv preprint arXiv:2112.10751}.

\bibitem[Feinberg et~al., 2018]{feinberg2018model}
Feinberg, V., Wan, A., Stoica, I., Jordan, M.~I., Gonzalez, J.~E., and Levine, S. (2018).
\newblock Model-based value estimation for efficient model-free reinforcement learning.
\newblock {\em arXiv preprint arXiv:1803.00101}.

\bibitem[Fu et~al., 2020]{fu2020d4rl}
Fu, J., Kumar, A., Nachum, O., Tucker, G., and Levine, S. (2020).
\newblock D4rl: Datasets for deep data-driven reinforcement learning.
\newblock {\em arXiv preprint arXiv:2004.07219}.

\bibitem[Fujimoto and Gu, 2021]{fujimoto2021minimalist}
Fujimoto, S. and Gu, S. (2021).
\newblock A minimalist approach to offline reinforcement learning.
\newblock In {\em Thirty-Fifth Conference on Neural Information Processing Systems}.

\bibitem[Fujimoto et~al., 2018]{fujimoto2018addressing}
Fujimoto, S., Hoof, H., and Meger, D. (2018).
\newblock Addressing function approximation error in actor-critic methods.
\newblock In {\em International conference on machine learning}, pages 1587--1596. PMLR.

\bibitem[Fujimoto et~al., 2019]{fujimoto2019off}
Fujimoto, S., Meger, D., and Precup, D. (2019).
\newblock Off-policy deep reinforcement learning without exploration.
\newblock In {\em International Conference on Machine Learning}, pages 2052--2062. PMLR.

\bibitem[Garg et~al., 2023]{garg2023extreme}
Garg, D., Hejna, J., Geist, M., and Ermon, S. (2023).
\newblock Extreme q-learning: Maxent rl without entropy.
\newblock {\em arXiv preprint arXiv:2301.02328}.

\bibitem[Ghasemipour et~al., 2022]{ghasemipour2022so}
Ghasemipour, K., Gu, S.~S., and Nachum, O. (2022).
\newblock Why so pessimistic? estimating uncertainties for offline rl through ensembles, and why their independence matters.
\newblock {\em Advances in Neural Information Processing Systems}, 35:18267--18281.

\bibitem[Ha and Schmidhuber, 2018]{ha2018world}
Ha, D. and Schmidhuber, J. (2018).
\newblock World models.
\newblock {\em arXiv preprint arXiv:1803.10122}.

\bibitem[Hafner et~al., 2019a]{hafner2019dream}
Hafner, D., Lillicrap, T., Ba, J., and Norouzi, M. (2019a).
\newblock Dream to control: Learning behaviors by latent imagination.
\newblock {\em arXiv preprint arXiv:1912.01603}.

\bibitem[Hafner et~al., 2019b]{hafner2019learning}
Hafner, D., Lillicrap, T., Fischer, I., Villegas, R., Ha, D., Lee, H., and Davidson, J. (2019b).
\newblock Learning latent dynamics for planning from pixels.
\newblock In {\em International conference on machine learning}, pages 2555--2565. PMLR.

\bibitem[Hafner et~al., 2020]{hafner2020mastering}
Hafner, D., Lillicrap, T., Norouzi, M., and Ba, J. (2020).
\newblock Mastering atari with discrete world models.
\newblock {\em arXiv preprint arXiv:2010.02193}.

\bibitem[Hafner et~al., 2023]{hafner2023mastering}
Hafner, D., Pasukonis, J., Ba, J., and Lillicrap, T. (2023).
\newblock Mastering diverse domains through world models.
\newblock {\em arXiv preprint arXiv:2301.04104}.

\bibitem[Hansen et~al., 2022a]{hansen2022modem}
Hansen, N., Lin, Y., Su, H., Wang, X., Kumar, V., and Rajeswaran, A. (2022a).
\newblock Modem: Accelerating visual model-based reinforcement learning with demonstrations.
\newblock {\em arXiv preprint arXiv:2212.05698}.

\bibitem[Hansen et~al., 2023]{hansen2023td}
Hansen, N., Su, H., and Wang, X. (2023).
\newblock Td-mpc2: Scalable, robust world models for continuous control.
\newblock {\em arXiv preprint arXiv:2310.16828}.

\bibitem[Hansen et~al., 2022b]{hansen2022temporal}
Hansen, N., Wang, X., and Su, H. (2022b).
\newblock Temporal difference learning for model predictive control.
\newblock {\em arXiv preprint arXiv:2203.04955}.

\bibitem[Hansen-Estruch et~al., 2023]{hansen2023idql}
Hansen-Estruch, P., Kostrikov, I., Janner, M., Kuba, J.~G., and Levine, S. (2023).
\newblock Idql: Implicit q-learning as an actor-critic method with diffusion policies.
\newblock {\em arXiv preprint arXiv:2304.10573}.

\bibitem[Ho et~al., 2020]{ho2020denoising}
Ho, J., Jain, A., and Abbeel, P. (2020).
\newblock Denoising diffusion probabilistic models.
\newblock {\em Advances in neural information processing systems}, 33:6840--6851.

\bibitem[Ho and Salimans, 2022]{ho2022classifier}
Ho, J. and Salimans, T. (2022).
\newblock Classifier-free diffusion guidance.
\newblock {\em arXiv preprint arXiv:2207.12598}.

\bibitem[Jackson et~al., 2024]{jackson2024policy}
Jackson, M.~T., Matthews, M.~T., Lu, C., Ellis, B., Whiteson, S., and Foerster, J. (2024).
\newblock Policy-guided diffusion.
\newblock {\em arXiv preprint arXiv:2404.06356}.

\bibitem[Janner et~al., 2022]{janner2022planning}
Janner, M., Du, Y., Tenenbaum, J.~B., and Levine, S. (2022).
\newblock Planning with diffusion for flexible behavior synthesis.
\newblock {\em arXiv preprint arXiv:2205.09991}.

\bibitem[Janner et~al., 2019]{janner2019trust}
Janner, M., Fu, J., Zhang, M., and Levine, S. (2019).
\newblock When to trust your model: Model-based policy optimization.
\newblock {\em Advances in neural information processing systems}, 32.

\bibitem[Janner et~al., 2021]{janner2021offline}
Janner, M., Li, Q., and Levine, S. (2021).
\newblock Offline reinforcement learning as one big sequence modeling problem.
\newblock In {\em Thirty-Fifth Conference on Neural Information Processing Systems}.

\bibitem[Janner et~al., 2020]{janner2020gamma}
Janner, M., Mordatch, I., and Levine, S. (2020).
\newblock gamma-models: Generative temporal difference learning for infinite-horizon prediction.
\newblock {\em Advances in Neural Information Processing Systems}, 33:1724--1735.

\bibitem[Jaques et~al., 2019]{jaques2019way}
Jaques, N., Ghandeharioun, A., Shen, J.~H., Ferguson, C., Lapedriza, A., Jones, N., Gu, S., and Picard, R. (2019).
\newblock Way off-policy batch deep reinforcement learning of implicit human preferences in dialog.
\newblock {\em arXiv preprint arXiv:1907.00456}.

\bibitem[Jia et~al., 2023]{jia2023chain}
Jia, Z., Liu, F., Thumuluri, V., Chen, L., Huang, Z., and Su, H. (2023).
\newblock Chain-of-thought predictive control.
\newblock {\em arXiv preprint arXiv:2304.00776}.

\bibitem[Kaiser et~al., 2019]{kaiser2019model}
Kaiser, L., Babaeizadeh, M., Milos, P., Osinski, B., Campbell, R.~H., Czechowski, K., Erhan, D., Finn, C., Kozakowski, P., Levine, S., et~al. (2019).
\newblock Model-based reinforcement learning for atari.
\newblock {\em arXiv preprint arXiv:1903.00374}.

\bibitem[Kidambi et~al., 2020]{kidambi2020morel}
Kidambi, R., Rajeswaran, A., Netrapalli, P., and Joachims, T. (2020).
\newblock Morel: Model-based offline reinforcement learning.
\newblock {\em Advances in neural information processing systems}, 33:21810--21823.

\bibitem[Kingma et~al., 2021]{kingma2021variational}
Kingma, D., Salimans, T., Poole, B., and Ho, J. (2021).
\newblock Variational diffusion models.
\newblock {\em Advances in neural information processing systems}, 34:21696--21707.

\bibitem[Kostrikov et~al., 2021]{kostrikov2021offline}
Kostrikov, I., Nair, A., and Levine, S. (2021).
\newblock Offline reinforcement learning with implicit q-learning.

\bibitem[Kumar et~al., 2019]{kumar2019stabilizing}
Kumar, A., Fu, J., Tucker, G., and Levine, S. (2019).
\newblock Stabilizing off-policy q-learning via bootstrapping error reduction.
\newblock {\em arXiv preprint arXiv:1906.00949}.

\bibitem[Kumar et~al., 2020]{kumar2020conservative}
Kumar, A., Zhou, A., Tucker, G., and Levine, S. (2020).
\newblock Conservative q-learning for offline reinforcement learning.
\newblock {\em arXiv preprint arXiv:2006.04779}.

\bibitem[Lambert et~al., 2022]{lambert2022investigating}
Lambert, N., Pister, K., and Calandra, R. (2022).
\newblock Investigating compounding prediction errors in learned dynamics models.
\newblock {\em arXiv preprint arXiv:2203.09637}.

\bibitem[Lee et~al., 2022]{lee2022multi}
Lee, K.-H., Nachum, O., Yang, M., Lee, L., Freeman, D., Xu, W., Guadarrama, S., Fischer, I., Jang, E., Michalewski, H., et~al. (2022).
\newblock Multi-game decision transformers.
\newblock {\em arXiv preprint arXiv:2205.15241}.

\bibitem[Lipman et~al., 2022]{lipman2022flow}
Lipman, Y., Chen, R.~T., Ben-Hamu, H., Nickel, M., and Le, M. (2022).
\newblock Flow matching for generative modeling.
\newblock {\em arXiv preprint arXiv:2210.02747}.

\bibitem[Lu et~al., 2023]{lu2023synthetic}
Lu, C., Ball, P.~J., and Parker-Holder, J. (2023).
\newblock Synthetic experience replay.
\newblock {\em arXiv preprint arXiv:2303.06614}.

\bibitem[Meng et~al., 2021]{meng2021offline}
Meng, L., Wen, M., Yang, Y., Le, C., Li, X., Zhang, W., Wen, Y., Zhang, H., Wang, J., and Xu, B. (2021).
\newblock Offline pre-trained multi-agent decision transformer: One big sequence model conquers all starcraftii tasks.
\newblock {\em arXiv preprint arXiv:2112.02845}.

\bibitem[Micheli et~al., 2022]{micheli2022transformers}
Micheli, V., Alonso, E., and Fleuret, F. (2022).
\newblock Transformers are sample efficient world models.
\newblock {\em arXiv preprint arXiv:2209.00588}.

\bibitem[Mish, 2019]{mish2019self}
Mish, M.~D. (2019).
\newblock A self regularized non-monotonic activation function [j].
\newblock {\em arXiv preprint arXiv:1908.08681}.

\bibitem[Mishra and Chen, 2023]{mishra2023reorientdiff}
Mishra, U.~A. and Chen, Y. (2023).
\newblock Reorientdiff: Diffusion model based reorientation for object manipulation.
\newblock {\em arXiv preprint arXiv:2303.12700}.

\bibitem[Nagabandi et~al., 2018]{nagabandi2018neural}
Nagabandi, A., Kahn, G., Fearing, R.~S., and Levine, S. (2018).
\newblock Neural network dynamics for model-based deep reinforcement learning with model-free fine-tuning.
\newblock In {\em 2018 IEEE international conference on robotics and automation (ICRA)}, pages 7559--7566. IEEE.

\bibitem[Nakamoto et~al., 2023]{nakamoto2023cal}
Nakamoto, M., Zhai, Y., Singh, A., Mark, M.~S., Ma, Y., Finn, C., Kumar, A., and Levine, S. (2023).
\newblock Cal-ql: Calibrated offline rl pre-training for efficient online fine-tuning.
\newblock {\em arXiv preprint arXiv:2303.05479}.

\bibitem[Nguyen et~al., 2022]{nguyen2022conserweightive}
Nguyen, T., Zheng, Q., and Grover, A. (2022).
\newblock Conserweightive behavioral cloning for reliable offline reinforcement learning.
\newblock {\em arXiv preprint arXiv:2210.05158}.

\bibitem[Nichol and Dhariwal, 2021]{nichol2021improved}
Nichol, A.~Q. and Dhariwal, P. (2021).
\newblock Improved denoising diffusion probabilistic models.
\newblock In {\em International Conference on Machine Learning}, pages 8162--8171. PMLR.

\bibitem[Peng et~al., 2019]{peng2019advantage}
Peng, X.~B., Kumar, A., Zhang, G., and Levine, S. (2019).
\newblock Advantage-weighted regression: Simple and scalable off-policy reinforcement learning.
\newblock {\em arXiv preprint arXiv:1910.00177}.

\bibitem[Rigter et~al., 2023]{rigter2023world}
Rigter, M., Yamada, J., and Posner, I. (2023).
\newblock World models via policy-guided trajectory diffusion.
\newblock {\em arXiv preprint arXiv:2312.08533}.

\bibitem[Robine et~al., 2023]{robine2023transformer}
Robine, J., H{\"o}ftmann, M., Uelwer, T., and Harmeling, S. (2023).
\newblock Transformer-based world models are happy with 100k interactions.
\newblock {\em arXiv preprint arXiv:2303.07109}.

\bibitem[Ronneberger et~al., 2015]{ronneberger2015u}
Ronneberger, O., Fischer, P., and Brox, T. (2015).
\newblock U-net: Convolutional networks for biomedical image segmentation.
\newblock In {\em Medical Image Computing and Computer-Assisted Intervention--MICCAI 2015: 18th International Conference, Munich, Germany, October 5-9, 2015, Proceedings, Part III 18}, pages 234--241. Springer.

\bibitem[Schrittwieser et~al., 2020]{schrittwieser2020mastering}
Schrittwieser, J., Antonoglou, I., Hubert, T., Simonyan, K., Sifre, L., Schmitt, S., Guez, A., Lockhart, E., Hassabis, D., Graepel, T., et~al. (2020).
\newblock Mastering atari, go, chess and shogi by planning with a learned model.
\newblock {\em Nature}, 588(7839):604--609.

\bibitem[Schulman et~al., 2015]{schulman2015high}
Schulman, J., Moritz, P., Levine, S., Jordan, M., and Abbeel, P. (2015).
\newblock High-dimensional continuous control using generalized advantage estimation.
\newblock {\em arXiv preprint arXiv:1506.02438}.

\bibitem[Silver et~al., 2014]{silver2014deterministic}
Silver, D., Lever, G., Heess, N., Degris, T., Wierstra, D., and Riedmiller, M. (2014).
\newblock Deterministic policy gradient algorithms.
\newblock In {\em International conference on machine learning}, pages 387--395. Pmlr.

\bibitem[Sohl-Dickstein et~al., 2015]{sohl2015deep}
Sohl-Dickstein, J., Weiss, E., Maheswaranathan, N., and Ganguli, S. (2015).
\newblock Deep unsupervised learning using nonequilibrium thermodynamics.
\newblock In {\em International conference on machine learning}, pages 2256--2265. PMLR.

\bibitem[Song et~al., 2023]{song2023consistency}
Song, Y., Dhariwal, P., Chen, M., and Sutskever, I. (2023).
\newblock Consistency models.
\newblock {\em arXiv preprint arXiv:2303.01469}.

\bibitem[Song et~al., 2020]{song2020score}
Song, Y., Sohl-Dickstein, J., Kingma, D.~P., Kumar, A., Ermon, S., and Poole, B. (2020).
\newblock Score-based generative modeling through stochastic differential equations.
\newblock {\em arXiv preprint arXiv:2011.13456}.

\bibitem[Sutton, 1990]{sutton1990integrated}
Sutton, R.~S. (1990).
\newblock Integrated architectures for learning, planning, and reacting based on approximating dynamic programming.
\newblock In {\em Machine learning proceedings 1990}, pages 216--224. Elsevier.

\bibitem[Sutton, 1991]{sutton1991dyna}
Sutton, R.~S. (1991).
\newblock Dyna, an integrated architecture for learning, planning, and reacting.
\newblock {\em ACM Sigart Bulletin}, 2(4):160--163.

\bibitem[Thrun and Schwartz, 2014]{thrun2014issues}
Thrun, S. and Schwartz, A. (2014).
\newblock Issues in using function approximation for reinforcement learning.
\newblock In {\em Proceedings of the 1993 connectionist models summer school}, pages 255--263. Psychology Press.

\bibitem[Vaswani et~al., 2017]{vaswani2017attention}
Vaswani, A., Shazeer, N., Parmar, N., Uszkoreit, J., Jones, L., Gomez, A.~N., Kaiser, {\L}., and Polosukhin, I. (2017).
\newblock Attention is all you need.
\newblock {\em Advances in neural information processing systems}, 30.

\bibitem[Wang et~al., 2019]{wang2019benchmarking}
Wang, T., Bao, X., Clavera, I., Hoang, J., Wen, Y., Langlois, E., Zhang, S., Zhang, G., Abbeel, P., and Ba, J. (2019).
\newblock Benchmarking model-based reinforcement learning.
\newblock {\em arXiv preprint arXiv:1907.02057}.

\bibitem[Wang et~al., 2022]{wang2022diffusion}
Wang, Z., Hunt, J.~J., and Zhou, M. (2022).
\newblock Diffusion policies as an expressive policy class for offline reinforcement learning.
\newblock {\em arXiv preprint arXiv:2208.06193}.

\bibitem[Watkins and Dayan, 1992]{watkins1992q}
Watkins, C.~J. and Dayan, P. (1992).
\newblock Q-learning.
\newblock {\em Machine learning}, 8:279--292.

\bibitem[Williams et~al., 2015]{williams2015model}
Williams, G., Aldrich, A., and Theodorou, E. (2015).
\newblock Model predictive path integral control using covariance variable importance sampling.
\newblock {\em arXiv preprint arXiv:1509.01149}.

\bibitem[Williams, 1992]{williams1992simple}
Williams, R.~J. (1992).
\newblock Simple statistical gradient-following algorithms for connectionist reinforcement learning.
\newblock {\em Machine learning}, 8:229--256.

\bibitem[Wu and He, 2018]{wu2018group}
Wu, Y. and He, K. (2018).
\newblock Group normalization.
\newblock In {\em Proceedings of the European conference on computer vision (ECCV)}, pages 3--19.

\bibitem[Wu et~al., 2019]{wu2019behavior}
Wu, Y., Tucker, G., and Nachum, O. (2019).
\newblock Behavior regularized offline reinforcement learning.
\newblock {\em arXiv preprint arXiv:1911.11361}.

\bibitem[Xiao et~al., 2019]{xiao2019learning}
Xiao, C., Wu, Y., Ma, C., Schuurmans, D., and M{\"u}ller, M. (2019).
\newblock Learning to combat compounding-error in model-based reinforcement learning.
\newblock {\em arXiv preprint arXiv:1912.11206}.

\bibitem[Xu et~al., 2023]{xu2023controllable}
Xu, Y., Li, N., Goel, A., Guo, Z., Yao, Z., Kasaei, H., Kasaei, M., and Li, Z. (2023).
\newblock Controllable video generation by learning the underlying dynamical system with neural ode.
\newblock {\em arXiv preprint arXiv:2303.05323}.

\bibitem[Yamagata et~al., 2023]{yamagata2023q}
Yamagata, T., Khalil, A., and Santos-Rodriguez, R. (2023).
\newblock Q-learning decision transformer: Leveraging dynamic programming for conditional sequence modelling in offline rl.
\newblock In {\em International Conference on Machine Learning}, pages 38989--39007. PMLR.

\bibitem[Yang et~al., 2023]{yang2023learning}
Yang, M., Du, Y., Ghasemipour, K., Tompson, J., Schuurmans, D., and Abbeel, P. (2023).
\newblock Learning interactive real-world simulators.
\newblock {\em arXiv preprint arXiv:2310.06114}.

\bibitem[Ye et~al., 2021]{ye2021mastering}
Ye, W., Liu, S., Kurutach, T., Abbeel, P., and Gao, Y. (2021).
\newblock Mastering atari games with limited data.
\newblock {\em Advances in Neural Information Processing Systems}, 34:25476--25488.

\bibitem[Yu et~al., 2020]{yu2020mopo}
Yu, T., Thomas, G., Yu, L., Ermon, S., Zou, J.~Y., Levine, S., Finn, C., and Ma, T. (2020).
\newblock Mopo: Model-based offline policy optimization.
\newblock {\em Advances in Neural Information Processing Systems}, 33:14129--14142.

\bibitem[Zhang et~al., 2023]{zhang2023learning}
Zhang, L., Xiong, Y., Yang, Z., Casas, S., Hu, R., and Urtasun, R. (2023).
\newblock Learning unsupervised world models for autonomous driving via discrete diffusion.
\newblock {\em arXiv preprint arXiv:2311.01017}.

\bibitem[Zheng et~al., 2023a]{zheng2023semi}
Zheng, Q., Henaff, M., Amos, B., and Grover, A. (2023a).
\newblock Semi-supervised offline reinforcement learning with action-free trajectories.
\newblock In {\em International Conference on Machine Learning}, pages 42339--42362. PMLR.

\bibitem[Zheng et~al., 2023b]{zheng2023guided}
Zheng, Q., Le, M., Shaul, N., Lipman, Y., Grover, A., and Chen, R.~T. (2023b).
\newblock Guided flows for generative modeling and decision making.
\newblock {\em arXiv preprint arXiv:2311.13443}.

\bibitem[Zheng et~al., 2022]{zheng2022online}
Zheng, Q., Zhang, A., and Grover, A. (2022).
\newblock Online decision transformer.
\newblock In {\em international conference on machine learning}, pages 27042--27059. PMLR.

\end{thebibliography}
\bibliographystyle{apalike}

\newpage
\appendix
\onecolumn

\section{Implementation Details of Diffusion World Model}
\label{subsec:app_diffusion_model}
We summarize the architecture and hyperparameters used for our experiments. 
For all the experiments, we use our own PyTorch implementation that is heavily influenced by the following codebases: 

Decision Diffuser~\cite{ajay2022conditional} \hskip5pt \url{https://github.com/anuragajay/decision-diffuser} \\
Diffuser~\cite{janner2022planning} \hskip5pt \url{https://github.com/jannerm/diffuser/}\\
SSORL~\cite{zheng2023semi} \hskip5pt \url{https://github.com/facebookresearch/ssorl/}


\paragraph{Architecture.}
As introduced in Section~\ref{sec:method_diffusion}, the diffusion world model $p_\theta$ used in this paper is chosen to model a length-$T$
subtrajecotires $(s_t, a_t, r_t, s_{t+1}, r_{t+1}, \dots, s_{t+T-1}, r_{t+T-1})$. At inference time,
it predicts the subsequent subtrajecotry of $T-1$ steps, conditioning on initial state-action pair $(s_t, a_t)$ and target RTG $y=g_t$:
\begin{align}
    \rhat_t, \shat_{t+1}, \rhat_{t+1}, \dots, \shat_{t+T-1}, \rhat _{t+T-1} \sim p_\theta(\cdot| s_t, a_t, y=g_t).
\end{align}
There are two reasons we choose not to model future actions in the sequence. First, our proposed diffusion model value expansion (Definition~\ref{def:dmve}) does not require the action information for future steps. Second, previous work have found that modeling continuous action through diffusion is less accurate~\cite{ajay2022conditional}.

Throughout the paper, we train guided diffusion models for state-reward sequences of length $T=8$. The number of diffusion steps is $K=5$.
The probability of null conditioning $p_\text{uncond}$ is set to $0.25$, and the batch size is $64$.
We use the cosine noise schedule proposed by \citet{nichol2021improved}.
The discount factor is $\gamma=0.99$, and we normalize the discounted RTG by a task-specific reward scale, which is $400$ for Hopper, $550$ for Walker, and $1200$ for Halfcheetah tasks.

Following~\citet{ajay2022conditional}, our noise predictor $\eps_\theta$ is a temporal U-net~\cite{janner2022planning, ronneberger2015u} that consists of 6 repeated residual blocks, where each block consists of 2 temporal convolutions followed by the group norm~\cite{wu2018group}
and a final Mish nonlinearity activation~\cite{mish2019self}.
The diffusion step $k$ is first transformed to its sinusoidal position encoding and projected to a latent space via a 2-layer MLP, and the RTG value is transformed into its latent embedding via a 3-layer MLP. In our diffusion world model, the initial action $a_t$ as additional condition is also transformed into latent embedding via a 3-layer MLP, and further concatenated with the embeddings of the diffusion step and RTG. 

\paragraph{Optimization.} We optimize our model by the Adam optimizer with a learning rate $1 \times 10^{-4}$ for all the datasets. The final model parameter $\bar{\theta}$ we consider is an exponential moving average (EMA) of the obtained parameters over the course of training. For every $10$ iteration, we update 
$\bar{\theta} = \beta \bar{\theta} + (1- \beta) \theta$,
where the exponential decay parameter $\beta = 0.995$. We train the diffusion model for $2 \times 10^6$ iterations.

\paragraph{Sampling with Guidance.}
\label{appendix:sampling}
To sample from the diffusion model, we need to first sample a random noise $x{(K)} \sim \N(0, \mathbf{I})$ and then run the reverse process.
Algorithm~\ref{algo:diffusion_sampling_general} presents the general process of sampling from a diffusion model trained under classifier-free guidance.

In the context of offline RL, the diffusion world model generates future states and rewards based on the current state $s_t$, the current action $a_t$ and the target return $g_\text{eval}$, see Section~\ref{sec:method}. Therefore, the sampling process is slightly different from Algorithm~\ref{algo:diffusion_sampling_general}, as we need to constrain the initial state and initial action to be $s_t$ and $a_t$, respectively. The adapted algorithm is summarized in Algorithm~\ref{algo:diffusion_sampling_rl}.

Following \citet{ajay2022conditional}, we apply the low temperature sampling technique for diffusion models. The temperature is set to be $\alpha=0.5$ for sampling at each
diffusion step from Gaussian $\mathcal{N}(\hat{\mu}_\theta, \alpha^2\hat{\Sigma}_\theta)$, with $\hat{\mu}_\theta$ and $\hat{\Sigma}_\theta$ being the predicted mean and covariance.

\paragraph{Accelerated Inference.}
\label{subsec:app_acc}
Algorithm~\ref{algo:diffusion_sampling_general} and \ref{algo:diffusion_sampling_rl} run the full reverse process, 
Building on top of them, we further apply the stride sampling technique as in~\citet{nichol2021improved} to speed up sampling process.
Formally, in the full reverse process, we generates $x^{(k-1)}$ by $x^{(k)}$ one by one, from $k=K$ till $k=1$:
\begin{equation}
    x^{(k-1)} = \frac{\sqrt{\bar{\alpha}_{k-1}}\beta_k}{1-\bar{\alpha}_k}\hat{x}^{(0)} + \frac{\sqrt{\alpha_k}(1-\bar{\alpha}_{k-1})}{1-\bar{\alpha}_k}x^{(k)}+ \sigma_k\varepsilon, \;\; \varepsilon\sim\mathcal{N}(0, \mathbf{I}),
    \label{eq:diffusion_reverse_computing}
\end{equation}
where $\hat{x}^{(0)}$ is the prediction of the true data point (line~\ref{algo:line_diffusion_sampling_estimate_x0} of Algorithm~\ref{algo:diffusion_sampling_general}),
$\sigma_k = \sqrt{\dfrac{\beta_k(1 - \bar{\alpha}_{k-1})}{1 - \bar{\alpha}_k}}$ is the standard deviation of noise at step $k$ (line 8 of in Algorithm~\ref{algo:diffusion_sampling_general}). We note that $\bar{\alpha}_k = \prod_{k'=1}^K (1 -\beta_{k'})$ where the noise schedule $\set{\beta_k}_{k=1}^K$ is predefined, see Section~\ref{sec:preliminary} and Appendix~\ref{subsec:app_diffusion_model}.

Running a full reverse process amounts to evaluating Equation~\eqref{eq:diffusion_reverse_computing} for $K$ times, which is time consuming. 
To speed up sampling, we choose $N$ diffusion steps equally spaced between $1$ and $K$, namely, $\tau_1, \ldots, \tau_N$, where $\tau_N=K$. We then evaluate Equation~\eqref{eq:diffusion_reverse_computing} for the chosen steps $\tau_1, \ldots, \tau_N$.
This effectively reduces the inference time to approximately $N/K$ of the original. In our experiments, we train the diffusion model with $K=5$ diffusion steps and sample with $N=3$ inference steps, see Section~\ref{sec:expr_ablation} for a justification of this number.

\begin{algorithm}[H]
\DontPrintSemicolon
\SetCommentSty{bluetcp}
\caption{Sampling from Guided Diffusion Models}
\label{algo:diffusion_sampling_general}
\textbf{Input:} trained noise prediction model $\eps_\theta$,
conditioning parameter $y$, guidance parameter $\omega$, number of diffusion steps $K$\; 
$ \x{K} \sim \N(0, \mathbf{I})$\;
\For{$k = K, \ldots, 1$}{
    $\hat{\eps} \leftarrow \omega \cdot \eps_\theta(\x{k} , k, y) + (1 -\omega) \cdot \eps_\theta(\x{k} , k, \varnothing)$\;
    \tcp{estimate true data point $\x{0}$}
    $\hat{x}^{(0)} \leftarrow \dfrac{1}{\sqrt{\bar{\alpha}_k}}\big(\x{k} - \sqrt{1-\bar{\alpha}_t}\hat{\eps}\big)$ \label{algo:line_diffusion_sampling_estimate_x0}\;
    \tcp{Sample from the posterior distribution $q(\x{k-1} | \x{k}, \x{0})$}
    \tcp{See Equation (6) and (7) of \citet{ho2020denoising}}
    $\hat{\mu} \leftarrow \dfrac{\sqrt{\bar{\alpha}_{k-1}} \beta_k}{1 - \bar{\alpha}_k} \hat{x}^{(0)}
    + \dfrac{\sqrt{\alpha_k}(1 - \bar{\alpha}_{k-1})}{1 - \bar{\alpha}_k}\x{k}$\;
    $\hat{\Sigma} \leftarrow \dfrac{\beta_k(1 - \bar{\alpha}_{k-1})}{1 - \bar{\alpha}_k}\mathbf{I}$\;
    $\x{k-1} \sim \N(\hat{\mu}, \hat{\Sigma})$
}
\textbf{Output:} $\x{0}$
\end{algorithm}

\begin{algorithm}[H]
\DontPrintSemicolon
\SetCommentSty{bluetcp}
\caption{Diffusion World Model Sampling}
\label{algo:diffusion_sampling_rl}
\textbf{Input:} 
trained noise prediction model $\eps_\theta$,
initial state $s_t$, initial action $a_t$, target return $g_\text{eval}$,
guidance parameter $\omega$,
number of diffusion steps $K$\;
$ \x{K} \sim \N(0, \mathbf{I})$\;
\tcp{apply conditioning of $s_t$ and $a_t$}
$\x{K}[0:\text{dim}(s_t) + \text{dim}(a_t)] \leftarrow \text{concatenate}(s_t, a_t)$\;
\For{$k = K, \ldots, 1$}{
    $\hat{\eps} \leftarrow \omega \cdot \eps_\theta(\x{k} , k, g_\text{eval}) + (1 -\omega) \cdot \eps_\theta(\x{k} , k, \varnothing)$\;
    \tcp{estimate true data point $\x{0}$}
    $\hat{x}^{(0)} \leftarrow \dfrac{1}{\sqrt{\bar{\alpha}_k}}\big(\x{k} - \sqrt{1-\bar{\alpha}_t}\hat{\eps}\big)$ \;
    \tcp{Sample from the posterior distribution $q(\x{k-1} | \x{k}, \x{0})$}
    \tcp{See Equation (6) and (7) of \citet{ho2020denoising}}
    $\hat{\mu} \leftarrow \dfrac{\sqrt{\bar{\alpha}_{k-1}} \beta_k}{1 - \bar{\alpha}_k} \hat{x}^{(0)} 
    + \dfrac{\sqrt{\alpha_k}(1 - \bar{\alpha}_{k-1})}{1 - \bar{\alpha}_k}\x{k}$\;
    $\hat{\Sigma} \leftarrow \dfrac{\beta_k(1 - \bar{\alpha}_{k-1})}{1 - \bar{\alpha}_k}\mathbf{I}$ \;
    $\x{k-1} \sim \N(\hat{\mu}, \hat{\Sigma})$\;
    \tcp{apply conditioning of $s_t$ and $a_t$}
    $\x{k-1}[0:\text{dim}(s_t) + \text{dim}(a_t)] \leftarrow \text{concatenate}(s_t, a_t)$\;
}
\textbf{Output:} $\x{0}$
\end{algorithm}

\section{Implementation Details of One-step Dynamics Model}
\label{sec:app_onestep}
The traditional one-step dynamics model $f_\theta(s_{t+1},r_t|s_t,a_t)$ is typically represented by a parameterized probability distribution over the state and reward spaces, and optimized through log-likelihood maximization of the single-step transitions:
\begin{align}
    \max_\theta \mathbb{E}_{(s_t,a_t,r_t,s_{t+1}) \sim \Doffline}\left[\log f_\theta(s_{t+1},r_t|s_t,a_t) \right],
\end{align}
where $(s_t,a_t,r_t,s_{t+1})$ is sampled from the offline data distribution $\Doffline$.
As in \citet{kidambi2020morel}, we model $f_\theta$ as a Gaussian distribution $\mathcal{N}(\mu_\theta, \Sigma_\theta)$, where the mean $\mu_\theta$ and the diagonal covariance matrix $\Sigma_\theta$ are parameterized by two 4-layer MLP neural networks with 256 hidden units per layer. We use the ReLU activation function for hidden layers. The final layer of $\Sigma_\theta$ is activated by a SoftPlus function to ensure validity. 
We train the dynamics models for $1\times 10^6$ iterations, using the Adam optimizer with learning rate $1\times 10^{-4}$.

\section{Diffusion World Model Based Offline RL Methods}
\label{sec:app_instantiations}
In Section~\ref{sec:expr}, we consider 3 instantiations of Algorithm~\ref{algo:diffusion_mbrl} where we integrate TD3+BC, IQL, Q-learning with pessimistic reward (PQL) into our framework. These algorithms are specifically designed for offline RL, with \textit{conservatism} notions defined on actions (TD3+BC), value function (IQL), and rewards (PQL) respectively. In the sequel, we refer to our instantiations as DWM-TD3BC, DWM-IQL and DWM-PQL. The detailed implementation of them will be introduced below.

\subsection{DWM-TD3BC: TD3+BC with Diffusion World Model}

Building on top of the TD3 algorithm~\cite{fujimoto2018addressing}, TD3+BC~\cite{fujimoto2021minimalist} employs explicit behavior cloning regularization to learn a deterministic policy. The algorithm works as follows.

The critic training follows the TD3 algorithm exactly. We learn two critic networks $Q_{\phi_1}$ and $Q_{\phi_2}$ as double Q-learning~\cite{fujimoto2018addressing} through TD learning.
The target value in TD learning for a transition $(s,a, r, s')$ is given by:
\begin{equation}
    y = r + \gamma \min_{i\in\{1,2\}}Q_{\phi_i}\big(s', a'=\text{Clip}(\pi_{\bar{\psi}}(s')+\eps, -C, C)\big),
    \label{eq:target_value_td3bc}
\end{equation}
where $\eps \sim \N(0, \sigma^2I)$ is a random noise, $\pi_{\bar{\psi}}$ is the target policy and Clip($\cdot$) is an operator that bounds the value of the action vector to be within $[-C, C]$ for each dimension. Both $Q_{\phi_1}$ and $Q_{\phi_2}$ will regress into the target value $y$ by minimizing the mean squared error (MSE), which amounts to solving the following problem: 
\begin{equation}
    \min_{\phi_i} \mathbb{E}_{(s,a,r,s')\sim \mathcal{D}_\text{offline}}\left[ \left( y(r,s')- Q_{\phi_i}(s,a) \right)^2\right], \; i\in\{1,2\}.
 \label{eq:q_loss_td3bc}
\end{equation}
 
For training the policy, TD3+BC optimizes the following regularized problem:
\begin{align}
   \max_\psi \E_{(s,a)\sim \Doffline } \left[ \lambda Q_{\phi_1}(s,\pi_\psi(s)) - \norm{a-\pi_\psi(s)}^2 \right],
   \label{eq:td3bc_policy}
\end{align}
which $\Doffline$ is the offline data distribution. Without the behavior cloning regularization term $\norm{a-\pi_\psi(s)}^2$, the above problem reduces to the objective corresponding to the deterministic policy gradient theorem~\cite{silver2014deterministic}. Note that $\pi$ is always trying to maximize one fixed proxy value function.

Both the updates of target policy $\pi_{\bar{\psi}}$ and the target critic networks $Q_{\bar{\phi}_i}$ are delayed in TD3+BC. The whole algorithm is summarized in Algorithm~\ref{algo:mbtd3bc}.

\begin{algorithm}[H]
\DontPrintSemicolon
\SetCommentSty{bluetcp}
\caption{DWM-TD3BC}\label{algo:mbtd3bc}
\textbf{Inputs:} offline dataset $\Doffline$, pretrained diffusion world model $p_\theta$,
simulation horizon $H$, conditioning RTG $\geval$, policy and target networks update frequency $n$, coefficient $\lambda$, parameters for action perturbation and clipping: $\sigma$, $C$\;

Initialize the actor and critic networks $\pi_\psi$, $Q_{\phi_1}$, $Q_{\phi_2}$\;
Initialize the weights of target networks $\bar{\psi} \leftarrow \psi$, $\bar{\phi}_1 \leftarrow \phi_1$, $\bar{\phi}_2 \leftarrow \phi_2$\;
\For{$i=1, 2, \ldots$ until convergence}{
    Sample state-action pair $(s_t, a_t)$ from $\Doffline$\;
    \tcp{diffuion model value expansion}
    Sample $\rhat_t, \shat_{t+1}, \rhat_{t+1}, \ldots, \shat_{t+T-1},\rhat_{t+T-1} \sim p_\theta(\cdot | s_t, a_t, \geval)$\;
    Sample $\eps \sim \N(0, \sigma^2 I)$\;
    $\ahat^\eps_{t+H} \leftarrow \text{Clip}(\pi_{\bar{\psi}}(\shat_{t+H}) + \eps, -C, C)$ \;
    Compute the target $Q$ value 
    $y = \sum_{h=0}^{H-1} \gamma^{h} \rhat_{t+h} + \gamma^H \min_{i \in \set{1, 2}}Q_{\bar{\phi}_i}(\shat_{t+H}, \ahat^\eps_{t+H}) $\;
    \tcp{update the critic}
    $\phi_1 \leftarrow \phi_1 - \eta \nabla_{\phi_1} \norm{Q_{\phi_1}(s_t, a_t) - y}^2_2$\;
    $\phi_2 \leftarrow \phi_2 - \eta \nabla_{\phi_2} \norm{Q_{\phi_2}(s_t, a_t) - y}^2_2$\;
    \tcp{update the actor (delayed) and the target networks}
    \If{$i$ mod $n$}{
        $\psi \leftarrow \psi  + \eta \nabla_\psi \left( \lambda Q_{\phi_1}\Big(s_t, \pi_\psi(s_t)\Big) - \norm{a_t - \pi_\psi(s_t)}^2 \right)$\tcp*{Update the actor network}
        $\bar{\phi}_1 \leftarrow \bar{\phi}_1 + w (\phi - \bar{\phi}_1) $ \;
        $\bar{\phi}_2 \leftarrow \bar{\phi}_2 + w (\phi - \bar{\phi}_2) $ \tcp*{Update the target networks}
        $\bar{\psi} \leftarrow \bar{\psi} + w (\psi - \bar{\psi}) $ \;
    }
}

\textbf{Output:} $\pi_\psi$
\end{algorithm}

\subsection{DWM-IQL: IQL with Diffusion World Model}
IQL~\cite{kostrikov2021offline} applies pessimistic value estimation on offline dataset. In addition to the double Q functions used in TD3+BC,
IQL leverages an additional state-value function $V_\xi(s)$,
which is estimated through expectile regression:
\begin{align}
   \min_\xi \E_{(s,a)\sim \Doffline} \left[ 
   L^\tau 
   \left( 
   \min_{i\in\{1,2\}} Q_{\bar{\phi}_i}(s,a)-V_\xi(s) 
   \right)
   \right],
   \label{eq:iql_v}
\end{align}
where $L^\tau(u)=|\tau - \mathbbm{1}_{u<0}|u^2$ with hyperparameter $\tau\in(0.5, 1)$,
As $\tau \rightarrow 1$, $V_\xi(s)$ is essentially estimating the maximum value of $Q(s,a)$. This can be viewed as implicitly performing the policy improvement step, without an explicit policy. Using a hyperparameter $\tau < 1$ regularizes the value estimation (of an implicit policy) and thus mitigates the overestimation issue of $Q$ function. The $Q$ function is updated also using Eq.~\eqref{eq:q_loss_td3bc} but with target $y=r+\gamma V_\xi(s')$. Finally, given the $Q$ and the $V$ functions, the policy is extracted by Advantage Weighted Regression~\cite{peng2019advantage}, i.e., solving
\begin{align}
    \max_\psi \mathbb{E}_{(s,a)\sim\Doffline}\left[\exp \left( \beta (Q_\phi(s,a)-V_\xi(s))\right) \log \pi_\psi(a|s) \right].
    \label{eq:iql_policy}
\end{align}
The update of the target critic networks $Q_{\bar{\phi}_i}$ are delayed in IQL. The whole algorithm is summarzied in Algorithm~\ref{algo:mbiql}.

\begin{algorithm}[htbp]
\DontPrintSemicolon
\SetCommentSty{bluetcp}
\caption{DWM-IQL}\label{algo:mbiql}
\textbf{Inputs:} offline dataset $\Doffline$, pretrained diffusion world model $p_\theta$,
simulation horizon $H$, conditioning RTG $\geval$, target network update frequency $n$, expectile loss parameter $\tau$\;

Initialize the actor, critic and value networks $\pi_\psi$, $Q_{\phi_1}$, $Q_{\phi_2}$, $V_\xi$\;
Initialize the weights of target networks $\bar{\phi}_1 \leftarrow \phi_1$, $\bar{\phi}_2 \leftarrow \phi_2$\;
\For{$i=1, 2, \ldots$ until convergence}{
    Sample state-action pair $(s_t, a_t)$ from $\Doffline$\;
    \tcp{diffuion model value expansion}
    Sample $\rhat_t, \shat_{t+1}, \rhat_{t+1}, \ldots, \shat_{t+T-1},\rhat_{t+T-1} \sim p_\theta(\cdot | s_t, a_t, \geval)$\;
    Compute the target $Q$ value 
    $y = \sum_{h=0}^{H-1} \gamma^{h} \rhat_{t+h} + \gamma^H V_\xi(\shat_{t+H}) $\;
    \tcp{update the $V$-value network}
    $\xi \leftarrow \xi - \eta \nabla_{\xi} L^\tau(\min_{i\in\{1,2\}}Q_{\bar{\phi}_i}(s,a)-V_\xi(s))$\;
    \tcp{update the critic ($Q$-value networks)}
    $\phi_1 \leftarrow \phi_1 - \eta \nabla_{\phi_1} \norm{Q_{\phi_1}(s_t, a_t) - y}^2_2$\;
    $\phi_2 \leftarrow \phi_2 - \eta \nabla_{\phi_2} \norm{Q_{\phi_2}(s_t, a_t) - y}^2_2$\;
    \tcp{update the actor}
    Update the actor network: $\psi \leftarrow \psi  + \eta \nabla_\psi \exp \big(\beta ( \min_{i \in \set{1,2} }Q_{\phi_i}(s,a)-V_\xi(s))\big) \log \pi_\psi(a|s)$\;
    \tcp{update the target networks}
    \If{$i$ mod $n$}{
        $\bar{\phi}_1 \leftarrow \bar{\phi}_1 + w (\phi - \bar{\phi}_1) $ \;
        $\bar{\phi}_2 \leftarrow \bar{\phi}_2 + w (\phi - \bar{\phi}_2) $ \;
    }
}

\textbf{Output:} $\pi_\psi$
\end{algorithm}

\subsection{DWM-PQL: Pessimistic Q-learning with Diffusion World Model}
Previous offline RL algorithms like MOPO~\cite{yu2020mopo} have applied the conservatism notion directly to the reward function, which we referred to as pessimistic Q-learning (PQL) in this paper. Specifically, the original algorithm proposed by \citet{yu2020mopo} learns an ensemble of $m$ one-step dynamics models $\{p_{\theta_i}\}_{i\in[m]}$, and use a modified reward
\begin{equation}
    \tilde{r}(s,a)=\hat{r}(s,a) - \kappa u(s, a|p_{\theta_1}, \ldots, p_{\theta_m})
\end{equation}
for learning the $Q$ functions, where $\hat{r}$ is the mean prediction from the ensemble, $u(s, a|p_{\theta_1}, \ldots, p_{\theta_m})$ is a measurement of prediction uncertainty using the ensemble. 

\citet{yu2020mopo} parameterize each dynamics model by a Gaussian distribution, and measure the prediction uncertainty using the maximum of the Frobenious norm of each covariance matrix. Since the diffusion model does not have such parameterization, and it is computationally daunting to train an ensemble of diffusion models, we propose an alternative uncertainty measurement similar to the one used in MoRel~\cite{kidambi2020morel}.

Given $(s_t, a_t)$ and $\geval$, we randomly sample $m$ sequences from the DWM, namely,
\begin{equation}
    \rhat^i_t, \shat^i_{t+1}, \rhat^i_{t+1}, \ldots, \shat^i_{t+T-1},\rhat^i_{t+T-1}, \;\; i \in [m].
\end{equation}
Then, we take the 1st sample as the DWM output with modified reward:
\begin{equation}
\label{eq:morel_uncertainty}
\tilde{r}_{t'} = \sum_{i=1}^m\frac{1}{m}\hat{r}_{t'}^i - \kappa \max_{i\in[m], j \in [m]} \left( \norm{\rhat^i_{t'} - \rhat^j_{t'} }^2 + \norm{\shat^i_{t'+1} - \shat^j_{t'+1} }^2_2 \right), \;\; t' = t, \ldots, t+T-2.
 \end{equation}   
This provides an efficient way to construct uncertainty-penalized rewards for each timestep along the diffusion predicted trajectories. Note that this does not apply to the reward predicted for the last timestep.
The rest of the algorithm follows IQL but using MSE loss instead of expectile loss for updating the value network. 

The DWM-PQL algorithm is summarized in Algorithm~\ref{algo:pqld}

\begin{algorithm}[htbp]
\DontPrintSemicolon
\SetCommentSty{bluetcp}
\caption{DWM-PQL}\label{algo:pqld}
\textbf{Inputs:} offline dataset $\Doffline$, pretrained diffusion world model $p_\theta$,
simulation horizon $H$, conditioning RTG $\geval$, target network update frequency $n$, pessimism coefficient $\lambda$, number of samples for uncertainty estimation $m$ \;

Initialize the actor, critic and value networks $\pi_\psi$, $Q_{\phi_1}$, $Q_{\phi_2}$, $V_\xi$\;
Initialize the weights of target networks $\bar{\phi}_1 \leftarrow \phi_1$, $\bar{\phi}_2 \leftarrow \phi_2$\;
\For{$i=1, 2, \ldots$ until convergence}{
    Sample state-action pair $(s_t, a_t)$ from $\Doffline$\;
    \tcp{diffuion model value expansion}
    Sample $m$ subtrajectories $\{\rhat^j_t, \shat^j_{t+1}, \rhat^j_{t+1}, \ldots, \shat^j_{t+T-1},\rhat^j_{t+T-1}\}_{j=1}^m \sim p_\theta(\cdot | s_t, a_t, \geval)$\;
    Modify the rewards of the first subtrajectory as in Eq.~\eqref{eq:morel_uncertainty}:  $\tilde{r}_t, \shat_{t+1}, 
    \tilde{r}_{t+1}, \ldots, \shat_{t+T-1},\rhat_{t+T-1}$\;
    Compute the target $Q$ value 
    $y = \sum_{h=0}^{H-1} \gamma^{h} \tilde{r}_{t+h} + \gamma^H V_\xi(\shat^1_{t+H}) $\;
    \tcp{update the $V$-value network}
    $\xi \leftarrow \xi - \eta \nabla_{\xi} ||\min_{i\in\{1,2\}}Q_{\bar{\phi}_i}(s,a)-V_\xi(s))||_2^2$\;
    \tcp{update the critic ($Q$-value networks)}
    $\phi_1 \leftarrow \phi_1 - \eta \nabla_{\phi_1} \norm{Q_{\phi_1}(s_t, a_t) - y}^2_2$\;
    $\phi_2 \leftarrow \phi_2 - \eta \nabla_{\phi_2} \norm{Q_{\phi_2}(s_t, a_t) - y}^2_2$\;
    \tcp{update the actor}
    Update the actor network: $\psi \leftarrow \psi  + \eta \nabla_\psi \exp \big(\beta (\min_{i \in \set{1,2} }Q_{\phi_i}(s,a)-V_\xi(s))\big) \log \pi_\psi(a|s)$\;
    \tcp{update the target networks}
    \If{$i$ mod $n$}{
        $\bar{\phi}_1 \leftarrow \bar{\phi}_1 + w (\phi - \bar{\phi}_1) $ \;
        $\bar{\phi}_2 \leftarrow \bar{\phi}_2 + w (\phi - \bar{\phi}_2) $ \;
    }
}
\textbf{Output:} $\pi_\psi$
\end{algorithm}

For baseline methods with one-step dynamics model, the imagined trajectories starting from sample $(s_t, a_t)\sim \Doffline$ are derived by recursively sample from the one-step dynamics model $f_\theta(\cdot|s,a)$ and policy $\pi_\psi(\cdot|s)$: $\hat{\tau}(s_t, a_t) \sim (f_\theta \circ \pi_\psi)^{H-1}(s_t, a_t)$. By keeping the rest same as above, it produces MBRL methods with one-step dynamics, namely O-IQL, O-TD3BC and O-PQL.

\section{Training and Evaluation Details of Offline RL Algorithms}
\label{subsec:app_mf_mb_details}
\subsection{Common Settings}
We conduct primary tests on TD3+BC and IQL for selecting the best practices for data normalization. Based on the results,
TD3+BC, O-TD3BC and DWM-TD3BC applies observation normalization, while other algorithms (O-PQL, DWM-PQL, IQL, O-IQL and DWM-IQL) applies both observation and reward normalization.

All models are trained on NVIDIA Tesla V100 PCle GPU devices. For training 200 epochs (1000 iterations per epoch), model-free algorithms like TD3+BC and IQL typically takes around 2000 seconds, DWM model-based algorithms like DWM-TD3BC and DWM-IQL typically takes around 18000 seconds, transformer model-based algorithms like T-TD3BC and T-IQL typically takes around  8000 seconds, one-step model-based algorithms like O-TD3BC and O-IQL typically takes around 2300 seconds. The specific computational time varies from task to task.

All algorithms are trained with a batch size of 128 using a fixed set of pretrained dynamics models (one-step and diffusion). The discount factor is set as $\gamma=0.99$ for all data.

\subsection{MF Algorithms}
TD3+BC and IQL are trained for $1\times 10^6$ iterations, with learning rate $3\times 10^{-4}$ for actor, critic and value networks. The actor, critic, and value networks are all parameterized by 3-layer MLPs with 256 hidden units per layer. We use the ReLU activation function for each hidden layer. IQL learns a stochastic policy which outputs a Tanh-Normal distribution, while TD3+BC has a deterministic policy with Tanh output activation. The hyperparameters for TD3+BC and IQL are provided in Table~\ref{tab:mf_hyperparam}. \looseness=-1

\begin{table}[h]
\centering
\begin{tabular}{cc|cc}
\toprule
\multicolumn{2}{c|}{TD3+BC} & \multicolumn{2}{c}{IQL}  \\ \hline
policy noise & 0.2 & expectile & 0.7  \\
noise clip & 0.5 & $\beta$ & 3.0\\
policy update frequency & 2 & max weight & 100.0 \\
target update frequence & 2 & policy update frequence & 1 \\
$\alpha$ & 2.5 & advantage normalization & False\\ 
EMA $w$ & 0.005 & EMA $w$ & 0.005  \\
\bottomrule
\end{tabular}
 \caption{Hyperparameters for training TD3+BC and IQL.}
\label{tab:mf_hyperparam}
\end{table}

The baseline DD~\cite{ajay2022conditional} algorithm uses diffusion models trained with sequence length $T=32$ and number of diffusion steps $K=5$.
It requires additionally training an inverse dynamics model (IDM) for action prediction, which is parameterized by a 3-layer MLP with 1024 hidden units for each hidden layer and ReLU activation function. The dropout rate for the MLP is 0.1. The IDMs are trained for $2\times 10^6$ iterations for each environment.
For a fair comparison with the other DWM methods, DD uses $N=3$ internal sampling steps as DWM. We search over the same range of evaluation RTG $g_\text{eval}$ for DD and the other DWM methods.

\subsection{MB algorithms}
DWM-TD3BC, DWM-IQL and DWM-PQL are trained for $5\times 10^5$ iterations. Table~\ref{tab:hyperparam} summarizes the hyperparameters we search for each experiment. The other hyperparameters and network architectures are the same as original TD3+BC and IQL in above sections. DWM-IQL with $\lambda$-return takes $\lambda=0.95$, following \citet{hafner2023mastering}.

The counterparts with one-step dynamics models are trained for $2\times 10^5$ iterations due to a relatively fast convergence from our empirical observation. Most of the hyperparameters also follow TD3+BC and IQL. The PQL-type algorithms (O-PQL and DWM-PQL) further search the pessimism coefficient $\kappa$ (defined in Eq.~\eqref{eq:morel_uncertainty}) among $\{0.01, 0.1, 1.0\}$.

\begin{table}[H]
\centering
\begin{tabular}{cccc}
\toprule
Env & Evaluation RTG &$H$ \\ \hline
hopper-medium-v2 & [0.6, 0.7, 0.8]  & [1,3,5,7]\\
walker2d-medium-v2 &  [0.6, 0.7, 0.8]  & [1,3,5,7] \\
halfcheetah-medium-v2 &  [0.4, 0.5, 0.6]   & [1,3,5,7]\\
hopper-medium-replay-v2 & [0.6, 0.7, 0.8]  & [1,3,5,7]\\
walker2d-medium-replay-v2 & [0.6, 0.7, 0.8]  & [1,3,5,7]\\
halfcheetah-medium-replay-v2 &  [0.4, 0.5, 0.6]  & [1,3,5,7]\\
hopper-medium-expert-v2 & [0.7, 0.8, 0.9]  & [1,3,5,7] \\
walker2d-medium-expert-v2 & [0.8, 0.9, 1.0]   & [1,3,5,7]\\
halfcheetah-medium-expert-v2 &  [0.6, 0.7, 0.8] & [1,3,5,7] \\
\bottomrule
\end{tabular}
 \caption{List of the hyperparameters we search for DWM methods.}
\label{tab:hyperparam}
\end{table}

\section{Additional Experiments}
\subsection{Detailed Results of Long Horizon Planning with DWM}
\label{subsec:app_rollout_length}
The section provides the detailed results of the experiments for long horizon planning with DWM in Section~\ref{sec:expr_dwm_vs_onestep}. Table~\ref{tab:rollout_length} summarizes the normalized returns (with means and standard deviations) of DWM-IQL and DWM-TD3BC for different simulation horizons $\{1,3,7,15, 31\}$. 

\begin{table}[H]
\centering
\begin{tabular}{cccc}
\toprule
& & \multicolumn{2}{c}{Return (mean$\pm$std)} \\ \cline{3-4}
Env. & Simulation Horizon & DWM-IQL & DWM-TD3BC \\ \hline
\multirow{5}{*}{\textbf{hopper-medium-v2}} 
& 1 & 0.54 ± 0.11 & 0.68 ± 0.12\\
& 3 & 0.55 ± 0.10  & 0.63 ± 0.11\\
& 7 & 0.56 ± 0.09  & 0.66 ± 0.13 \\
& 15 & 0.58 ± 0.12  & 0.77 ± 0.15\\
& 31 & 0.61 ± 0.11  & 0.79 ± 0.15\\ \hline
\multirow{5}{*}{\textbf{walker2d-medium-v2}} 
& 1 & 0.65 ± 0.23 & 0.56 ± 0.13 \\
& 3 & 0.74 ± 0.11  & 0.74 ± 0.13 \\
& 7 & 0.71 ± 0.13  & 0.74 ± 0.11 \\
& 15 & 0.66 ± 0.15 & 0.73 ± 0.13\\
& 31 & 0.67 ± 0.20 & 0.75 ± 0.12\\\hline
\multirow{5}{*}{\textbf{halfcheetah-medium-v2}} 
& 1 & 0.44 ± 0.01 & 0.35 ± 0.03 \\
& 3 & 0.44 ± 0.01  & 0.39 ± 0.01 \\
& 7 & 0.44 ± 0.01  & 0.40 ± 0.01\\
& 15 & 0.44 ± 0.02   & 0.40 ± 0.01 \\
& 31 & 0.44 ± 0.01 & 0.40 ± 0.01 \\\hline
\multirow{5}{*}{\textbf{hopper-medium-replay-v2}} 
& 1 & 0.18 ± 0.06 & 0.52 ± 0.21 \\
& 3 & 0.37 ± 0.18  & 0.44 ± 0.23 \\
& 7 & 0.39 ± 0.14  & 0.52 ± 0.28 \\
& 15 & 0.37 ± 0.18   & 0.67 ± 0.25 \\
& 31 & 0.37 ± 0.15 & 0.59 ± 0.22 \\\hline
\multirow{5}{*}{\textbf{walker2d-medium-replay-v2}} 
& 1 & 0.32 ± 0.15 & 0.13 ± 0.02 \\
& 3 & 0.27 ± 0.24  & 0.19 ± 0.10 \\
& 7 & 0.25 ± 0.20  & 0.22 ± 0.14 \\
& 15 & 0.26 ± 0.19   & 0.22 ± 0.10 \\
& 31 & 0.27 ± 0.19 & 0.17 ± 0.12  \\\hline
\multirow{5}{*}{\textbf{halfcheetah-medium-replay-v2}}  
& 1 & 0.38 ± 0.05 & 0.02 ± 0.00\\
& 3 & 0.39 ± 0.02  & 0.17 ± 0.05  \\
& 7 & 0.39 ± 0.02  & 0.22 ± 0.03 \\
& 15 & 0.38 ± 0.03   & 0.26 ± 0.03\\
& 31 & 0.37 ± 0.03 & 0.26 ± 0.05 \\\hline
\multirow{5}{*}{\textbf{hopper-medium-expert-v2}} 
& 1 & 0.86 ± 0.25 & 0.88 ± 0.17\\
& 3 & 0.90 ± 0.19 & 0.94 ± 0.22 \\
& 7 & 0.88 ± 0.28  & 0.93 ± 0.24\\
& 15 & 0.85 ± 0.20  & 0.91 ± 0.19 \\
& 31 & 0.84 ± 0.23 & 0.93 ± 0.23 \\\hline
\multirow{5}{*}{\textbf{walker2d-medium-expert-v2}} 
& 1 & 0.80 ± 0.22 &  0.74 ± 0.21\\
& 3 & 1.02 ± 0.09  & 0.89 ± 0.13 \\
& 7 & 0.98 ± 0.2  & 0.82 ± 0.19  \\
& 15 & 1.06 ± 0.05 & 0.84 ± 0.14 \\
& 31 & 1.05 ± 0.06 & 0.87 ± 0.03 \\\hline
\multirow{5}{*}{\textbf{halfcheetah-medium-expert-v2}} 
& 1 & 0.60 ± 0.18 & 0.39 ± 0.01\\
& 3 & 0.52 ± 0.14  & 0.43 ± 0.07 \\
& 7 & 0.63 ± 0.13  & 0.44 ± 0.03\\
& 15 & 0.66 ± 0.14   & 0.50 ± 0.08\\
& 31 & 0.65 ± 0.17 & 0.49 ± 0.09 \\
\bottomrule
\end{tabular}
 \caption{Comparison of the normalized returns with different simulation horizons for DWM-TD3BC and DWM-IQL. The reported values are the best performances across different RTG values (listed in Table~\ref{tab:hyperparam}).}
\label{tab:rollout_length}
\end{table}

\subsection{World Modeling: Prediction Error Analysis}
\label{subsec:compounding}
An additional experiment is conducted to evaluate the prediction errors of the observations and rewards with DWM under simulation horizon $H=8$, for an example task \textit{walker2d-medium-expert-v2}.

We randomly sample a subsequence $(s_t,\dots, s_{t+7})$ from the offline dataset, and let DWM predict the subsequent states and rewards, conditioned on the first state $s_t$ and true action $a_t$.
For the one-step model, the model iteratively predicts the reward $r_t$ and next state $s_{t+1}$, conditioned on the current state $s_t$ (which is predicted in the previous step) and true action $a_t$.

We report the mean squared error (MSE) between the predicted samples and the  ground truth for each rollout timestep in Table ~\ref{tab:obs_rew_compounding}. Each method is evaluated with five models and 100 sequences per model, and the mean and standard deviations are reported.  The average prediction errors over the entire sequence are also calculated in the last row.  It shows the significant reduction of prediction errors in sequence modeling by using DWM over traditional one-step models, especially when the prediction timestep is large.

\begin{table}[H]
\centering
\resizebox{\textwidth}{!}{
\begin{tabular}{c||c|c||c|c}
\hline
\multirow{2}{*}{\textbf{Step}} & \multicolumn{2}{c||}{\textbf{One-step Model}} & \multicolumn{2}{c}{\textbf{DWM}} \\
\cline{2-5}
 & \textbf{Observation} & \textbf{Reward} & \textbf{Observation} & \textbf{Reward} \\
\hline
1 & $0.0000 \pm 0.0000$ & $8.05e-05 \pm 0.00011$ & $0.0000 \pm 0.0000$ & $8.89e-05 \pm 0.00018$ \\
2 & $0.0363 \pm 0.0455$ & $2.03e-04 \pm 0.00031$ & $0.1050 \pm 0.1668$ & $1.50e-04 \pm 0.00020$ \\
3 & $0.1576 \pm 0.2308$ & $4.73e-04 \pm 0.00084$ & $0.4173 \pm 0.3902$ & $2.34e-04 \pm 0.00021$ \\
4 & $0.3503 \pm 0.3547$ & $7.68e-04 \pm 0.00148$ & $0.4525 \pm 0.5021$ & $2.33e-04 \pm 0.00018$ \\
5 & $0.6173 \pm 0.4945$ & $1.13e-03 \pm 0.00222$ & $0.4796 \pm 0.5035$ & $2.71e-04 \pm 0.00039$ \\
6 & $0.9185 \pm 0.8678$ & $2.00e-03 \pm 0.00417$ & $0.4854 \pm 0.4759$ & $4.01e-04 \pm 0.00102$ \\
7 & $1.2394 \pm 0.9788$ & $3.40e-03 \pm 0.00757$ & $0.5353 \pm 0.5076$ & $3.89e-04 \pm 0.00094$ \\
8 & $1.4890 \pm 0.9612$ & $4.47e-03 \pm 0.00844$ & $0.5146 \pm 0.5814$ & $3.77e-04 \pm 0.00064$ \\
\hline
\textbf{Average} & $0.6010 \pm 0.4916$ & $0.0015 \pm 0.0031$ & $0.3737 \pm 0.3984$ & $0.0003 \pm 0.0004$ \\
\hline
\end{tabular}
}
\caption{Comparison of observation and prediction error (MSE) over rollout timesteps ($H=8$) for one-step model and DWM for \textit{walker2d-medium-expert-v2}.}
\label{tab:obs_rew_compounding}
\end{table}

\subsection{Algorithm~\ref{algo:diffusion_mbrl} with Transformer-based World Model}
\label{subsec:app_transformer}
Following the same protocol as DWM, the Transformer model is trained to predict future state-reward sequences, conditioning on the initial state-action pair. We use a 4-layer transformer architecture with 4 attention heads, similar to the one in \citet{zheng2022online}. Specifically, all the actions except for the first one are masked out as zeros in the state-action-reward sequences.
Distinct from the original DT~\cite{chen2021decision} where the loss function only contains the action prediction error, here the Transformer is trained with state and reward prediction loss. The Transformers are trained with optimizers and hyperparameters following ODT~\cite{zheng2022online}.
The evaluation RTG for Transformers takes values $3600/400=9.0, 5000/550\approx 9.1, 6000/1200=5.0$ for hopper, walker2d and halfcheetah environments, respectively. The complete results of T-TD3BC and T-IQL are provided in Table~\ref{tab:dt_td3bc} and Table~\ref{tab:dt_iql} respectively.

\begin{table}[H]
\centering
\small
\begin{tabular}{c|cccc}
\toprule
& \multicolumn{4}{c}{Simulation Horizon} \\
Env & 1 & 3 & 5 & 7   \\  \hline
hopper-m  & 0.50$\pm$0.05 & 0.57$\pm$0.08 & 0.58$\pm$0.08 & 0.57$\pm$0.08 \\
walker2d-m & 0.36$\pm$0.15 &  0.40$\pm$0.20 & 0.60$\pm$0.16 & 0.53$\pm$0.17  \\
halfcheetah-m & 0.18$\pm$0.07 &  0.41 $\pm$0.03 & 0.38$\pm$0.08  &  0.42$\pm$0.03   \\
hopper-mr &   0.24$\pm$0.01 & 0.23$\pm$0.05 & 0.25$\pm$0.06 & 0.22$\pm$0.08  \\ 
walker2d-mr &  0.12 $\pm$0.04 & 0.09$\pm$0.05 & 0.13$\pm$0.06 & 0.12$\pm$0.05 \\
halfcheetah-mr &  0.40$\pm$0.01 & 0.39$\pm$0.02 & 0.39$\pm$0.03 & 0.39$\pm$0.02  \\
hopper-me &  0.41$\pm$0.13 & 0.57$\pm$0.19 & 0.66$\pm$0.25 & 0.52$\pm$0.15 \\
walker2d-me & 0.34$\pm$0.22 & 0.58$\pm$0.15 & 0.58$\pm$0.26 & 0.46$\pm$0.37  \\
halfcheetah-me &  0.14$\pm$0.06 & 0.31$\pm$0.09 & 0.36$\pm$0.17 & 0.29$\pm$0.12 \\ 
\bottomrule
\end{tabular}
\caption{The normalized returns of T-TD3BC.}
\label{tab:dt_td3bc}
\end{table}

\begin{table}[H]
\small
\centering
\begin{tabular}{c|cccc}
\toprule
& \multicolumn{4}{c}{Simulation Horizon} \\
Env & 1 & 3 & 5 & 7   \\  \hline
hopper-m  & 0.48$\pm$0.08 & 0.54$\pm$0.10 & 0.55$\pm$0.08 & 0.51$\pm$0.09 \\
walker2d-m & 0.54$\pm$0.18 &  0.62$\pm$0.19 & 0.72$\pm$0.12 & 0.72$\pm$0.14  \\
halfcheetah-m & 0.42$\pm$0.03 & 0.42$\pm$0.02 & 0.43$\pm$0.01 & 0.43$\pm$0.01     \\
hopper-mr & 0.17$\pm$0.05 & 0.24$\pm$0.09 & 0.26$\pm$0.09 & 0.20$\pm$0.07\\
walker2d-mr & 0.17$\pm$0.12 & 0.17$\pm$0.14 & 0.23$\pm$0.12 & 0.16$\pm$0.11\\
halfcheetah-mr & 0.38$\pm$0.04 & 0.39$\pm$0.01 & 0.38$\pm$0.04 & 0.39$\pm$ 0.03\\
hopper-me &  0.62$\pm$0.16 & 0.59$\pm$0.21 & 0.47$\pm$0.21 &  0.47$\pm$0.21 \\
walker2d-me & 0.67$\pm$0.23 & 0.87$\pm$0.21 & 1.03$\pm$0.09 & 0.71$\pm$0.22 \\
halfcheetah-me & 0.39$\pm$0.19 & 0.43$\pm$0.13 & 0.44$\pm$0.08 &  0.43$\pm$0.09 \\
\bottomrule
\end{tabular}
\caption{The normalized returns of T-IQL.}
\label{tab:dt_iql}
\end{table}

\subsection{Additional Baselines}
\label{subsec:add_baselines}
\subsubsection{Data Augmentation}
\label{subsec:data_aug}
As a data augmentation (DA) method, SynTHER~\cite{lu2023synthetic} learns an unconditional diffusion model at the transition level, where the generated data are used for augmenting the training distribution. We conduct additional experiments for DWM and SynTHER-type data augmentation for this section.

The two data augmentation methods based on TD3+BC and IQL are referred to as DA-TD3BC and DA-IQL. For the sake of fair comparison, for DA-TD3BC and DA-IQL, the models are trained in the same manner as DWM with the same parameter sweeping for RTG values and horizons. The results show that the DWM consistently performs better than the DA algorithms across all tasks, for both TD3+BC and IQL.

For the sake of fair comparison with DWM, we use the same data preprocessing as DWM for this experiment, which is different from state reward normalizations as the original SynTHER paper. 
In our experiments, we use CDF normalizers to transform each dimension of the state vector to $[-1, 1]$ independently, \emph{i.e.}, making the data uniform over each dimension by transforming with marginal CDFs. Specifically, 
we transform the raw reward $r_\text{raw}$ to $r = 2(r_\text{raw}-r_\text{min})/(r_\text{max}-r_\text{min})-1$, where $r_\text{min}$ and $r_\text{max}$
are max and min raw reward of the offline dataset.
SynTHER applies ``whitening" that makes each (non-terminal) continuous dimension mean 0 and std 1, and the terminal states are rounded without normalization. To enable fast sampling, we use a very low diffusion steps: 5 at training and 3 at testing. The original SynthER paper uses 128 steps, which requires more computational time for sample generation. We use a set of consistent parameters like model sizes and batch sizes cross all the environments, the same as our previous experiments.

\begin{table}[ht]
\centering
\begin{tabular}{c|cc|cc}
\toprule
Env. & DA-TD3BC & DWM-TD3BC & DA-IQL & DWM-IQL \\
\midrule
hopper-m & 0.65 $\pm$ 0.10 & \textbf{0.65 $\pm$ 0.10} & 0.51 $\pm$ 0.10 & 0.54 $\pm$ 0.11 \\
walker2d-m & 0.63 $\pm$ 0.18 & \textbf{0.70 $\pm$ 0.15} & 0.74 $\pm$ 0.09 & 0.76 $\pm$ 0.05 \\
halfcheetah-m & 0.44 $\pm$ 0.01 & \textbf{0.46 $\pm$ 0.01} & 0.44 $\pm$ 0.01 & 0.44 $\pm$ 0.01 \\
hopper-mr & 0.53 $\pm$ 0.09 & 0.53 $\pm$ 0.09 & 0.25 $\pm$ 0.04 & \textbf{0.61 $\pm$ 0.13} \\
walker2d-mr & 0.37 $\pm$ 0.22 & \textbf{0.46 $\pm$ 0.19} & 0.42 $\pm$ 0.24 & 0.35 $\pm$ 0.14 \\
halfcheetah-mr & \textbf{0.43 $\pm$ 0.01} & \textbf{0.43 $\pm$ 0.01} & 0.42 $\pm$ 0.04 & 0.41 $\pm$ 0.01 \\
hopper-me & \textbf{1.03 $\pm$ 0.14} & \textbf{1.03 $\pm$ 0.14} & 0.55 $\pm$ 0.19 & 0.90 $\pm$ 0.25 \\
walker2d-me & 1.09 $\pm$ 0.04 & \textbf{1.10 $\pm$ 0.00} & 0.76 $\pm$ 0.13 & 1.04 $\pm$ 0.10 \\
halfcheetah-me & 0.72 $\pm$ 0.14 & \textbf{0.75 $\pm$ 0.16} & 0.62 $\pm$ 0.14 & 0.71 $\pm$ 0.14 \\
\midrule
Average & 0.654 $\pm$ 0.103 & \textbf{0.679 $\pm$ 0.098} & 0.523 $\pm$ 0.109 & 0.641 $\pm$ 0.117 \\
\bottomrule
\end{tabular}
\caption{\small Comparison of our DWM method and data-augmentation (DA) methods on the D4RL dataset. Results are aggregated over 5 random seeds. }
\label{tab:data_aug_compare}
\end{table}

\subsubsection{Autoregressive Diffusion}
We conduct experiments on Autoregressive Diffusion (AD) mentioned in previous work~\cite{rigter2023world} as an additional Baseline. AD is essentially a one-step model using diffusion instead of MLP, which can be autoregressively rolled out and used for Model Based Value Expansion (MVE). We find that AD is computationally inefficient to be practically integrated into the MVE framework. We have checked the wallclock time (in seconds) for sampling a batch of 128 future trajectories with different values of horizon, for the \textit{walker2d-medium-v2} environment.

We set the number of trajectories to be 128 because this is the batch size we use for training RL agents. Results are averaged over 100 trials. For both approaches we use diffusion models of the same model size, where the number of sampling diffusion steps is 3 (to enable fast inference). This experiment is conducted on a A6000 GPU and time unit is second. 

The results are displayed in Table~\ref{tab:compare_ad}. The sampling time of DWM is a constant because it's a sequence model, and in practice we diffuse the whole sequence and take a part of it according to $H$; while the sampling time of AD scale linearly as $H$ increases. When $H=7$, the sampling time is roughly 
$6.67\times$ compared with DWM. The MB methods we reported in the paper are trained for $5\times 10^5$ iterations (see Section D.3). That means, even only generating trajectories will take 27.5 hours if we use AD (as opposed to $\sim4$ hours for DWM).This suggests that AD is too computationally expensive to be incorporated into the MBRL framework.

\begin{table}[htbp]
\centering
\begin{tabular}{l|c|c|c|c}
\hline
\textbf{Method} & \textbf{H=1} & \textbf{H=3} & \textbf{H=5} & \textbf{H=7} \\ \hline
DWM (trained with $T=8$) & 0.031 & 0.031 & 0.031 & 0.031 \\ \hline
Autoregressive Diffusion & 0.030 & 0.087 & 0.145 & 0.198 \\ \hline
\end{tabular}
\caption{Comparison AD and DWM for sampling time (seconds) under different horizon $H$ values}
\label{tab:compare_ad}
\end{table}

\subsection{Additional Environments: Sparse-reward Tasks}
\label{sec:add_sparse_reward}
To further verify the method on various tasks, we conduct experiments on the maze-type tasks with sparse rewards. The methods include TD3+BC, IQL, the DWM counterparts and DD. The training and evaluation protocol follow exactly the same as the main experiments in Appendix~\ref{subsec:app_mf_mb_details}. The results are summarized in Table~\ref{tab:maze}, which shows the superior performance of DWM-based algorithms in sparse-reward settings.

\begin{table}[ht]
\centering
\begin{tabular}{c|ccccc}
\toprule
Env. & DWM-TD3BC & DWM-IQL & TD3+BC & IQL & DD\\
\midrule
maze2d-umaze   & $0.36\pm0.23$ & $0.39\pm0.29$ & $0.05\pm0.15$ & $0.08\pm0.16$ & $0.40\pm0.52$ \\ \hline
maze2d-medium  & $0.57\pm0.50$ & $0.41\pm0.09$ & $0.00\pm0.006$ & $0.10\pm0.10$ & $0.20\pm0.19$ \\ \hline
maze2d-large   & $0.28\pm0.13$ & $0.11\pm0.13$ & $-0.01\pm0.03$ & $0.01\pm0.07$ & $0.02\pm0.07$ \\ \hline
antmaze-umaze  & $0.86\pm0.29$ & $0.66\pm0.47$ & $0.58\pm0.49$ & $0.64\pm0.48$ & $0.31\pm0.45$ \\
\bottomrule
\end{tabular}
\caption{\small Comparison of different methods on sparse-reward tasks: three \textit{maze2d} tasks and one \textit{antmaze} task.}
\label{tab:maze}
\end{table}

\subsection{Ablation: Number of Diffusion Steps for Training and Inference.} 
\label{app:diffusion_steps}
The number of training diffusion steps $K$ can heavily influence the modeling quality, where a larger value of $K$ generally leads to better performance.
At the same time, sampling from the diffusion models is recognized as a slow procedure, as it involves $K$ internal denoising steps.
We apply the \textit{stride sampling} technique~\cite{nichol2021improved} to accelerate the sampling process with reduced internal steps $N$, see Appendix~\ref{subsec:app_acc} for more details.
However, the sampling speed comes at the cost of quality. It is important to strike a balance between inference speed and prediction accuracy.
We investigate how to choose the number of $K$ and $N$ to significantly accelerate sampling without sacrificing model performance. \looseness=-1

\begin{figure}[t]
    \centering
    \includegraphics[width=0.65\columnwidth]{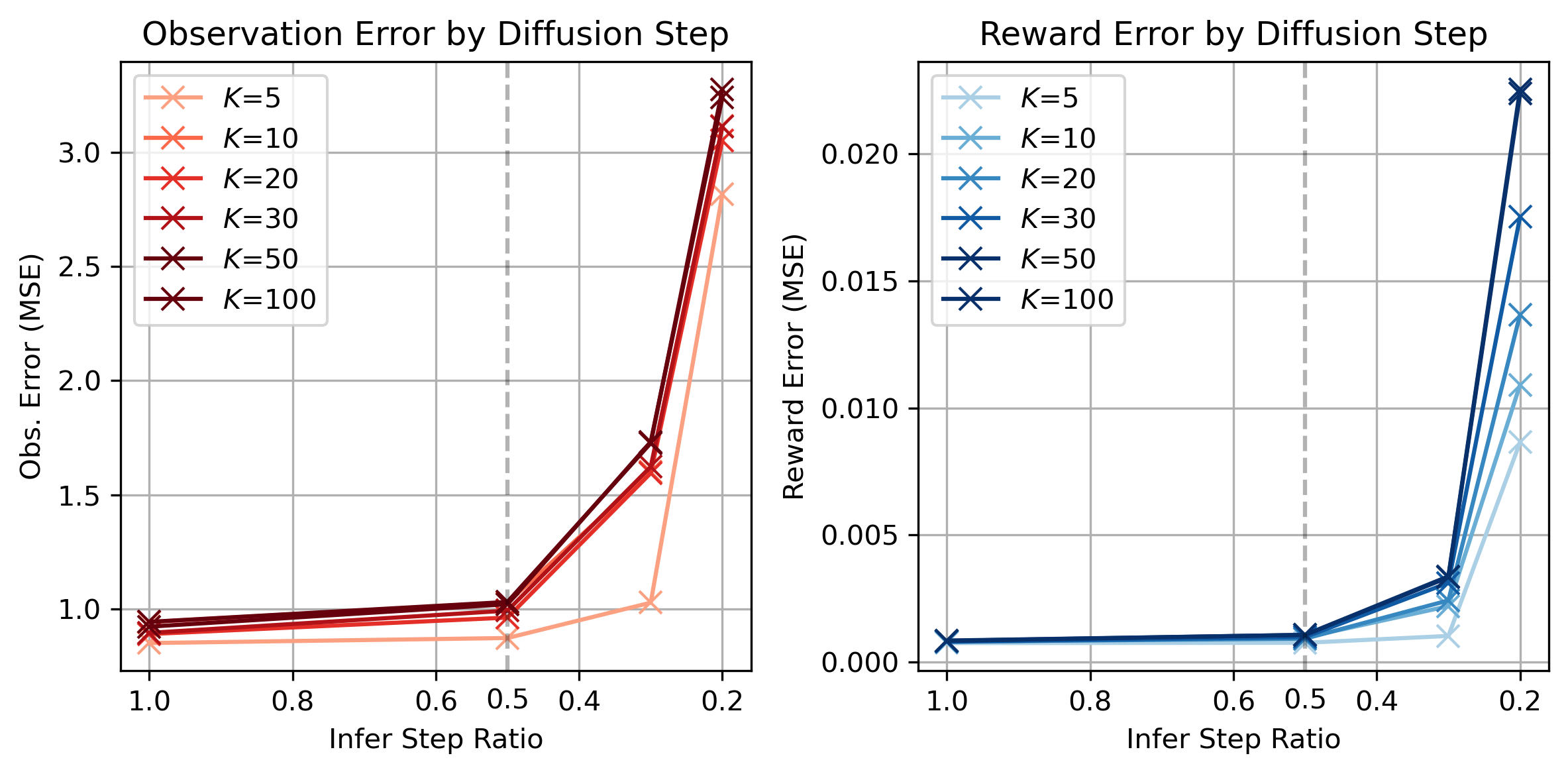}
    \caption{
    Average observation and reward prediction errors (across 9 tasks and simulation horizon $H\in[7]$) for DWM  DWM trained with $T=8$ and different diffusion steps $K$, as the inference step ratio $r_\text{ratio}$ changes.
    }
    \label{fig:prediction_error_T8}

    \includegraphics[width=0.65\columnwidth]{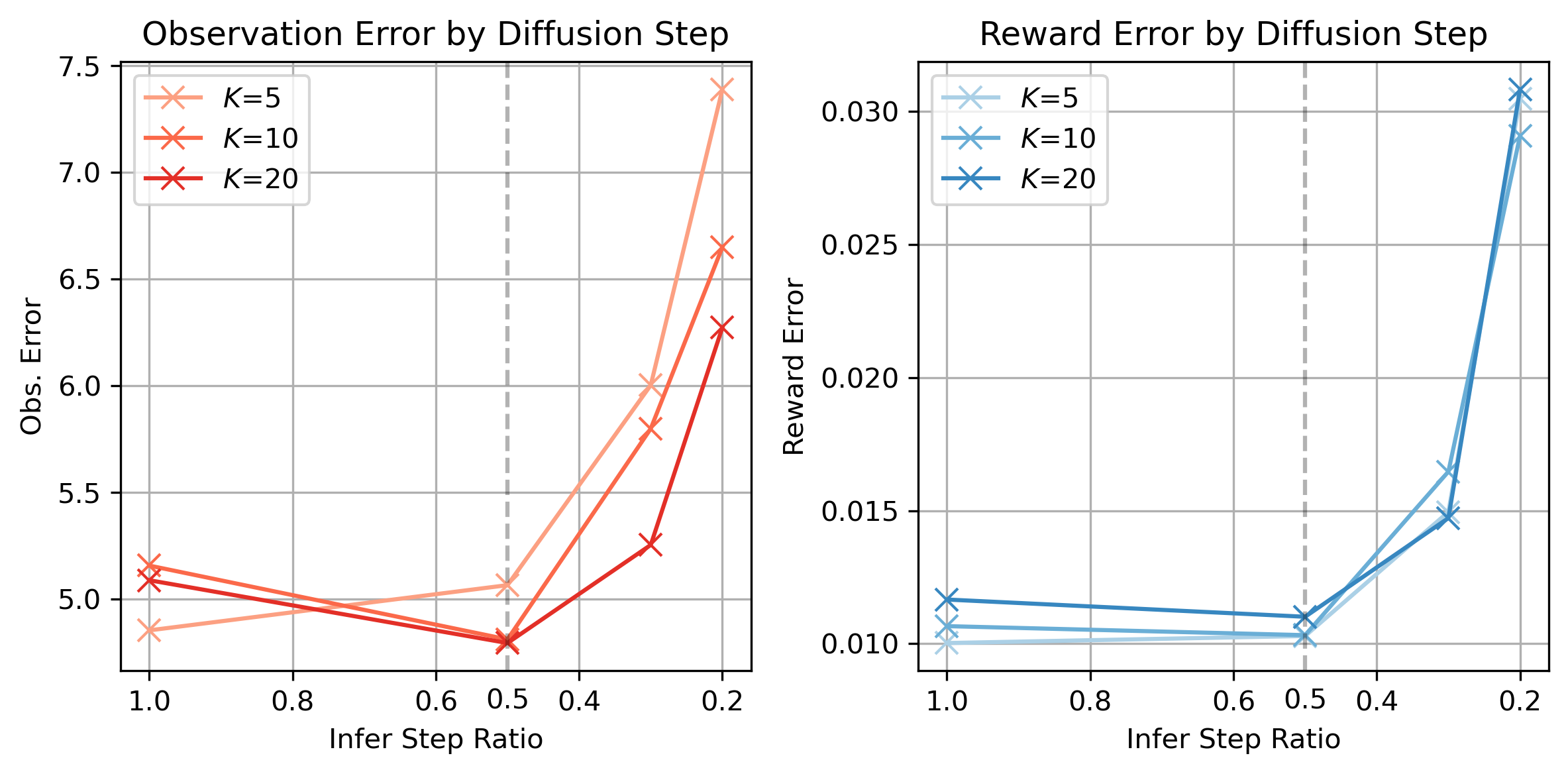}
    \caption{Average observation and reward prediction errors (across 9 tasks and simulation horizon $H \in [31]$) for DWM trained with $T=32$ and different diffusion steps $K$, as the inference step ratio $r_\text{ratio}$ changes.}
    \label{fig:prediction_error_T32}
\end{figure}

We train DWM with different numbers of diffusion steps $K\in\{5,10,20,30,50,100\}$, where the sequence length is $T=8$. We set four inference step ratios $r_\text{infer}\in\{0.2, 0.3, 0.5, 1.0\}$ and use $N=\lceil r_\text{infer}\cdot K \rceil$ internal steps in stride sampling. Figure~\ref{fig:prediction_error_T8} reports the prediction errors of DMW for both observation and reward sequences, defined in Equation~\eqref{eq:avg_pred_error}. We note that the prediction error depends on the evaluation RTG, and we report the best results across multiple values of it (listed in Table~\ref{tab:hyperparam}). An important observation is that $r_\text{infer}=0.5$ is a critical ridge for distinguishing the performances with different inference steps, where $N < K/2$ hurts the prediction accuracy significantly. 
Moreover, within the regime $r_\text{infer} \geq 0.5$, a small diffusion steps $K=5$ performs roughly the same as larger values.
Therefore, we choose $K=5$ and $r_\text{infer}=0.5$ for our main experiments, which leads to the number of sampling steps $N=3$. We have also repeated the above experiments for DWM with longer sequence length $T=32$. The results also support the choice $r_\text{infer}=0.5$ but favors $K=10$, see Figure~\ref{fig:prediction_error_T32}.

\subsubsection{Details Results}
Let $\tau$ denote a length-$T$ subtrajectory $(s_t, a_t, r_t, s_{t+1}, \ldots, s_{t+T-1}, r_{t+T-1})$.
The average prediction errors of a DWM $p_\theta$ for states and rewards along sequences are defined as:
\begin{align}
    \bar{\epsilon}_s = \E_{ \tau \sim \Doffline, \hat{s}_t\sim p_\theta(\cdot|s_1, a_1, g_\text{eval})}\left[\frac{1}{T}\sum_{t'=t}^{t+T-1} \norm{\hat{s}_{t'} - s_{t'}}^2 \right],  \; \text{and}\nonumber \\
    \bar{\epsilon}_r = \E_{\tau \sim \Doffline, \hat{s}_t\sim p_\theta(\cdot|s_1, a_1, g_\text{eval})}\left[\frac{1}{T}\sum_{t'=t}^{t+T-1} \norm{\hat{r}_{t'} - r_{t'}}^2 \right]. \nonumber \\
  \label{eq:avg_pred_error}
\end{align}
We first note that all the prediction errors depend on the evaluation RTG $\geval$. For the ease of clean presentation, we search over multiple values of $\geval$, listed in Table~\ref{tab:hyperparam}, and report the best results.

In addition to Figure~\ref{fig:prediction_error_T8}, the average prediction errors for diffusion models with $T=32$ (longer sequence) and diffusion steps $K\in\{5, 10, 20\}$ are shown in Figure~\ref{fig:prediction_error_T32}. 
Based on the results, $K=10$ and $r_\text{infer}=0.5$ are selected to strike a good balance between prediction accuracy and inference speed. The corresponding numerical results are listed in Table~\ref{tab:prediction_error_T8} and \ref{tab:prediction_error_T32}.

\begin{table}[H]
\centering
\small
\begin{tabular}{ccccc}
\toprule
Diffusion Step $K$ & Infer Step Ratio $r_\text{ratio}$ & Infer Step $N$ & Obs. Error & Reward Error \\
\midrule
\multirow{4}{*}{5} & 0.2 & 1 & 2.815 & 0.009 \\
& 0.3 & 2 & 1.028 & 0.001 \\
& 0.5 & 3 &  0.873 & 0.001 \\
& 1.0 & 5 & 0.851 & 0.001 \\ \hline
\multirow{4}{*}{10} & 0.2  & 1 & 3.114 & 0.011 \\
& 0.3  & 2 & 1.601 & 0.002 \\
& 0.5  & 3 & 1.028 & 0.001 \\
& 1.0  & 5 & 0.943 & 0.001 \\ \hline
\multirow{4}{*}{20} & 0.2  & 1 & 3.052 & 0.014 \\
& 0.3  & 2 & 1.595 & 0.002 \\
& 0.5  & 3 & 0.963 & 0.001 \\
& 1.0  & 5 & 0.890 & 0.001 \\ \hline
\multirow{4}{*}{30} & 0.2  & 1 & 3.112 & 0.018 \\
& 0.3  & 2 & 1.623 & 0.003 \\
& 0.5  & 3 & 0.993 & 0.001 \\
& 1.0  & 5 & 0.896 & 0.001 \\ \hline
\multirow{4}{*}{50} & 0.2  & 1 & 3.275 & 0.022 \\
& 0.3  & 2 & 1.726 & 0.003 \\
& 0.5  & 3 & 1.031 & 0.001 \\
& 1.0  & 5 & 0.944 & 0.001 \\ \hline
\multirow{4}{*}{100} & 0.2  & 1 & 3.239 & 0.023 \\
& 0.3  & 2 & 1.732 & 0.003 \\
& 0.5  & 3 & 1.021 & 0.001 \\
& 1.0  & 5 & 0.923 & 0.001 \\
\bottomrule
\end{tabular}
\caption{The average (across tasks and simulation horizon $H\in[7]$) observation and reward prediction errors for DWM with $T=8$ and different inference steps $N=\lceil r_\text{infer}\cdot K \rceil$.}
\label{tab:prediction_error_T8}
\end{table}

\begin{table}[H]
\centering
\small
\begin{tabular}{ccccc}
\toprule
Diffusion Step $K$ & Infer Step Ratio $r_\text{infer}$ & Infer Step $N$ & Obs. Error & Reward Error \\
\midrule
\multirow{4}{*}{5} & 0.2 & 1 & 7.390 & 0.030 \\
& 0.3  & 2 & 6.003 & 0.015 \\
& 0.5 & 3 & 5.065 & 0.010 \\
& 1.0  & 5 & 4.853 & 0.010 \\ \hline
\multirow{4}{*}{10} & 0.2  & 1 & 6.650 & 0.029 \\
& 0.3  & 2 & 5.799 & 0.016 \\
& 0.5  & 3 & 4.811 & 0.010 \\
& 1.0  & 5 & 5.157 & 0.011 \\ \hline
\multirow{4}{*}{20} & 0.2  & 1 & 6.273 & 0.031 \\
& 0.3  & 2 & 5.254 & 0.015 \\
& 0.5  & 3 & 4.794 & 0.011 \\
& 1.0  & 5 & 5.088 & 0.012 \\
\bottomrule
\end{tabular}
\caption{The average (across tasks and simulation horizon $H\in[31]$) observation and reward prediction errors for DWM with $T=32$ and different inference steps $N=\lceil r_\text{infer}\cdot K \rceil$.}
\label{tab:prediction_error_T32}
\end{table}

\subsection{Ablation: Sequence Length of Diffusion World Model}
\label{app:seq_length_dwm}
We further compare the average performances of algorithms with DWM trained with sequence length $T=8$ and $T=32$.
Table~\ref{tab:rollout_compare} presents average best return across 9 tasks (searched over RTG values and simulation horizon $H$). 
Even though DWM is robust to long-horizon simulation and in certain cases we have found the optimal $H$ is larger than $8$, 
we found using $T=32$ improves the performance of DWM-IQL, but slightly hurts the performance of DWM-TD3BC. 
\begin{table}[H]
\centering
\small
\begin{tabular}{cc|cc}
\toprule
\multicolumn{2}{c|}{DWM-TD3BC} & \multicolumn{2}{c}{DWM-IQL (w/o $\lambda$)}\\
T=8 & T=32 & T=8 & T=32    \\  \hline
0.68 $\pm$ 0.10 & 0.60 $\pm$ 0.12 & 0.57 $\pm$ 0.09 & 0.61$\pm$ 0.10\\
\bottomrule
\end{tabular}
\caption{The average performance of DWM algorithms across 9 tasks, using DWM with different sequence lengths.}
\label{tab:rollout_compare}
\end{table}
Therefore, we choose $T=8$ for our main experiments.

\subsection{Ablation: OOD Evaluation RTG Values} 
\label{app:ood_rtg}
We found that the evaluation RTG values play a critical role in determining the performance of our algorithm. Our preliminary experiments on trajectory preidction have suggested that in distribution evaluation RTGs underperforms OOD RTGs, see Appendix~\ref{subsec:app_in_vs_ood}. Figure~\ref{fig:rtg} reports the return of DWM-IQL and DWM-TD3BC across 3 tasks, with different values of $g_\text{eval}$\footnote{We note that the return and RTG are normalized in different ways: the return computed by the D4RL benchmark is undiscounted and normalized by the performance of one SAC policy, whereas the RTG we use in training is discounted and normalized by hand-selected constants.}. We report the results averaged over different simulation horizons 1, 3, 5 and 7. 
The compared RTG values are different for each task, but are all OOD. Appendix~\ref{subsec:app_data_dist} shows the distributions of training RTGs for each task.  The results show that the actual return does not always match with the specified $\geval$. This is a well-known issue of return-conditioned RL methods~\cite{emmons2021rvs, zheng2022online, nguyen2022conserweightive}. Nonetheless, 
OOD evaluation RTGs generally performs well. Figure~\ref{fig:rtg} shows both DWM-TD3BC and DWM-IQL are robust to OOD evaluation RTGs. We emphasize the reported return is averaged over training instances with different simulation horizons, where the peak performance, reported in Table~\ref{tab:compare_mbrl} is higher. Our intuition is to encourage the diffusion model to take
an optimistic view of the future return for the current state. On the other hand, the evaluation RTG cannot be overly high. As shown in task \textit{halfcheetah-mr}, increasing RTG $g_\text{eval}>0.4$ will further decrease the actual performances for both methods. The optimal RTG values vary from task to task, and the complete experiment results are provided below.

\begin{figure}[htbp]
    \centering
    \includegraphics[height=0.2\columnwidth]{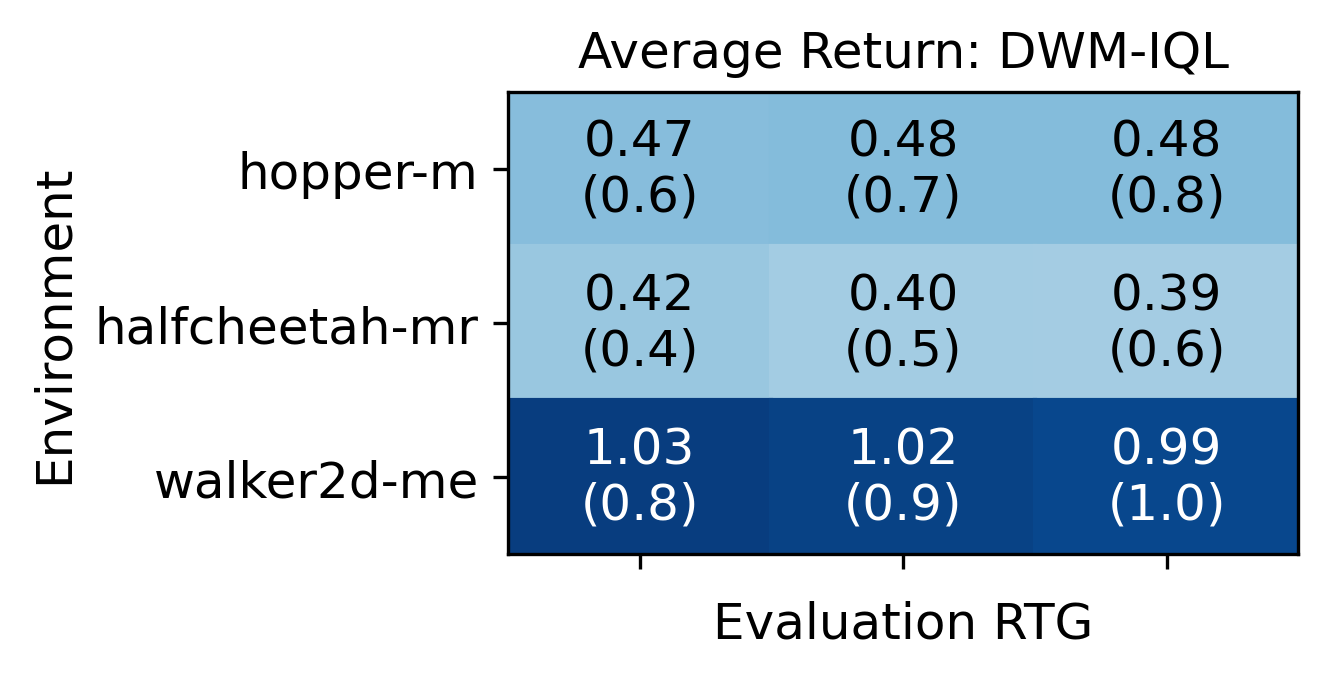}
    \includegraphics[height=0.2\columnwidth]{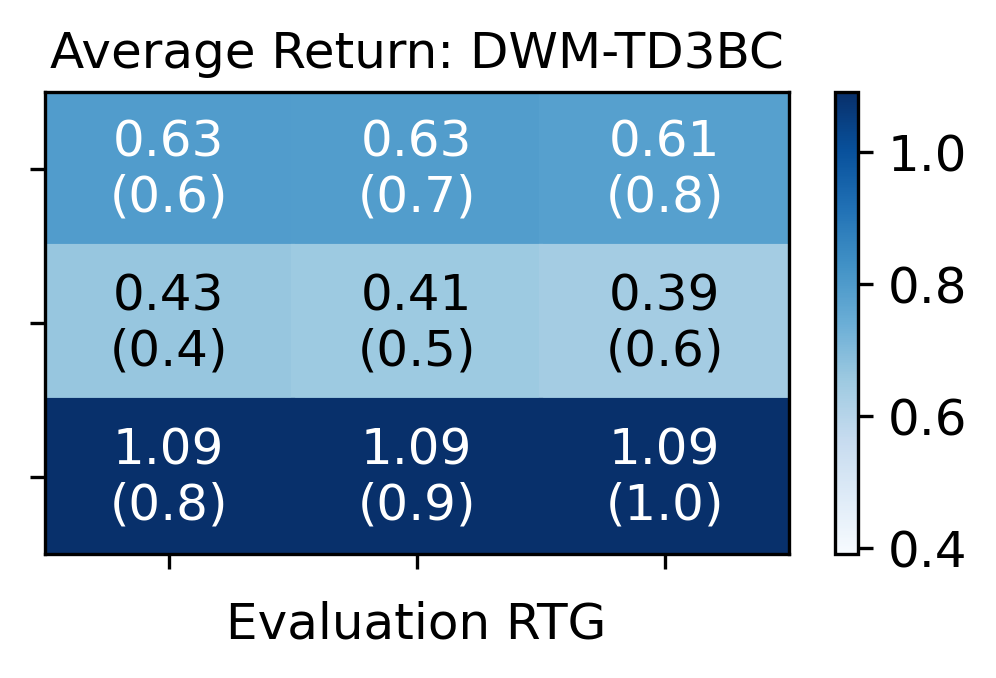}
    \caption{Comparison of DWM methods using different evaluation RTG values (displayed in parenthesis).}
    \label{fig:rtg}
\end{figure}

\subsubsection{Offline Data Distribution}
\label{subsec:app_data_dist}
\begin{figure}[H]
    \includegraphics[width=0.3\columnwidth]{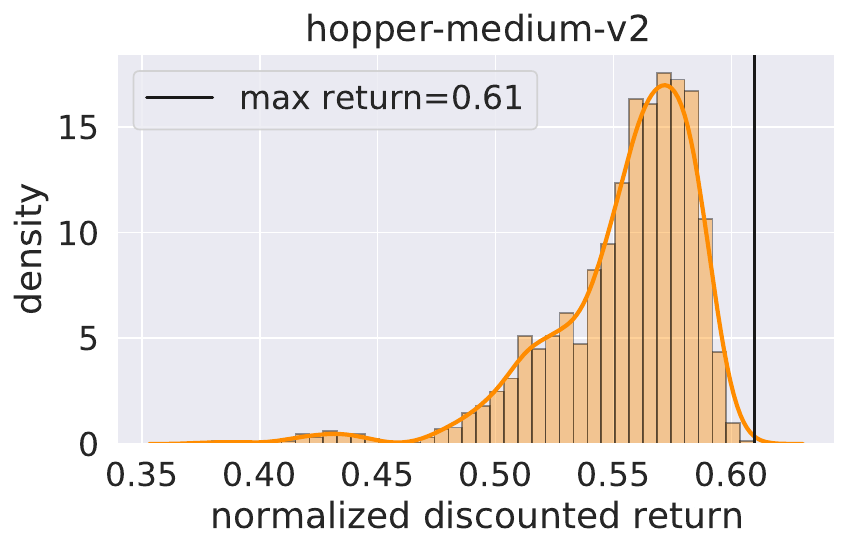}
    \includegraphics[width=0.3\columnwidth]{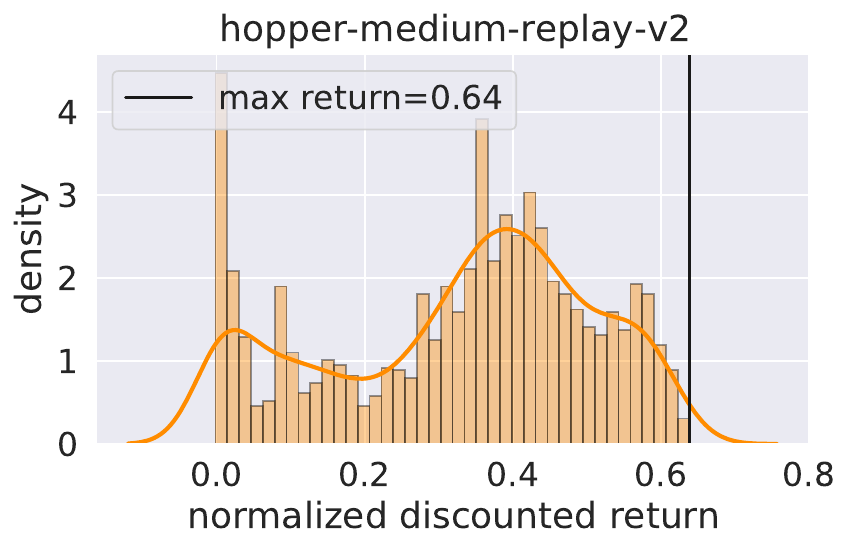}
    \includegraphics[width=0.3\columnwidth]{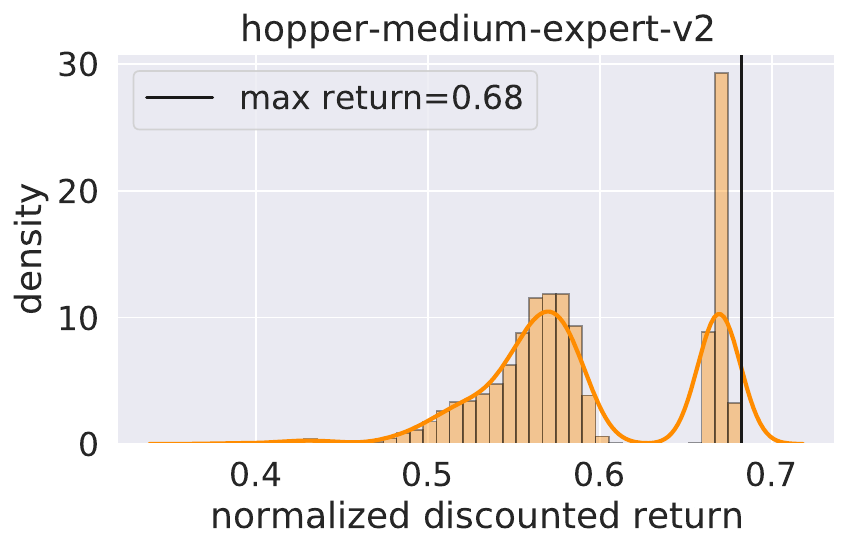}
    \includegraphics[width=0.3\columnwidth]{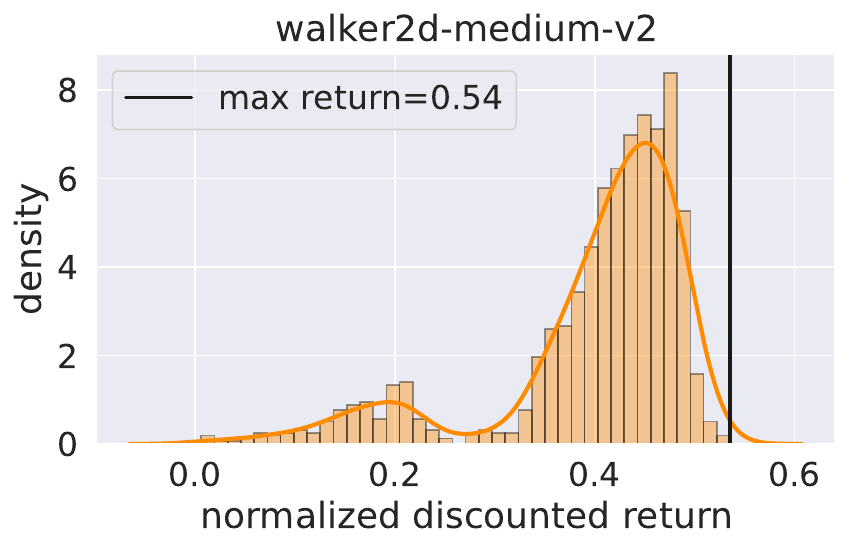}
    \includegraphics[width=0.3\columnwidth]{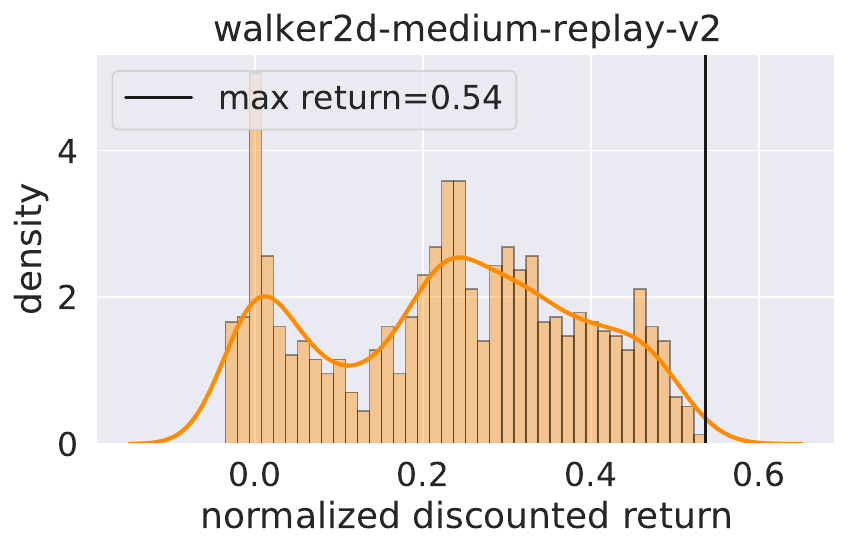}
    \includegraphics[width=0.3\columnwidth]{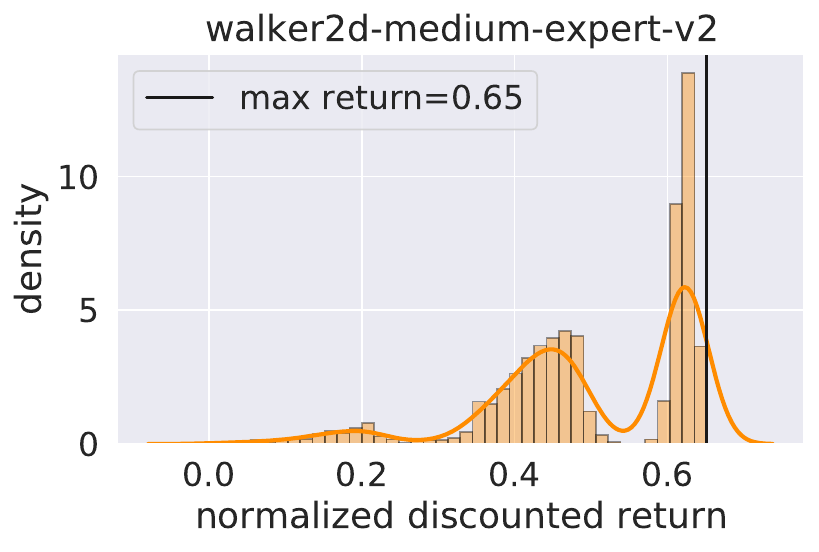}
    \includegraphics[width=0.3\columnwidth]{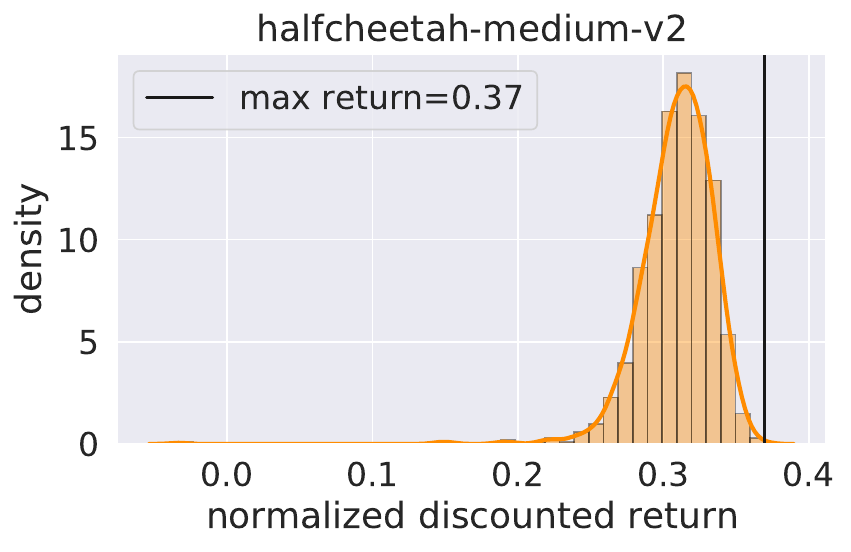}
    \includegraphics[width=0.33\columnwidth]{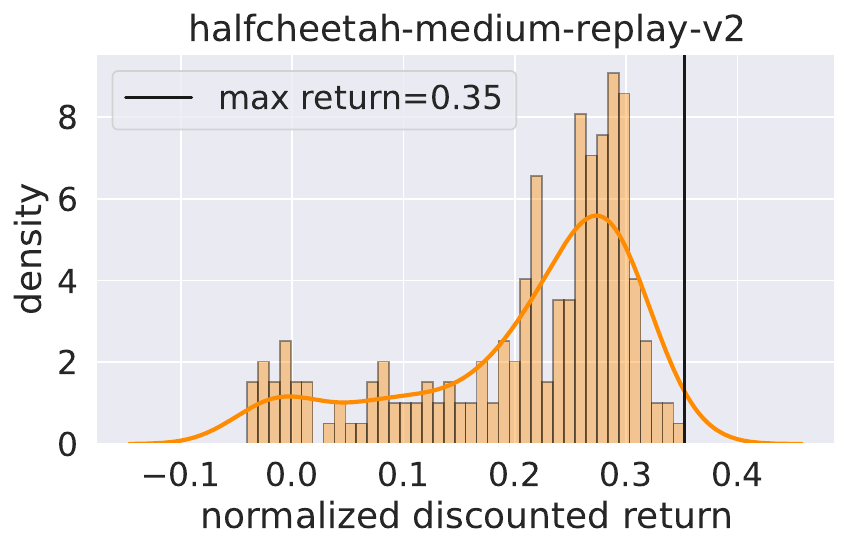}
    \includegraphics[width=0.34\columnwidth]{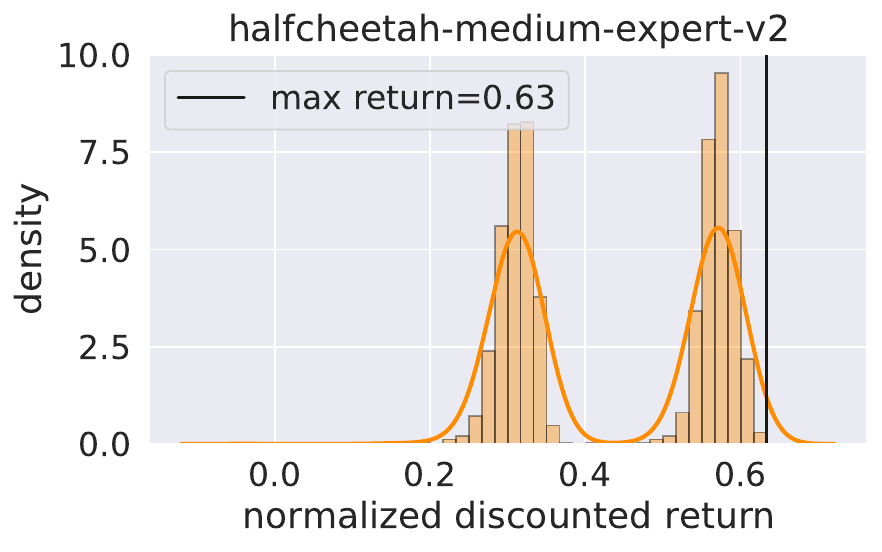}
    \caption{Normalized discounted returns for each environment.}
    \label{fig:norm_return}
\end{figure}

The normalized discounted return (as RTG labels in training) for the entire D4RL dataset over the nine tasks are analyzed in Fig.~\ref{fig:norm_return}. Compared with RTG values in our experiments as Table~\ref{tab:hyperparam}, 
the data maximum is usually smaller than the evaluation RTG values that leads to higher performances, as observed in our empirical experiments.

\newpage
\subsubsection{In-Distribution v.s. Out-of-Distribution RTG}
\label{subsec:app_in_vs_ood}
\begin{figure}[H]
    \includegraphics[width=\columnwidth]{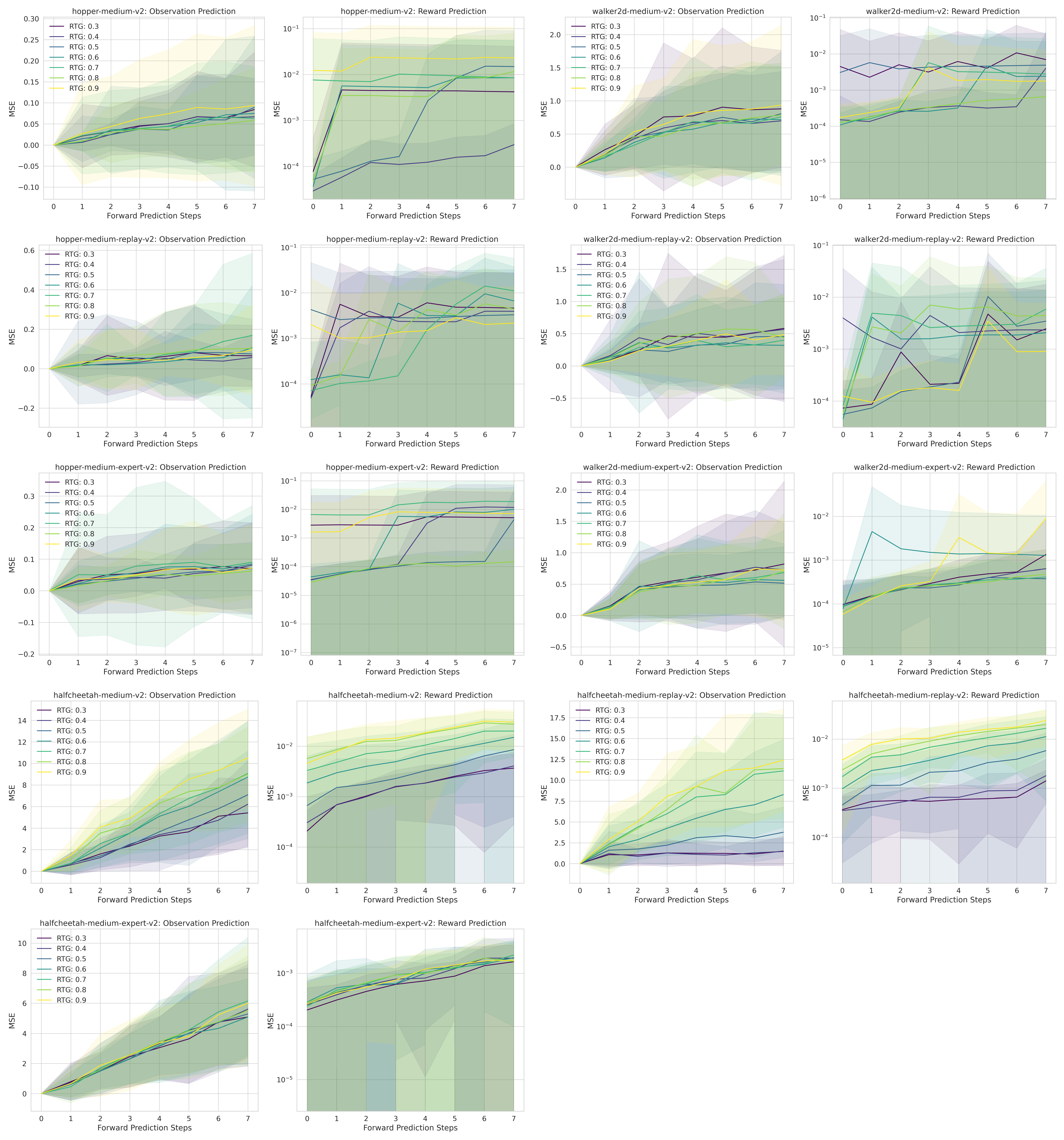}
    \caption{The breakdown prediction errors of DWM at each prediction timestep with different RTGs. The DWM is trained with $T=8$ and $K=5$.}
    \label{fig:breakdown_prediction_error_T8}
\end{figure}

\begin{figure}[H]
    \includegraphics[width=\columnwidth]{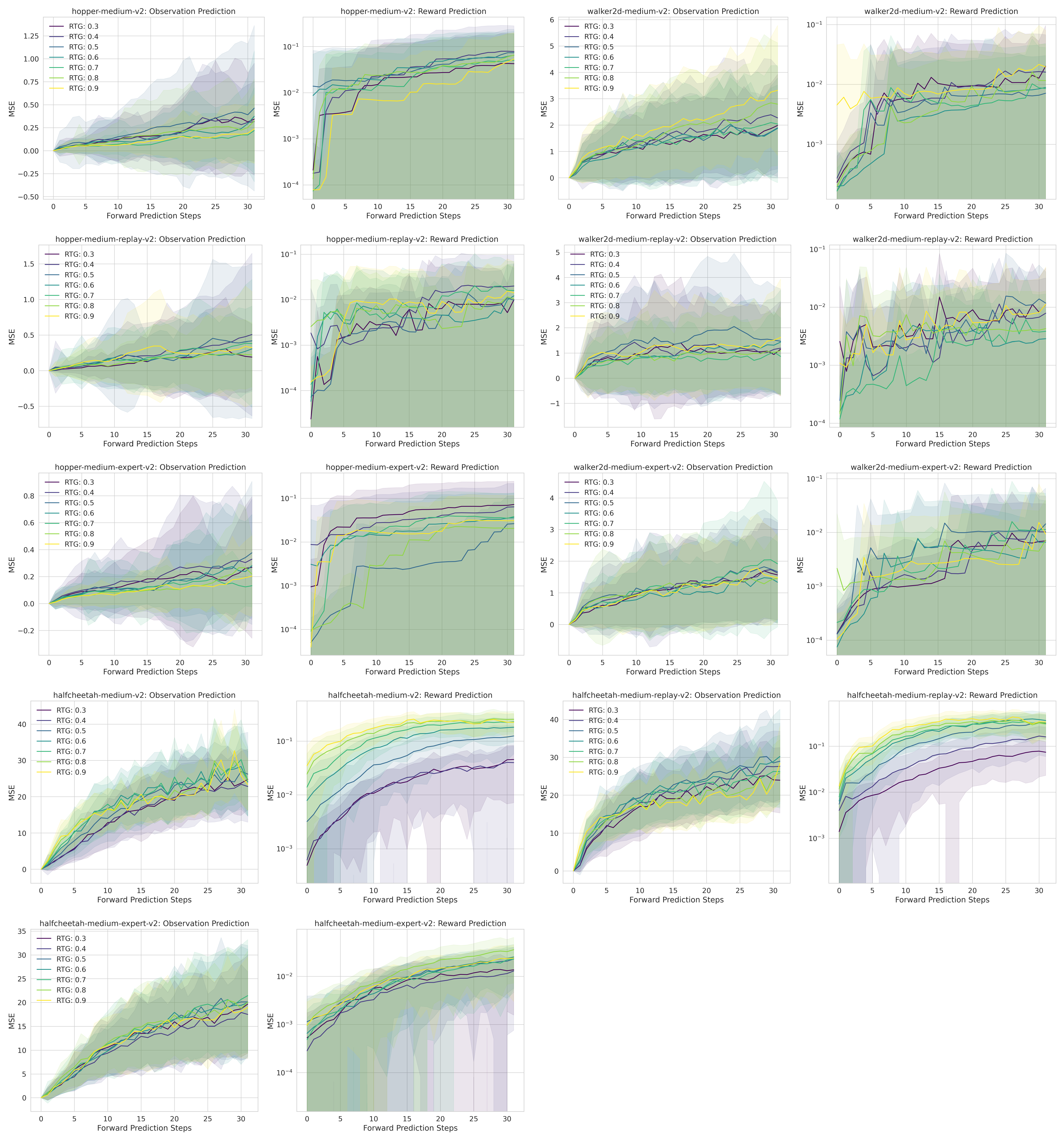}
    \caption{The breakdown prediction errors of DWM at each prediction timestep with different RTG.  The DWM is trained with $T=32$ and $K=10$.}
    \label{fig:breakdown_prediction_error_T32}
\end{figure}

Analogous to Equation~\eqref{eq:avg_pred_error}, we can define the breakdown prediction errors for each timestep $t'$, $t \leq t' \leq t+T-1$.
Figure~\ref{fig:breakdown_prediction_error_T8} and Figure~\ref{fig:breakdown_prediction_error_T32} plot the results,
using different values of $\geval$. The OOD RTGs generally perform better.

It is worth noting that the prediction error naturally grows as the horizon increases. Intuitively, given a fixed environment, the initial state of the D4RL datasets are very similar, whereas the subsequent states after multiple timesteps become quite different.

\subsection{Ablation: $\lambda$-Return Value Estimation}
\label{app:lambda_return}
The Dreamer series of work~\cite{hafner2019dream, hafner2020mastering, hafner2023mastering} applies the $\lambda$-return technique \cite{schulman2015high} for value estimation, used the imagined trajectory.
This technique can be seamlessly embedded into our framework as a modification of the standard Diff-MVE.
More precisely, given a state-action pair $(s_t, a_t)$ sampled from the offline dataset,
we recursively compute the $\lambda$-target value for $h=H, \ldots, 0$:
\begin{align}
  \hat{Q}^\lambda_{t+h} = \rhat_{t+h} + \gamma \begin{cases} (1-\lambda)Q_{\bar{\phi}}(\shat_{t+h+1}, \pi_{\bar{\psi}}(\shat_{t+h+1})) + \lambda \hat{Q}^\lambda_{t+h+1} & \text{ if }h<H  \\ Q_{\bar{\phi}}(\shat_{t+H}, \pi_{\bar{\psi}}(\shat_{t+H})) & \text{ if }h=H \end{cases} 
\label{eq:lambda_return}
\end{align}
using DWM predicted states $\set{\shat_{t+h}}_{h=0}^H$ and rewards $\{\rhat_t\}_{h=0}^H$.
We can use $\Qhat^\lambda_t$ as the target $Q$ value for TD learning, as a modification of line~\ref{algo:diffusion_mve} of Algorithm~\ref{algo:diffusion_mbrl}. 
For algorithms that also learn the state-only value function, like IQL, the $Q_{\bar{\phi}}$ function can be replaced by the $V_{\bar{\psi}}$ function.
Worth noting, Equation~\eqref{eq:lambda_return} reduces to the vanilla Diff-MVE when $\lambda=1$. \looseness=-1

We conduct experiments to compare the vanilla diff-MVE and the $\lambda$-return variant for DWM-TD3BC and DWM-IQL, using $\lambda=0.95$. 
We search over RTG values (specified in Appendix Table~\ref{tab:hyperparam})  and simulation horizons $1,3,5,7$. The results are summarized in Table~\ref{tab:compare_lambda_return}. The $\lambda$-return technique
is beneficial for DWM-IQL, but harmful for DWM-TD3BC. We speculate that since Equation~\eqref{eq:lambda_return} iteratively invokes the $Q_{\bar{\phi}}$ or the $V_{\bar{\psi}}$ function, it favors approaches with more accurate value estimations. While IQL regularizes the value functions, TD3+BC only 
has policy regularization and is shown to be more prone to the value over-estimation issue in our experiments. 
Based on these results, we incorporated the $\lambda$-return technique into DWM-IQL, but let DWM-TD3BC use the vanilla Diff-MVE.
We let DWM-PQL uses the vanilla Diff-MVE for the sake of algorithmic simplicity.

\begin{table}[htbp]
\centering
\small
\begin{tabular}{c|cc|cc}
\toprule
 & \multicolumn{2}{c|}{\textbf{DWM-TD3BC}} & \multicolumn{2}{c}{\textbf{DWM-IQL}}  \\  
 Env. & w/o $\lambda$ &  w/ $\lambda$  & w/o $\lambda$ &  w/ $\lambda$  \\  \midrule
hopper-m  & 0.65 $\pm$ 0.10 & \textbf{0.68 $\pm$ 0.13} & 0.50 $\pm$ 0.08 & 0.54 $\pm$ 0.11 \\
walker2d-m & 0.70 $\pm$ 0.15 & 0.74 $\pm$ 0.08  & 0.62 $\pm$ 0.19 & \textbf{0.76 $\pm$ 0.05}\\
halfcheetah-m & \textbf{0.46 $\pm$ 0.01} & 0.40 $\pm$ 0.01  & \textbf{0.46 $\pm$ 0.01} & 0.44 $\pm$ 0.01    \\
hopper-mr & 0.53 $\pm$ 0.09 & 0.50 $\pm$ 0.23  & 0.29 $\pm$ 0.04 & \textbf{0.61 $\pm$ 0.13} \\
walker2d-mr & \textbf{0.46 $\pm$ 0.19} & 0.23 $\pm$ 0.10 & 0.27 $\pm$ 0.09 &  0.35 $\pm$ 0.14  \\
halfcheetah-mr & \textbf{0.43 $\pm$ 0.01} & 0.39 $\pm$ 0.02 & \textbf{0.43 $\pm$ 0.01} & 0.41 $\pm$ 0.01 \\
hopper-me & 1.03 $\pm$ 0.14 & \textbf{1.05 $\pm$ 0.16}  & 0.78 $\pm$ 0.24 & 0.90 $\pm$ 0.25  \\
walker2d-me & \textbf{1.10 $\pm$ 0.00} & 0.89 $\pm$ 0.13 & 1.08 $\pm$ 0.03 & 1.04 $\pm$ 0.10 \\
halfcheetah-me & \textbf{0.75 $\pm$ 0.16} & 0.71 $\pm$ 0.22 & 0.73 $\pm$ 0.14 & 0.74 $\pm$ 0.16  \\ \hline
Avg. & \textbf{0.68} &  0.62 & 0.57 & 0.64   \\
\bottomrule
\end{tabular}
\caption{Comparison of the performance of DWM methods using vanilla Diff-MVE and the $\lambda$-return variant. }
\label{tab:compare_lambda_return}
\end{table}

\subsection{Ablation: RTG Relabeling and Model Fine-tuning}
\label{app:rtg_relabel}
Unlike dynamic programming in traditional RL, sequential modeling methods like diffusion models and DT are suspected to fail to stitch suboptimal trajectories. RTG relabeling is proposed to alleviate this problem for DT~\cite{yamagata2023q}, through iteratively relabeling RTG $g$ from training dataset to be: 
\begin{equation}
    \tilde{g}_t = r_t + \gamma \max (g_{t+1}, \hat{V}(s_{t+1})) = \max(g_t, r_t + \hat{V}(s_{t+1}),
    \label{eq:rtg_relabel_discounted}
\end{equation}
where the $\hat{V}$ function is separately learned from the offline dataset using CQL~\cite{kumar2020conservative}, and the max operator is used to prevent underestimation due to the pessimism of $\hat{V}$. The original formulation by \citet{yamagata2023q} does not include the $\gamma$ term as DT uses undiscounted RTG, i.e., $\gamma=1$.

We apply the RTG relabeling for DWM fine-tuning in the policy learning phase of vanilla DWM-IQL algorithm, without the $\lambda$-return technique. The value function $\hat{V}$ comes from the IQL algorithm. We take the first $10\%$ steps of the entire policy learning as warm up steps, where we do not apply RTG relabeling. This is because 
$\hat{V}$ can be inaccurate at the beginning of training. The modified algorithm DWM-IQL(R) achieves an average score of $0.61$, improved over score $0.57$ for DWM-IQL(w/o $\lambda$), under exactly the same training and test settings. Results are provided in Table~\ref{tab:compare_rtg_relabel} with details in Table~\ref{tab:mbiql_relabel}. Nonetheless, the improvement is of limited unity compared with the $\lambda$-return, thus we do not include it in the final design.

\begin{table}[H]
\centering
\small
\begin{tabular}{ccccc}
\toprule
Env & DWM-IQL(w/o $\lambda$) & DWM-IQL(w/ RTG Relabel)   \\  \hline
hopper-m  & 0.50 $\pm$ 0.08   & 0.59 ± 0.13\\
walker2d-m & 0.62 $\pm$ 0.19   & 0.65 ± 0.17 \\
halfcheetah-m & 0.46 $\pm$ 0.01  & 0.47 ± 0.01   \\
hopper-mr & 0.29 $\pm$ 0.04  & 0.27 ± 0.02 \\
walker2d-mr & 0.27 $\pm$ 0.09  & 0.32 ± 0.15 \\
halfcheetah-mr & 0.43 $\pm$ 0.01  & 0.43 ± 0.02 \\
hopper-me & 0.78 $\pm$ 0.24 & 0.88 ± 0.26 \\
walker2d-me & 1.08 $\pm$ 0.03  & 1.1 ± 0.0 \\
halfcheetah-me& 0.73 $\pm$ 0.14  & 0.79 ± 0.10 \\ \hline
Avg. & 0.57 & 0.61   \\
\bottomrule
\end{tabular}
\caption{The results of finetuning DMW via RTG relabeling in the policy training phase: normalized return (mean $\pm$ std) }
\label{tab:compare_rtg_relabel}
\end{table}

\begin{table}[H]
\centering
\resizebox{\textwidth}{!}{
\begin{tabular}{ccc|ccc|ccc}
\hline
\multicolumn{3}{c|}{\textbf{hopper-medium-v2}} & \multicolumn{3}{c|}{\textbf{hopper-medium-replay-v2}} & \multicolumn{3}{c}{\textbf{hopper-medium-expert-v2}} \\
\hline
Simulation Horizon & RTG & Return (Mean $\pm$ Std) & Simulation Horizon & RTG & Return (Mean $\pm$ Std) & Simulation Horizon & RTG & Return (Mean $\pm$ Std)\\
\hline
1 & 0.6 &  0.54 $\pm$ 0.06 & 1 & 0.6  & 0.21 $\pm$ 0.02 & 1 & 0.7  & 0.71 $\pm$ 0.30\\
3 & 0.6 &  0.51 $\pm$ 0.09 & 3 & 0.6  & 0.22 $\pm$ 0.02 & 3 & 0.7  & 0.79 $\pm$ 0.24\\
5 & 0.6 &  0.51 $\pm$ 0.12 & 5 & 0.6  & 0.23 $\pm$ 0.01 & 5 & 0.7  & 0.71 $\pm$ 0.18\\
7 & 0.6 &  0.59 $\pm$ 0.13 & 7 & 0.6  & 0.23 $\pm$ 0.01 & 7 & 0.7  & 0.79 $\pm$ 0.24\\
1 & 0.7 &  0.51 $\pm$ 0.04 & 1 & 0.7  & 0.21 $\pm$ 0.01 & 1 & 0.8  & 0.59 $\pm$ 0.25\\
3 & 0.7 &  0.49 $\pm$ 0.07 & 3 & 0.7  & 0.23 $\pm$ 0.01 & 3 & 0.7  & 0.79 $\pm$ 0.24 \\
5 & 0.7 &  0.49 $\pm$ 0.08 & 5 & 0.7  & 0.25 $\pm$ 0.02 & 5 & 0.7  & 0.71 $\pm$ 0.18\\
7 & 0.7 &  0.48 $\pm$ 0.07 & 7 & 0.7  & 0.25 $\pm$ 0.03 & 7 & 0.7  & 0.79 $\pm$ 0.24 \\
1 & 0.8 &  0.52 $\pm$ 0.04 & 1 & 0.8  & 0.2 $\pm$ 0.03 & 1 & 0.9  & 0.60 $\pm$ 0.28\\
3 & 0.8 &  0.48 $\pm$ 0.06 & 3 & 0.8  & 0.23 $\pm$ 0.03 & 3 & 0.9  & 0.88 $\pm$ 0.26\\
5 & 0.8 &  0.48 $\pm$ 0.08 & 5 & 0.8  & 0.26 $\pm$ 0.05 & 5 & 0.9  & 0.83 $\pm$ 0.23\\
7 & 0.8 &  0.47 $\pm$ 0.07 & 7 & 0.8  & 0.27 $\pm$ 0.02 & 7 & 0.9  & 0.75 $\pm$ 0.28\\
\hline
\multicolumn{3}{c|}{\textbf{walker2d-medium-v2}} & \multicolumn{3}{c|}{\textbf{walker2d-medium-replay-v2}} & \multicolumn{3}{c}{\textbf{walker2d-medium-expert-v2}} \\
\hline
Simulation Horizon & RTG & Return (Mean $\pm$ Std) & Simulation Horizon & RTG & Return (Mean $\pm$ Std) & Simulation Horizon & RTG & Return (Mean $\pm$ Std) \\
\hline
1 & 0.6 & 0.59 $\pm$ 0.19 & 1 & 0.6 & 0.25 $\pm$ 0.12 & 1 & 0.8 & 1.08 $\pm$ 0.00 \\
3 & 0.6 & 0.57 $\pm$ 0.18 & 3 & 0.6 & 0.21 $\pm$ 0.11 & 3 & 0.8 & 1.07 $\pm$ 0.06\\
5 & 0.6 & 0.58 $\pm$ 0.21 & 5 & 0.6 & 0.18 $\pm$ 0.05 & 5 & 0.8 & 1.09 $\pm$ 0.00 \\
7 & 0.6 & 0.50 $\pm$ 0.2 & 7 & 0.6 & 0.15 $\pm$ 0.09 & 7 & 0.8 & 1.09 $\pm$ 0.02\\
1 & 0.7 & 0.65 $\pm$ 0.17 & 1 & 0.7 & 0.32 $\pm$ 0.15 & 1 & 0.9 & 1.08 $\pm$ 0.01\\
3 & 0.7 & 0.55 $\pm$ 0.19 & 3 & 0.7 & 0.16 $\pm$ 0.08 & 3 & 0.9 & 1.07 $\pm$ 0.05 \\
5 & 0.7 & 0.54 $\pm$ 0.23 & 5 & 0.7 & 0.20 $\pm$ 0.03 & 5 & 0.9 & 1.10 $\pm$ 0.00 \\
7 & 0.7  & 0.52 $\pm$ 0.18 & 7 & 0.7 & 0.15 $\pm$ 0.10 & 7 & 0.9 & 1.10 $\pm$ 0.01\\
1 & 0.8 & 0.60 $\pm$ 0.22 & 1 & 0.8 & 0.25 $\pm$ 0.11 & 1 & 1.0 & 1.08 $\pm$ 0.0\\
3 & 0.8 & 0.57 $\pm$ 0.21 & 3 & 0.8 & 0.18 $\pm$ 0.08 & 3 & 1.0 & 1.08 $\pm$ 0.03 \\
5 & 0.8 & 0.56 $\pm$ 0.21 & 5 & 0.8 & 0.21 $\pm$ 0.02 & 5 & 1.0 & 1.08 $\pm$ 0.04 \\
7 & 0.8 & 0.54 $\pm$ 0.23 & 7 & 0.8 & 0.18 $\pm$ 0.12 & 7 & 1.0 & 1.08 $\pm$ 0.04 \\
\hline
\multicolumn{3}{c|}{\textbf{halfcheetah-medium-v2}} & \multicolumn{3}{c|}{\textbf{halfcheetah-medium-replay-v2}} & \multicolumn{3}{c}{\textbf{halfcheetah-medium-expert-v2}} \\
\hline
Simulation Horizon & RTG & Return (Mean $\pm$ Std) & Simulation Horizon & RTG & Return (Mean $\pm$ Std) & Simulation Horizon & RTG & Return (Mean $\pm$ Std) \\
\hline
1 & 0.4 & 0.46 $\pm$ 0.01 & 1 & 0.4 & 0.43 $\pm$ 0.02 & 1 & 0.6 & 0.69 $\pm$ 0.12 \\
3 & 0.4 & 0.47 $\pm$ 0.01 & 3 & 0.4 & 0.43 $\pm$ 0.03 & 3 & 0.6 & 0.76 $\pm$ 0.11\\
5 & 0.4 & 0.47 $\pm$ 0.01 & 5 & 0.4 & 0.42 $\pm$ 0.01 & 5 & 0.6 & 0.76 $\pm$ 0.16\\
7 & 0.4 & 0.47 $\pm$ 0.01 & 7 & 0.4 & 0.42 $\pm$ 0.02 & 7 & 0.6 & 0.79 $\pm$ 0.10\\
1 & 0.5 & 0.45 $\pm$ 0.01 & 1 & 0.5 & 0.41 $\pm$ 0.03 & 1 & 0.7 & 0.70 $\pm$ 0.16\\
3 & 0.5 & 0.45 $\pm$ 0.01 & 3 & 0.5 & 0.41 $\pm$ 0.01 & 3 & 0.7 & 0.74 $\pm$ 0.14\\
5 & 0.5 & 0.44 $\pm$ 0.02 & 5 & 0.5 & 0.4 $\pm$ 0.01 & 5 & 0.7 & 0.75 $\pm$ 0.12\\
7 & 0.5 & 0.45 $\pm$ 0.01 & 7 & 0.5 & 0.4 $\pm$ 0.03 & 7 & 0.7 & 0.75 $\pm$ 0.11\\
1 & 0.6 & 0.43 $\pm$ 0.02 & 1 & 0.6 & 0.38 $\pm$ 0.03 & 1 & 0.8 & 0.68 $\pm$ 0.20\\
3 & 0.6 & 0.43 $\pm$ 0.01 & 3 & 0.6 & 0.38 $\pm$ 0.03 & 3 & 0.8 & 0.73 $\pm$ 0.19\\
5 & 0.6 & 0.43 $\pm$ 0.01 & 5 & 0.6 & 0.39 $\pm$ 0.03 & 5 & 0.8 & 0.72 $\pm$ 0.13\\
7 & 0.6 & 0.43 $\pm$ 0.02 & 7 & 0.6 & 0.38 $\pm$ 0.03 & 7 & 0.8 & 0.75 $\pm$ 0.13\\
\hline
\end{tabular}}
\caption{The normalized returns of DWM-IQL (w/ RTG relabel) with diffusion step $K=5$, sampling steps $N=3$, and simulation horizon $H\in\{1,3,5,7\}$.}
\label{tab:mbiql_relabel}
\end{table}

\newpage
\subsection{Example Raw Results}

While we cannot enumerate all the experiment results, we provide some raw results of DWM methods. Table~\ref{tab:mbiql} and \ref{tab:mbtd3bc} report the performance of DWM-IQL (without $\lambda$-return) and DQM-TD3BC with parameters swept over both simulation horizon and RTG values. Table~\ref{tab:mbpql} report the performance of DWM-PQL with parameters swept over pessimism coefficient $\lambda$ and RTG values, for a fixed horizon $H=5$.
\begin{table}[H]
\centering
\resizebox{\textwidth}{!}{
\begin{tabular}{ccc|ccc|ccc}
\hline
\multicolumn{3}{c|}{\textbf{hopper-medium-v2}} & \multicolumn{3}{c|}{\textbf{hopper-medium-replay-v2}} & \multicolumn{3}{c}{\textbf{hopper-medium-expert-v2}} \\
\hline
Simulation Horizon & RTG & Return (Mean $\pm$ Std) & Simulation Horizon & RTG & Return (Mean $\pm$ Std) & Simulation Horizon & RTG & Return (Mean $\pm$ Std)\\
\hline
1 & 0.6 & 0.47 $\pm$ 0.05 & 1 & 0.6 & 0.26 $\pm$ 0.06 & 1 & 0.7 & 0.57 $\pm$ 0.26 \\
3 & 0.6 & 0.47 $\pm$ 0.09 & 3 & 0.6 & 0.23 $\pm$ 0.01 & 3 & 0.7 & 0.66 $\pm$ 0.21 \\
5 & 0.6 & 0.46 $\pm$ 0.06 & 5 & 0.6 & 0.25 $\pm$ 0.04 & 5 & 0.7 & 0.78 $\pm$ 0.24 \\
7 & 0.6 & 0.49 $\pm$ 0.09 & 7 & 0.6 & 0.23 $\pm$ 0.02 & 7 & 0.7 & 0.73 $\pm$ 0.27 \\
1 & 0.7 & 0.5 $\pm$ 0.08 & 1 & 0.7 & 0.29 $\pm$ 0.04 & 1 & 0.8 & 0.43 $\pm$ 0.19 \\
3 & 0.7 & 0.46 $\pm$ 0.07 & 3 & 0.7 & 0.25 $\pm$ 0.02 & 3 & 0.8 & 0.7 $\pm$ 0.19 \\
5 & 0.7 & 0.48 $\pm$ 0.06 & 5 & 0.7 & 0.23 $\pm$ 0.01 & 5 & 0.8 & 0.72 $\pm$ 0.21 \\
7 & 0.7 & 0.48 $\pm$ 0.07 & 7 & 0.7 & 0.26 $\pm$ 0.01 & 7 & 0.8 & 0.75 $\pm$ 0.25 \\
1 & 0.8 & 0.49 $\pm$ 0.1 & 1 & 0.8 & 0.2 $\pm$ 0.05 & 1 & 0.9 & 0.47 $\pm$ 0.14 \\
3 & 0.8 & 0.47 $\pm$ 0.06 & 3 & 0.8 & 0.19 $\pm$ 0.03 & 3 & 0.9 & 0.52 $\pm$ 0.26 \\
5 & 0.8 & 0.47 $\pm$ 0.07 & 5 & 0.8 & 0.3 $\pm$ 0.06 & 5 & 0.9 & 0.59 $\pm$ 0.16 \\
7 & 0.8 & 0.49 $\pm$ 0.08 & 7 & 0.8 & 0.25 $\pm$ 0.05 & 7 & 0.9 & 0.61 $\pm$ 0.22 \\
\hline
\multicolumn{3}{c|}{\textbf{walker2d-medium-v2}} & \multicolumn{3}{c|}{\textbf{walker2d-medium-replay-v2}} & \multicolumn{3}{c}{\textbf{walker2d-medium-expert-v2}} \\
\hline
Simulation Horizon & RTG & Return (Mean $\pm$ Std) & Simulation Horizon & RTG & Return (Mean $\pm$ Std) & Simulation Horizon & RTG & Return (Mean $\pm$ Std) \\
\hline
1 & 0.6 & 0.46 $\pm$ 0.22 & 1 & 0.6 & 0.27 $\pm$ 0.09 & 1 & 0.8 & 0.97 $\pm$ 0.22 \\
3 & 0.6 & 0.55 $\pm$ 0.23 & 3 & 0.6 & 0.15 $\pm$ 0.02 & 3 & 0.8 & 1.03 $\pm$ 0.12 \\
5 & 0.6 & 0.54 $\pm$ 0.25 & 5 & 0.6 & 0.13 $\pm$ 0.01 & 5 & 0.8 & 1.05 $\pm$ 0.09 \\
7 & 0.6 & 0.5 $\pm$ 0.2 & 7 & 0.6 & 0.11 $\pm$ 0.0 & 7 & 0.8 & 1.08 $\pm$ 0.03 \\
1 & 0.7 & 0.62 $\pm$ 0.19 & 1 & 0.7 & 0.26 $\pm$ 0.14 & 1 & 0.9 & 1.01 $\pm$ 0.13 \\
3 & 0.7 & 0.59 $\pm$ 0.2 & 3 & 0.7 & 0.13 $\pm$ 0.12 & 3 & 0.9 & 1.05 $\pm$ 0.1 \\
5 & 0.7 & 0.56 $\pm$ 0.2 & 5 & 0.7 & 0.11 $\pm$ 0.01 & 5 & 0.9 & 1.05 $\pm$ 0.11 \\
7 & 0.7 & 0.47 $\pm$ 0.23 & 7 & 0.7 & 0.11 $\pm$ 0.0 & 7 & 0.9 & 0.96 $\pm$ 0.25 \\
1 & 0.8 & 0.59 $\pm$ 0.21 & 1 & 0.8 & 0.26 $\pm$ 0.17 & 1 & 1.0 & 0.97 $\pm$ 0.14 \\
3 & 0.8 & 0.6 $\pm$ 0.2 & 3 & 0.8 & 0.16 $\pm$ 0.12 & 3 & 1.0 & 1.0 $\pm$ 0.11 \\
5 & 0.8 & 0.56 $\pm$ 0.21 & 5 & 0.8 & 0.13 $\pm$ 0.11 & 5 & 1.0 & 1.03 $\pm$ 0.16 \\
7 & 0.8 & 0.59 $\pm$ 0.19 & 7 & 0.8 & 0.14 $\pm$ 0.1 & 7 & 1.0 & 0.97 $\pm$ 0.21 \\
\hline
\multicolumn{3}{c|}{\textbf{halfcheetah-medium-v2}} & \multicolumn{3}{c|}{\textbf{halfcheetah-medium-replay-v2}} & \multicolumn{3}{c}{\textbf{halfcheetah-medium-expert-v2}} \\
\hline
Simulation Horizon & RTG & Return (Mean $\pm$ Std) & Simulation Horizon & RTG & Return (Mean $\pm$ Std) & Simulation Horizon & RTG & Return (Mean $\pm$ Std) \\
\hline
1 & 0.4 & 0.45 $\pm$ 0.01 & 1 & 0.4 & 0.42 $\pm$ 0.01 & 1 & 0.6 & 0.61 $\pm$ 0.16 \\
3 & 0.4 & 0.46 $\pm$ 0.01 & 3 & 0.4 & 0.43 $\pm$ 0.01 & 3 & 0.6 & 0.7 $\pm$ 0.13 \\
5 & 0.4 & 0.46 $\pm$ 0.01 & 5 & 0.4 & 0.42 $\pm$ 0.02 & 5 & 0.6 & 0.73 $\pm$ 0.15 \\
7 & 0.4 & 0.46 $\pm$ 0.01 & 7 & 0.4 & 0.43 $\pm$ 0.01 & 7 & 0.6 & 0.73 $\pm$ 0.14 \\
1 & 0.5 & 0.43 $\pm$ 0.02 & 1 & 0.5 & 0.39 $\pm$ 0.01 & 1 & 0.7 & 0.43 $\pm$ 0.08 \\
3 & 0.5 & 0.44 $\pm$ 0.01 & 3 & 0.5 & 0.41 $\pm$ 0.01 & 3 & 0.7 & 0.66 $\pm$ 0.11 \\
5 & 0.5 & 0.44 $\pm$ 0.02 & 5 & 0.5 & 0.39 $\pm$ 0.03 & 5 & 0.7 & 0.67 $\pm$ 0.13 \\
7 & 0.5 & 0.44 $\pm$ 0.02 & 7 & 0.5 & 0.39 $\pm$ 0.03 & 7 & 0.7 & 0.68 $\pm$ 0.17 \\
1 & 0.6 & 0.42 $\pm$ 0.02 & 1 & 0.6 & 0.38 $\pm$ 0.03 & 1 & 0.8 & 0.59 $\pm$ 0.17 \\
3 & 0.6 & 0.43 $\pm$ 0.01 & 3 & 0.6 & 0.39 $\pm$ 0.02 & 3 & 0.8 & 0.71 $\pm$ 0.13 \\
5 & 0.6 & 0.43 $\pm$ 0.01 & 5 & 0.6 & 0.39 $\pm$ 0.03 & 5 & 0.8 & 0.69 $\pm$ 0.14 \\
7 & 0.6 & 0.43 $\pm$ 0.01 & 7 & 0.6 & 0.39 $\pm$ 0.03 & 7 & 0.8 & 0.67 $\pm$ 0.14 \\
\hline
\end{tabular}}
\caption{The normalized return of DWM-IQL (without $\lambda$-return) with diffusion step $K=5$, $N=3$, and simulation horizon $H\in\{1,3,5,7\}$.}
\label{tab:mbiql}
\end{table}

\begin{table}[H]
\centering
\resizebox{\textwidth}{!}{
\begin{tabular}{ccc|ccc|ccc}
\hline
\multicolumn{3}{c|}{\textbf{hopper-medium-v2}} & \multicolumn{3}{c|}{\textbf{hopper-medium-replay-v2}} & \multicolumn{3}{c}{\textbf{hopper-medium-expert-v2}} \\
\hline
Simulation Horizon & RTG & Return (Mean $\pm$ Std) & Simulation Horizon & RTG & Return (Mean $\pm$ Std) & Simulation Horizon & RTG & Return (Mean $\pm$ Std)\\
\hline
1 & 0.6 & 0.62 $\pm$ 0.1 & 1 & 0.6 & 0.4 $\pm$ 0.12 & 1 & 0.7 & 1.02 $\pm$ 0.18 \\
3 & 0.6 & 0.65 $\pm$ 0.11 & 3 & 0.6 & 0.3 $\pm$ 0.06 & 3 & 0.7 & 1.0 $\pm$ 0.21 \\
5 & 0.6 & 0.63 $\pm$ 0.12 & 5 & 0.6 & 0.29 $\pm$ 0.05 & 5 & 0.7 & 0.98 $\pm$ 0.17 \\
7 & 0.6 & 0.63 $\pm$ 0.11 & 7 & 0.6 & 0.29 $\pm$ 0.05 & 7 & 0.7 & 1.03 $\pm$ 0.14 \\
1 & 0.7 & 0.6 $\pm$ 0.11 & 1 & 0.7 & 0.27 $\pm$ 0.11 & 1 & 0.8 & 0.94 $\pm$ 0.24 \\
3 & 0.7 & 0.65 $\pm$ 0.1 & 3 & 0.7 & 0.38 $\pm$ 0.06 & 3 & 0.8 & 0.94 $\pm$ 0.24 \\
5 & 0.7 & 0.62 $\pm$ 0.14 & 5 & 0.7 & 0.36 $\pm$ 0.14 & 5 & 0.8 & 0.92 $\pm$ 0.15 \\
7 & 0.7 & 0.65 $\pm$ 0.11 & 7 & 0.7 & 0.35 $\pm$ 0.08 & 7 & 0.8 & 0.9 $\pm$ 0.28 \\
1 & 0.8 & 0.58 $\pm$ 0.07 & 1 & 0.8 & 0.25 $\pm$ 0.06 & 1 & 0.9 & 0.95 $\pm$ 0.21 \\
3 & 0.8 & 0.63 $\pm$ 0.13 & 3 & 0.8 & 0.53 $\pm$ 0.09 & 3 & 0.9 & 0.99 $\pm$ 0.17 \\
5 & 0.8 & 0.63 $\pm$ 0.11 & 5 & 0.8 & 0.43 $\pm$ 0.18 & 5 & 0.9 & 0.97 $\pm$ 0.2 \\
7 & 0.8 & 0.62 $\pm$ 0.09 & 7 & 0.8 & 0.4 $\pm$ 0.11 & 7 & 0.9 & 0.97 $\pm$ 0.17 \\
\hline
\multicolumn{3}{c|}{\textbf{walker2d-medium-v2}} & \multicolumn{3}{c|}{\textbf{walker2d-medium-replay-v2}} & \multicolumn{3}{c}{\textbf{walker2d-medium-expert-v2}} \\
\hline
Simulation Horizon & RTG & Return (Mean $\pm$ Std) & Simulation Horizon & RTG & Return (Mean $\pm$ Std) & Simulation Horizon & RTG & Return (Mean $\pm$ Std) \\
\hline
1 & 0.6 & 0.7 $\pm$ 0.15 & 1 & 0.6 & 0.46 $\pm$ 0.19 & 1 & 0.8 & 1.08 $\pm$ 0.01 \\
3 & 0.6 & 0.67 $\pm$ 0.15 & 3 & 0.6 & 0.29 $\pm$ 0.16 & 3 & 0.8 & 1.09 $\pm$ 0.0 \\
5 & 0.6 & 0.61 $\pm$ 0.2 & 5 & 0.6 & 0.3 $\pm$ 0.18 & 5 & 0.8 & 1.09 $\pm$ 0.0 \\
7 & 0.6 & 0.6 $\pm$ 0.17 & 7 & 0.6 & 0.23 $\pm$ 0.17 & 7 & 0.8 & 1.1 $\pm$ 0.0 \\
1 & 0.7 & 0.65 $\pm$ 0.17 & 1 & 0.7 & 0.37 $\pm$ 0.14 & 1 & 0.9 & 1.08 $\pm$ 0.03 \\
3 & 0.7 & 0.68 $\pm$ 0.17 & 3 & 0.7 & 0.31 $\pm$ 0.19 & 3 & 0.9 & 1.09 $\pm$ 0.0 \\
5 & 0.7 & 0.63 $\pm$ 0.12 & 5 & 0.7 & 0.27 $\pm$ 0.18 & 5 & 0.9 & 1.09 $\pm$ 0.02 \\
7 & 0.7 & 0.63 $\pm$ 0.2 & 7 & 0.7 & 0.28 $\pm$ 0.28 & 7 & 0.9 & 1.09 $\pm$ 0.04 \\
1 & 0.8 & 0.64 $\pm$ 0.13 & 1 & 0.8 & 0.43 $\pm$ 0.2 & 1 & 1.0 & 1.08 $\pm$ 0.0 \\
3 & 0.8 & 0.7 $\pm$ 0.16 & 3 & 0.8 & 0.31 $\pm$ 0.14 & 3 & 1.0 & 1.09 $\pm$ 0.01 \\
5 & 0.8 & 0.68 $\pm$ 0.15 & 5 & 0.8 & 0.37 $\pm$ 0.22 & 5 & 1.0 & 1.09 $\pm$ 0.0 \\
7 & 0.8 & 0.61 $\pm$ 0.19 & 7 & 0.8 & 0.29 $\pm$ 0.23 & 7 & 1.0 & 1.1 $\pm$ 0.0 \\
\hline
\multicolumn{3}{c|}{\textbf{halfcheetah-medium-v2}} & \multicolumn{3}{c|}{\textbf{halfcheetah-medium-replay-v2}} & \multicolumn{3}{c}{\textbf{halfcheetah-medium-expert-v2}} \\
\hline
1 & 0.4 & 0.44 $\pm$ 0.01 & 1 & 0.4 & 0.43 $\pm$ 0.02 & 1 & 0.6 & 0.44 $\pm$ 0.14 \\
3 & 0.4 & 0.46 $\pm$ 0.01 & 3 & 0.4 & 0.43 $\pm$ 0.01 & 3 & 0.6 & 0.58 $\pm$ 0.17 \\
5 & 0.4 & 0.45 $\pm$ 0.01 & 5 & 0.4 & 0.43 $\pm$ 0.02 & 5 & 0.6 & 0.71 $\pm$ 0.16 \\
7 & 0.4 & 0.45 $\pm$ 0.01 & 7 & 0.4 & 0.43 $\pm$ 0.02 & 7 & 0.6 & 0.75 $\pm$ 0.16 \\
1 & 0.5 & 0.43 $\pm$ 0.02 & 1 & 0.5 & 0.41 $\pm$ 0.03 & 1 & 0.7 & 0.43 $\pm$ 0.08 \\
3 & 0.5 & 0.44 $\pm$ 0.01 & 3 & 0.5 & 0.41 $\pm$ 0.02 & 3 & 0.7 & 0.66 $\pm$ 0.11 \\
5 & 0.5 & 0.44 $\pm$ 0.01 & 5 & 0.5 & 0.41 $\pm$ 0.01 & 5 & 0.7 & 0.69 $\pm$ 0.13 \\
7 & 0.5 & 0.44 $\pm$ 0.02 & 7 & 0.5 & 0.41 $\pm$ 0.02 & 7 & 0.7 & 0.49 $\pm$ 0.13 \\
1 & 0.6 & 0.42 $\pm$ 0.02 & 1 & 0.6 & 0.39 $\pm$ 0.03 & 1 & 0.8 & 0.66 $\pm$ 0.17 \\
3 & 0.6 & 0.43 $\pm$ 0.01 & 3 & 0.6 & 0.39 $\pm$ 0.02 & 3 & 0.8 & 0.67 $\pm$ 0.17 \\
5 & 0.6 & 0.43 $\pm$ 0.01 & 5 & 0.6 & 0.39 $\pm$ 0.03 & 5 & 0.8 & 0.67 $\pm$ 0.2 \\
7 & 0.6 & 0.43 $\pm$ 0.01 & 7 & 0.6 & 0.39 $\pm$ 0.03 & 7 & 0.8 & 0.65 $\pm$ 0.14 \\
\hline
\end{tabular}}
\caption{The normalized return of DWM-TD3BC with diffusion step $K=5$, $N=0.5$, and simulation horizon $H\in\{1,3,5,7\}$.}
\label{tab:mbtd3bc}
\end{table}

\begin{table}[H]
\centering
\resizebox{\textwidth}{!}{
\begin{tabular}{ccc|ccc|ccc}
\hline
\multicolumn{3}{c|}{\textbf{hopper-medium-v2}} & \multicolumn{3}{c|}{\textbf{hopper-medium-replay-v2}} & \multicolumn{3}{c}{\textbf{hopper-medium-expert-v2}} \\
\hline
Pessimism $\kappa$ & RTG & Return (Mean $\pm$ Std) & Pessimism $\kappa$ & RTG & Return (Mean $\pm$ Std) & Pessimism $\kappa$ & RTG & Return (Mean $\pm$ Std)\\
\hline
0.01 & 0.6 &  0.49 $\pm$ 0.11 & 0.01 & 0.6  & 0.39 $\pm$ 0.03 & 0.01 & 0.7  & 0.78 $\pm$ 0.18\\
0.01 & 0.7 &  0.48 $\pm$ 0.08 & 0.01 & 0.7  & 0.26 $\pm$ 0.02 & 0.01 & 0.8  & 0.80 $\pm$ 0.18\\
0.01 & 0.8 &  0.48 $\pm$ 0.10 & 0.01 & 0.8  & 0.32 $\pm$ 0.07 & 0.01 & 0.9  & 0.65 $\pm$ 0.16\\
0.1 & 0.6 &  0.47 $\pm$ 0.04 & 0.1 & 0.6  & 0.28 $\pm$ 0.06 & 0.1 & 0.7  & 0.74 $\pm$ 0.27\\
0.1 & 0.7 &  0.49 $\pm$ 0.07 & 0.1 & 0.7  & 0.33 $\pm$ 0.02 & 0.1 & 0.8  & 0.75 $\pm$ 0.21\\
0.1 & 0.8 &  0.48 $\pm$ 0.08 & 0.1 & 0.8  & 0.28 $\pm$ 0.04 & 0.1 & 0.9  & 0.71 $\pm$ 0.19\\
1.0 & 0.6 &  0.48 $\pm$ 0.07 & 1.0 & 0.6  & 0.24 $\pm$ 0.02 & 1.0 & 0.7  & 0.63 $\pm$ 0.13\\
1.0 & 0.7 &  0.50 $\pm$ 0.09 & 1.0 & 0.7  & 0.25 $\pm$ 0.03 & 1.0 & 0.8  & 0.52 $\pm$ 0.13\\
1.0 & 0.8 &  0.47 $\pm$ 0.07 & 1.0 & 0.8  & 0.27 $\pm$ 0.05 & 1.0 & 0.9  & 0.48 $\pm$ 0.21\\
\hline
\multicolumn{3}{c|}{\textbf{walker2d-medium-v2}} & \multicolumn{3}{c|}{\textbf{walker2d-medium-replay-v2}} & \multicolumn{3}{c}{\textbf{walker2d-medium-expert-v2}} \\
\hline
Pessimism $\kappa$ & RTG & Return (Mean $\pm$ Std) & Pessimism $\kappa$ & RTG & Return (Mean $\pm$ Std) & Pessimism $\kappa$ & RTG & Return (Mean $\pm$ Std) \\
\hline
0.01 & 0.6 & 0.61 $\pm$ 0.19 & 0.01 & 0.6 & 0.12 $\pm$ 0.10 & 0.01 & 0.8 & 1.09 $\pm$ 0.04 \\
0.01 & 0.7 & 0.62 $\pm$ 0.21 & 0.01 & 0.7 & 0.18 $\pm$ 0.11 & 0.01 & 0.9 & 1.07 $\pm$ 0.05 \\
0.01 & 0.8 & 0.61 $\pm$ 0.18 & 0.01 & 0.8 & 0.19 $\pm$ 0.12 & 0.01 & 1.0 & 1.06 $\pm$ 0.07 \\
0.1 & 0.6 & 0.62 $\pm$ 0.19 & 0.1 & 0.6 & 0.14 $\pm$ 0.09 & 0.1 & 0.8 & 1.08 $\pm$ 0.06 \\
0.1 & 0.7 & 0.62 $\pm$ 0.21 & 0.1 & 0.7 & 0.16 $\pm$ 0.09 & 0.1 & 0.9 & 1.06 $\pm$ 0.09 \\
0.1 & 0.8 & 0.65 $\pm$ 0.18 & 0.1 & 0.8 & 0.19 $\pm$ 0.11 & 0.1 & 1.0 & 1.04 $\pm$ 0.12 \\
1.0 & 0.6 & 0.76 $\pm$ 0.13 & 1.0 & 0.6 & 0.24 $\pm$ 0.11 & 1.0 & 0.8 & 1.09 $\pm$ 0.01 \\
1.0 & 0.7 & 0.76 $\pm$ 0.14 & 1.0 & 0.7 & 0.35 $\pm$ 0.13 & 1.0 & 0.9 & 1.09 $\pm$ 0.01 \\
1.0 & 0.8 & 0.79 $\pm$ 0.08 & 1.0 & 0.8 & 0.34 $\pm$ 0.18 & 1.0 & 1.0 & 1.10 $\pm$ 0.01 \\
\hline
\multicolumn{3}{c|}{\textbf{halfcheetah-medium-v2}} & \multicolumn{3}{c|}{\textbf{halfcheetah-medium-replay-v2}} & \multicolumn{3}{c}{\textbf{halfcheetah-medium-expert-v2}} \\
\hline
Pessimism $\kappa$ & RTG & Return (Mean $\pm$ Std) & Pessimism $\kappa$ & RTG & Return (Mean $\pm$ Std) & Pessimism $\kappa$ & RTG & Return (Mean $\pm$ Std) \\
\hline
0.01 & 0.4 & 0.43 $\pm$ 0.01 & 0.01 & 0.4 & 0.42 $\pm$ 0.01 & 0.01 & 0.6 & 0.55 $\pm$ 0.09 \\
0.01 & 0.5 & 0.44 $\pm$ 0.01 & 0.01 & 0.5 & 0.39 $\pm$ 0.04 & 0.01 & 0.7 & 0.66 $\pm$ 0.17\\
0.01 & 0.6 & 0.42 $\pm$ 0.02 & 0.01 & 0.6 & 0.39 $\pm$ 0.03 & 0.01 & 0.8 & 0.64 $\pm$ 0.11\\
0.1 & 0.4 & 0.43 $\pm$ 0.01 & 0.1 & 0.4 & 0.42 $\pm$ 0.01 & 0.1 & 0.6 & 0.54 $\pm$ 0.13\\
0.1 & 0.5 & 0.44 $\pm$ 0.01 & 0.1 & 0.5 & 0.4 $\pm$ 0.03 & 0.1 & 0.7 & 0.62 $\pm$ 0.09\\
0.1 & 0.6 & 0.42 $\pm$ 0.02 & 0.1 & 0.6 & 0.38 $\pm$ 0.04 & 0.1 & 0.8 & 0.67 $\pm$ 0.13\\
1.0 & 0.4 & 0.43 $\pm$ 0.01 & 1.0 & 0.4 & 0.43 $\pm$ 0.01 & 1.0 & 0.6 & 0.61 $\pm$ 0.18\\
1.0 & 0.5 & 0.44 $\pm$ 0.01 & 1.0 & 0.5 & 0.4 $\pm$ 0.04 & 1.0 & 0.7 & 0.69 $\pm$ 0.13\\
1.0 & 0.6 & 0.44 $\pm$ 0.02 & 1.0 & 0.6 & 0.38 $\pm$ 0.04 & 1.0 & 0.8 & 0.64 $\pm$ 0.16\\
\hline
\end{tabular}
}
\caption{The normalized return of DWM-PQL with diffusion step $K=5$, $N=3$, and simulation horizon $H=5$.}
\label{tab:mbpql}
\end{table}


\end{document}